\documentclass{article}
\usepackage[paper=a4paper,textwidth=170mm,textheight=260mm]{geometry}
\usepackage{times}
\usepackage[colorlinks=true,
            linkcolor=red,
            urlcolor=blue,
            citecolor=blue]{hyperref}
\usepackage{url}
\usepackage{graphicx}
\usepackage{subfigure}
\usepackage{amsfonts,amssymb,amsmath,amsthm}
\usepackage{color}
\usepackage{url}
\usepackage{tablefootnote}
\usepackage{mathrsfs}
\usepackage{array}
\usepackage{enumerate}
\usepackage{enumitem}

\usepackage{graphicx} 
\usepackage{subfigure} 

\usepackage{natbib}

\usepackage{hyperref}
\newcommand*\samethanks[1][\value{footnote}]{\footnotemark[#1]}

\title{Analysis of classifiers' robustness to adversarial perturbations}

\title{Analysis of classifiers' robustness to adversarial perturbations}
\date{}
\author{Alhussein Fawzi\thanks{Ecole Polytechnique Federale de Lausanne (EPFL), Signal Processing Laboratory (LTS4), Lausanne 1015-Switzerland. Email: (alhussein.fawzi@epfl.ch, pascal.frossard@epfl.ch)} \and
Omar Fawzi\thanks{ENS de Lyon, LIP, UMR 5668 ENS Lyon - CNRS - UCBL - INRIA, Universit\'e de Lyon, France. Email: omar.fawzi@ens-lyon.fr}
\and Pascal Frossard\samethanks[1]}

\newcommand{\bb}[1]{\mathbb{#1}}

\DeclareMathOperator{\dist}{dist}

\theoremstyle{definition}
 \newtheorem{theorem}{Theorem}[section]
 \newtheorem{lemma}[theorem]{Lemma}

\begin{document} 

\maketitle

\begin{abstract}
The goal of this paper is to analyze an intriguing phenomenon recently discovered in deep networks, namely their instability to adversarial perturbations \citep{szegedy2013intriguing}. We provide a theoretical framework for analyzing the robustness of classifiers to adversarial perturbations, and show fundamental upper bounds on the robustness of classifiers. Specifically, we establish a general upper bound on the robustness of classifiers to adversarial perturbations, and then illustrate the obtained upper bound on two practical classes of classifiers, namely the linear and quadratic classifiers. In both cases, our upper bound depends on a \textit{distinguishability} measure that captures the notion of \textit{difficulty} of the classification task. Our results for both classes imply that in tasks involving small distinguishability, \textit{no classifier} in the considered set will be robust to adversarial perturbations, even if a good accuracy is achieved. Our theoretical framework moreover suggests that the phenomenon of adversarial instability is due to the low flexibility of classifiers, compared to the difficulty of the classification task (captured mathematically by the distinguishability measure).
Moreover, we further show the existence of a clear distinction between the robustness of a classifier to random noise and its robustness to adversarial perturbations. Specifically, the former is shown to be larger than the latter by a factor that is proportional to $\sqrt{d}$ (with $d$ being the signal dimension) for linear classifiers. This result gives a theoretical explanation for the discrepancy between the two robustness properties in high dimensional problems, which was empirically observed in \citet{szegedy2013intriguing} in the context of neural networks. To the best of our knowledge, our results provide the first theoretical work that addresses the phenomenon of adversarial instability recently observed for deep networks. We finally show experimental results on controlled and real-world data that confirm the theoretical analysis and extends its spirit to more complex classification schemes.
\end{abstract} 

\section{Introduction}
\label{sec:introduction}


State-of-the-art deep networks have recently been shown to be surprisingly unstable to adversarial perturbations \citep{szegedy2013intriguing}. Unlike random noise, adversarial perturbations are \textit{minimal} perturbations that are sought to switch the estimated label of the classifier. On vision tasks, the results of \cite{szegedy2013intriguing} have shown that perturbations that are hardly perceptible to the human eye are sufficient to change the decision of a deep network, even if the classifier has a performance that is close to the human visual system. This surprising instability raises interesting theoretical questions that we initiate in this paper.
What causes classifiers to be unstable to adversarial perturbations? Are deep networks the only classifiers that have such unstable behaviour? Is it at all possible to design training algorithms to build deep networks that are robust or is the instability to adversarial noise an inherent feature of all deep networks? Can we quantify the difference between random noise and adversarial noise? Providing theoretical answers to these questions is crucial in order to achieve the goal of building classifiers that are robust to adversarial hostile perturbations.

In this paper, we introduce a framework to formally study the robustness of classifiers to adversarial perturbations in the binary setting. We provide a general upper bound on the robustness of classifiers to adversarial perturbations, and then illustrate and specialize the obtained upper bound for the families of linear and quadratic classifiers. In both cases, our results show the existence of a fundamental limit on the robustness to adversarial perturbations. This limit is expressed in terms of a \textit{distinguishability} measure between the classes, which depends on the considered family of classifiers. Specifically, for linear classifiers, the distinguishability is defined as the distance between the means of the two classes, while for quadratic classifiers, it is defined as the distance between the matrices of second order moments of the two classes. For both classes of functions, our upper bound on the robustness is valid \textit{for all classifiers in the family independently of the training procedure}, and we see the fact that the bound is independent of the training procedure as a strength. This result has the following important implication: in difficult classification tasks involving a small value of distinguishability, \textit{any} classifier in the set with low misclassification rate is vulnerable to adversarial perturbations. Importantly, the distinguishability parameter related to quadratic classifiers is much larger than that of linear classifiers for many datasets of interest, and suggests that it is harder to find adversarial examples for more \textit{flexible} classifiers.
We further compare the robustness to adversarial perturbations of linear classifiers to the more traditional notion of robustness to random uniform noise. The latter robustness is shown to be larger than the former by a factor of $\sqrt{d}$ (with $d$ the dimension of input signals), thereby showing that in high dimensional classification tasks, linear classifiers can be robust to random noise even for small values of the distinguishability. We illustrate the newly introduced concepts and our theoretical results on a running example used throughout the paper. We complement our theoretical analysis with experimental results, and show that the intuition obtained from the theoretical analysis also holds for more complex classifiers. 

The phenomenon of adversarial instability has recently attracted a lot of attention from the deep network community. Following the original paper \citep{szegedy2013intriguing}, several attempts have been made to make deep networks robust to adversarial perturbations \citep{chalupka2014visual, gu2014towards}. Moreover, a distinct but related phenomenon has been explored in \cite{nguyen2014deep}. Closer to our work, the authors of \citep{goodfellow2014explaining} provided an empirical explanation of the phenomenon of adversarial instability, and designed an efficient method to find adversarial examples. Specifically, contrarily to the original explanation provided in \cite{szegedy2013intriguing}, the authors argue that it is the ``linear'' nature of deep nets that causes the adversarial instability. Instead, our paper adopts a rigorous mathematical perspective to the problem of adversarial instability and shows that adversarial instability is due to the \textit{low flexibility} of classifiers compared to the difficulty of the classification task. 

Our work should not be confused with works on the security of machine learning algorithms under adversarial attacks \citep{biggio2012poisoning, barreno2006can, dalvi2004adversarial}. These works specifically study attacks that manipulate \textit{the learning system} (e.g., change the decision function by injecting malicious training points), as well as defense strategies to counter these attacks. This setting significantly differs from ours, as we examine the robustness of \textit{a fixed} classifier to adversarial perturbations (that is, the classifier cannot be manipulated). The stability of learning algorithms has also been defined and extensively studied in \citep{bousquet2002stability, lugosi1994posterior}. Again, this notion of stability differs from the one studied here, as we are interested in the robustness of fixed classifiers, and not of learning algorithms.

The construction of learning algorithms that achieve robustness of classifiers to data corruption has been an active area of research in machine learning and robust optimization (see e.g., \cite{sra2012optimization} and references therein). For  a specific disturbance model on the data samples, the robust optimization approach for constructing robust classifiers seeks to minimize the worst possible empirical error under such disturbances \citep{lanckriet2003robust, xu2009robustness}. It is shown that, for many disturbance models, the desired objective function can be written as a tractable convex optimization problem. Our work studies the robustness of classifiers from a different perspective; we establish upper bounds on the robustness of classifiers \textit{independently of the learning algorithms}. That is, using our bounds, we can certify the instability of a class of classifiers to adversarial perturbations, 
 independently of the learning mechanism. In other words, while algorithmic and optimization aspects of robust classifiers have been studied in the above works, we focus on fundamental limits on the adversarial robustness of classifiers that are independent of the learning scheme.  


The paper is structured as follows: Sec. \ref{sec:problem_setting} introduces the problem setting. In Sec. \ref{sec:horizontal_vertical}, we introduce a running example that is used throughout the paper. We introduce in Sec. \ref{sec:upper_limit} a theoretical framework for studying the robustness to adversarial perturbations. In the following two sections, two case studies are analyzed in detail. The robustness of linear classifiers (to adversarial and random noise) is studied in Sec. \ref{sec:lin_classifiers}. In Sec. \ref{sec:quad_classifiers}, we study the adversarial robustness of quadratic classifiers. Experimental results illustrating our theoretical analysis are given in Section \ref{sec:exp_results}. Proofs and additional discussion on the choice of the norms to measure perturbations are finally deferred to the appendix. 

\section{Problem setting}
\label{sec:problem_setting}

We first introduce the framework and notations that are used for analyzing the robustness of classifiers to adversarial and uniform random noise. We restrict our analysis to the binary classification task, for simplicity. We expect similar conclusions for the multi-class case, but we leave that for future work. 
Let $\mu$ denote the probability measure on $\bb{R}^d$ of the data points that we wish to classify, and $y(x) \in \{ -1, 1\}$ be the label of a point $x \in \bb{R}^d$. The distribution $\mu$ is assumed to be of bounded support. That is, $\bb{P}_{\mu} (x \in \mathcal{B}) = 1$, with $\mathcal{B} = \{ x \in \mathbb{R}^d: \| x \|_2 \leq M \}$ for some $M > 0$. We further denote by $\mu_{1}$ and $\mu_{-1}$ the distributions of class $1$ and class $-1$ in $\bb{R}^d$, respectively. Let $f: \bb{R}^d \rightarrow \bb{R}$ be an arbitrary classification function. The classification rule associated to $f$ is simply obtained by taking the sign of $f(x)$. 
The performance of a classifier $f$ is usually measured by its \textit{risk}, defined as the probability of misclassification according to $\mu$: 
\begin{align*}
R(f) & = \bb{P}_{\mu} (\text{sign} (f(x)) \neq y(x)) \\
	  & = p_1 \bb{P}_{\mu_1} (f(x) < 0) + p_{-1} \bb{P}_{\mu_{-1}} (f(x) \geq 0),
\end{align*}
where $p_{\pm 1} = \bb{P}_{\mu} (y(x) = \pm 1)$.

The focus of this paper is to study the robustness of classifiers to adversarial perturbations in the ambient space $\bb{R}^d$.  Given a datapoint $x \in \bb{R}^d$ sampled from $\mu$, we denote by $\Delta_{\text{adv}} (x;f)$ the norm of the smallest perturbation that switches the sign\footnote{We make the assumption that a perturbation $r$ that satisfies the equality $f(x+r) = 0$ flips the estimated label of $x$.} of $f$:
\begin{align}
\label{eq:adversarial_function}
\Delta_{\text{adv}} (x; f) = \min_{r \in \bb{R}^d} \| r \|_2 \text{ subject to } f(x) f(x+r) \leq 0.
\end{align}
Here, we use the $\ell_2$ norm to quantify the perturbation; we refer the reader to Appendix \ref{sec:discussion_norms} for a discussion of the norm choice. Unlike random noise, the above definition corresponds to a minimal noise, where the perturbation $r$ is sought to flip the estimated label of $x$. This justifies the \textit{adversarial} nature of the perturbation. 
It is important to note that, while $x$ is a datapoint sampled according to $\mu$, the perturbed point $x+r$ is not required to belong to the dataset (i.e., $x+r$ can be outside the support of $\mu$). 
The robustness to adversarial perturbation of $f$ is defined as the average of $\Delta_{\text{adv}} (x;f)$ over all $x$: 
\begin{align}
\label{eq:adv_noise}
\rho_{\text{adv}}(f) = \bb{E}_\mu (\Delta_{\text{adv}}(x; f)). 
\end{align}
In words, $\rho_{\text{adv}} (f)$ is defined as the average norm of the minimal perturbations required to flip the estimated labels of the datapoints. Note that $\rho_{\text{adv}} (f)$ is a property of both the classifier $f$ and the distribution $\mu$, but it is independent of the true labels of the datapoints $y$.\footnote{In that aspect, our definition slightly differs from the one proposed in \cite{szegedy2013intriguing}, which defines the robustness to adversarial perturbations as the average of the norms of the minimal perturbations required to \textit{misclassify} all datapoints. As our notion of robustness is larger, the upper bounds derived in our paper also directly apply for the definition of robustness in \cite{szegedy2013intriguing}.} Moreover, it should be noted that $\rho_{\text{adv}}$ is different from the margin considered by SVMs. In fact, SVM margins are traditionally defined as the \textit{minimal} distance to the (linear) boundary over all training points, while $\rho_{\text{adv}}$ is defined as the \textit{average} distance to the boundary over all training points. In addition, distances in our case are measured in the input space, while the margin is defined in the feature space for kernel SVMs.
\begin{figure}[t]
\centering
\includegraphics[width=0.3\textwidth]{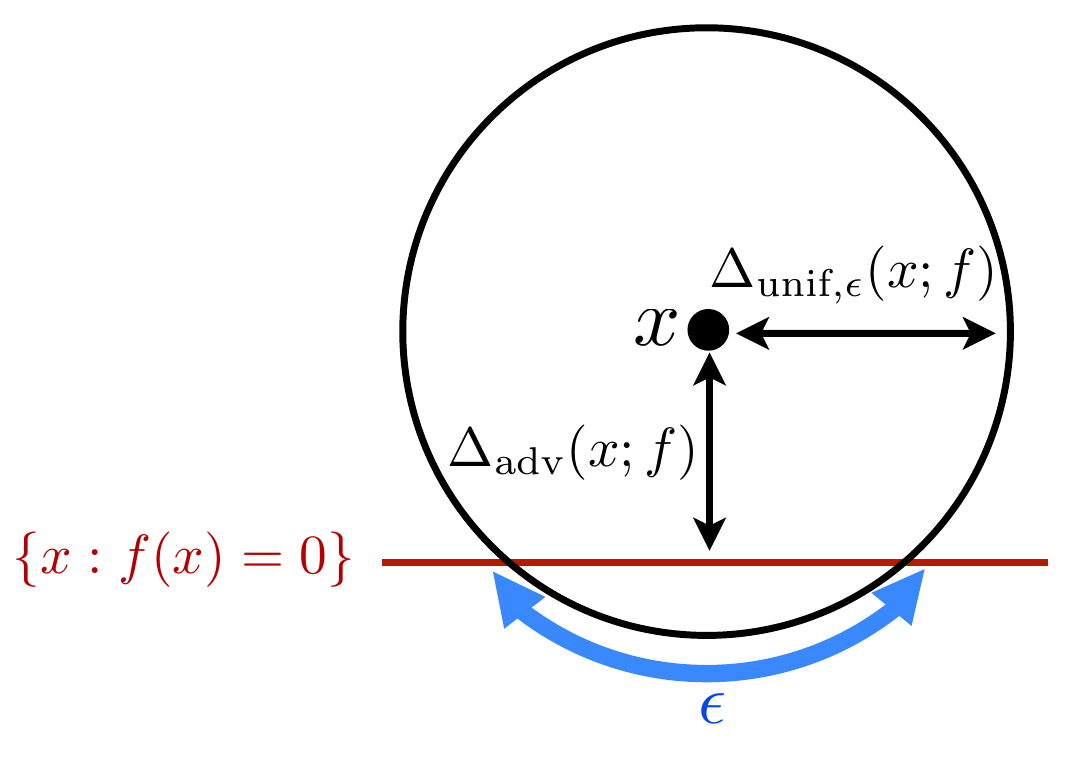}
\caption{\label{fig:schema_delta_unif_adv} Illustration of $\Delta_{\text{adv}} (x;f)$ and  $\Delta_{\text{unif}, \epsilon} (x;f)$. The red line represents the classifier boundary. In this case, the quantity $\Delta_{\text{adv}} (x;f)$ is equal to the distance from $x$ to this line. The radius of the sphere drawn around $x$ is $\Delta_{\text{unif}, \epsilon} (x;f)$. Assuming $f(x) > 0$, observe that the spherical cap in the region below the line has measure $\epsilon$, which means that the probability that a random point sampled on the sphere has label $+1$ is $1-\epsilon$.}
\end{figure}
In this paper, we also study the robustness of classifiers to random uniform noise, that we define as follows. For a given $\epsilon \in [0,1]$, let
\begin{align}
\label{eq:Delta_unif}
\Delta_{\text{unif}, \epsilon}(x;f) = & \max_{\eta \geq 0} \eta \\ 
& \text{ s.t. } \bb{P}_{n \sim \eta \bb{S}} (f(x) f(x+n) \leq 0) \leq \epsilon, \nonumber
\end{align}
where $\eta \bb{S}$ denotes the uniform measure on the sphere centered at $0$ and of radius $\eta$ in $\bb{R}^d$. In words, $\Delta_{\text{unif}, \epsilon}(x;f)$ denotes the maximal radius of the sphere centered at $x$, such that perturbed points sampled uniformly at random from this sphere are classified similarly to $x$ with high probability. 
An illustration of $\Delta_{\text{unif}, \epsilon} (x;f)$ and $\Delta_{\text{adv}} (x;f)$ is given in Fig. \ref{fig:schema_delta_unif_adv}.
Similarly to adversarial perturbations, the point $x+n$ will lie outside the support of $\mu$, in general. Note moreover that $\Delta_{\text{unif}, \epsilon} (x;f)$ provides an upper bound on $\Delta_{\text{adv}} (x;f)$, for all $\epsilon$. 
The $\epsilon$-robustness of $f$ to random uniform noise is defined by:
\begin{align}
\rho_{\text{unif}, \epsilon} (f) = \bb{E}_{\mu} (\Delta_{\text{unif}, \epsilon} (x; f)).
\end{align}
We summarize the quantities of interest in Table \ref{tab:quantites_interest}. 

\begin{table}
\centering
\begin{small}
\begin{tabular}{|l|c|c|}
\hline
 \textbf{Quantity} & \textbf{Definition} & \textbf{Dependence} \\
 \hline
Risk & $R(f) = \bb{P}_{\mu} (\text{sign} (f(x)) \neq y(x))$ & $\mu, y, f$ \\ \hline
Robustness to adversarial perturbations & $\rho_{\text{adv}} (f) = \bb{E}_{\mu} (\Delta_{\text{adv}}(x; f))$ & $\mu, f$ \\ \hline
Robustness to random uniform noise & $\rho_{\text{unif}, \epsilon} (f) = \bb{E}_{\mu} (\Delta_{\text{unif}, \epsilon} (x;f))$ & $\mu, f$ \\ \hline
\end{tabular}
\end{small}
\caption{\label{tab:quantites_interest}Quantities of interest in the paper and their dependencies.}
\end{table}
\section{Running example}
\label{sec:horizontal_vertical}
\begin{figure}[t]
\centering
\subfigure[]{
\includegraphics[width=0.08\textwidth]{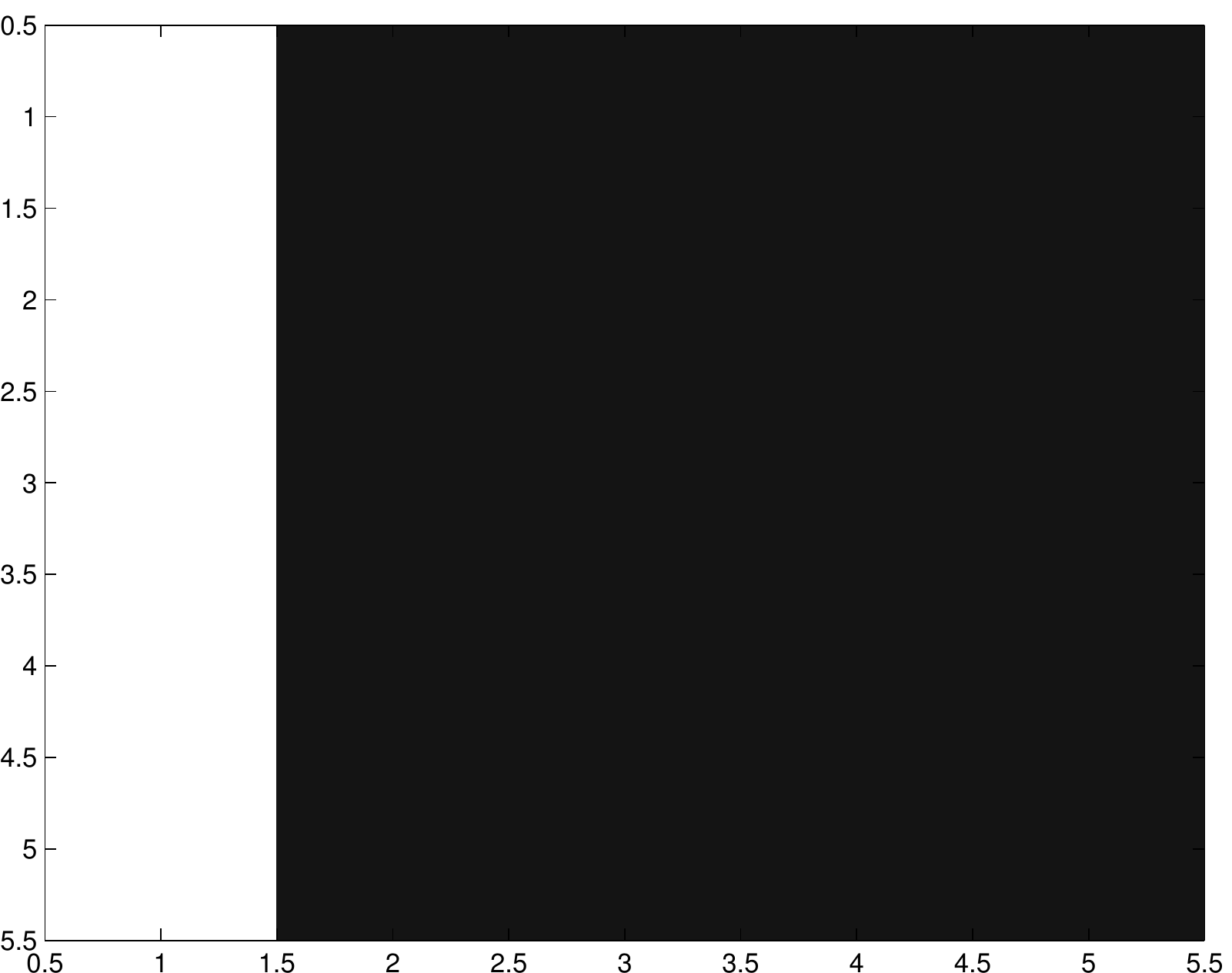}
}
\centering
\subfigure[]{
\includegraphics[width=0.08\textwidth]{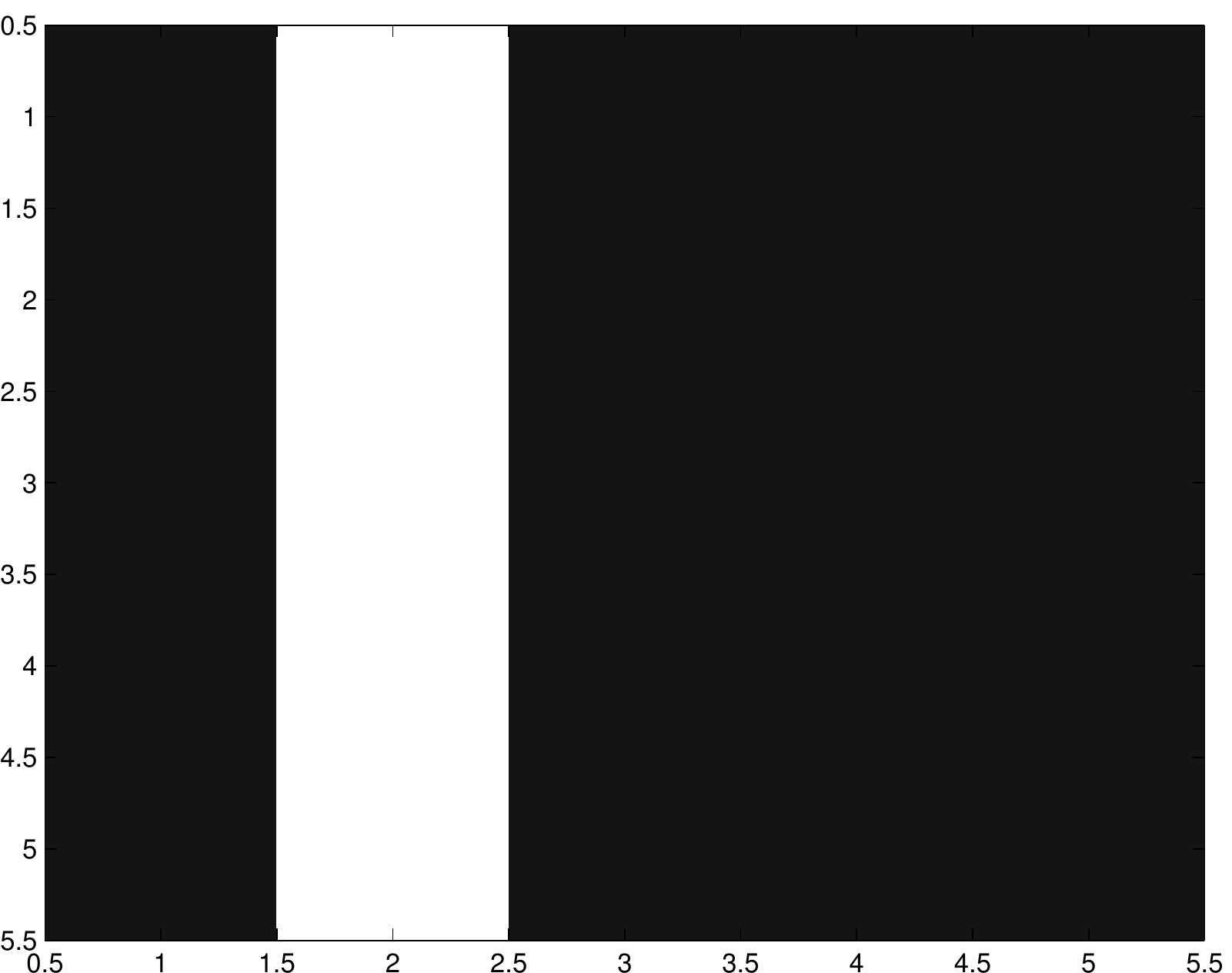}
} 
\centering
\subfigure[]{
\includegraphics[width=0.08\textwidth]{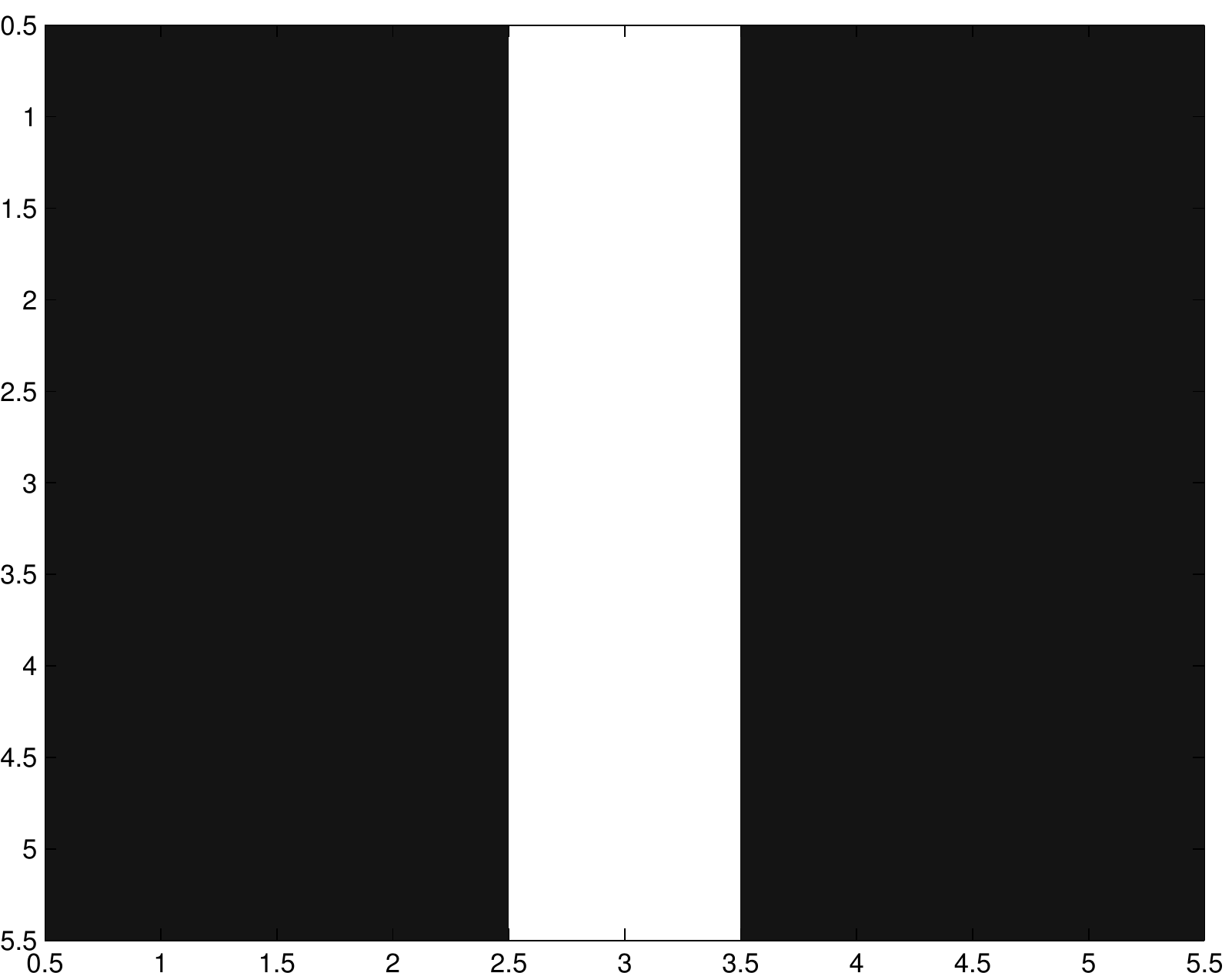}
}
\centering
\subfigure[]{
\includegraphics[width=0.08\textwidth]{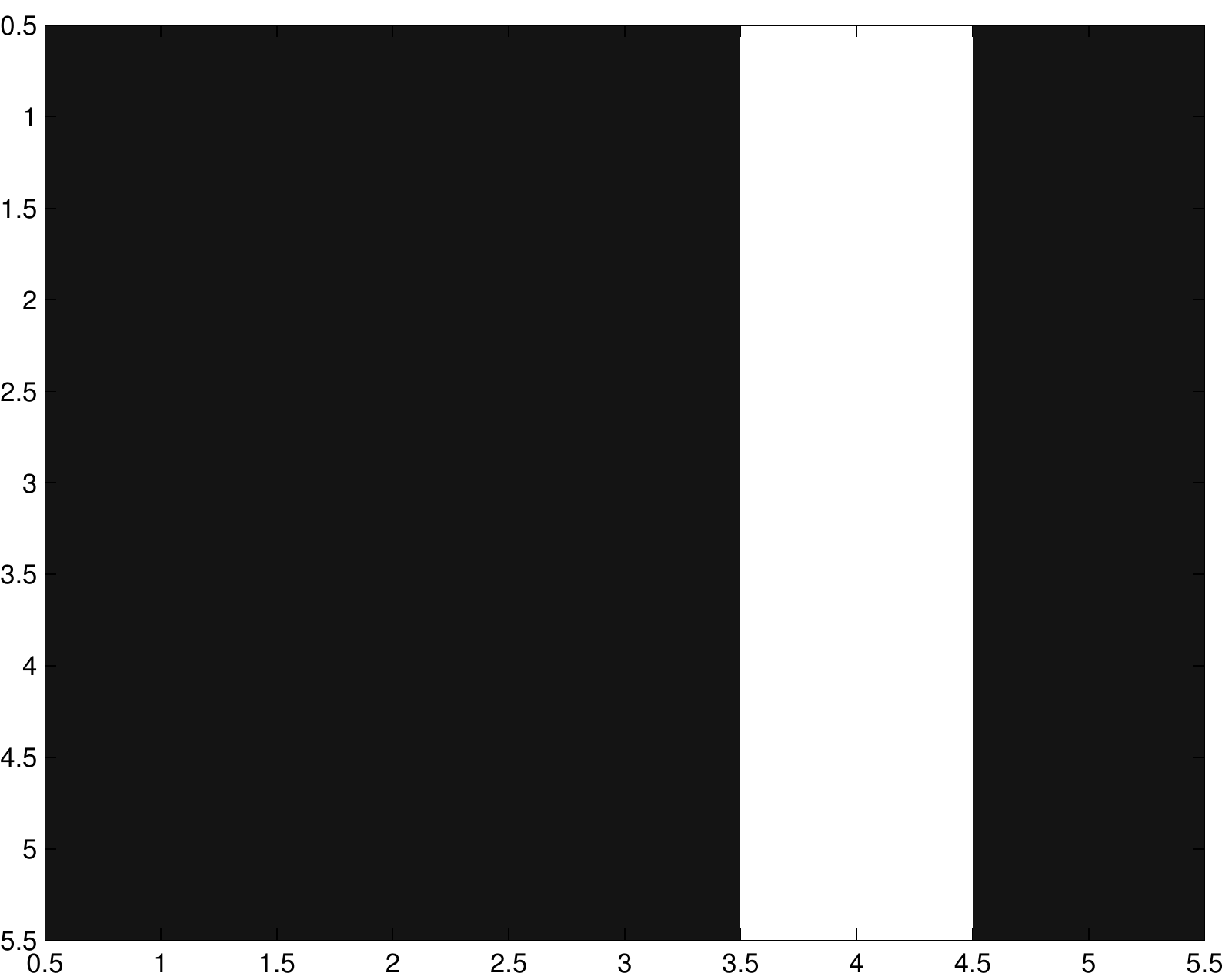}
} 
\centering
\subfigure[]{
\includegraphics[width=0.08\textwidth]{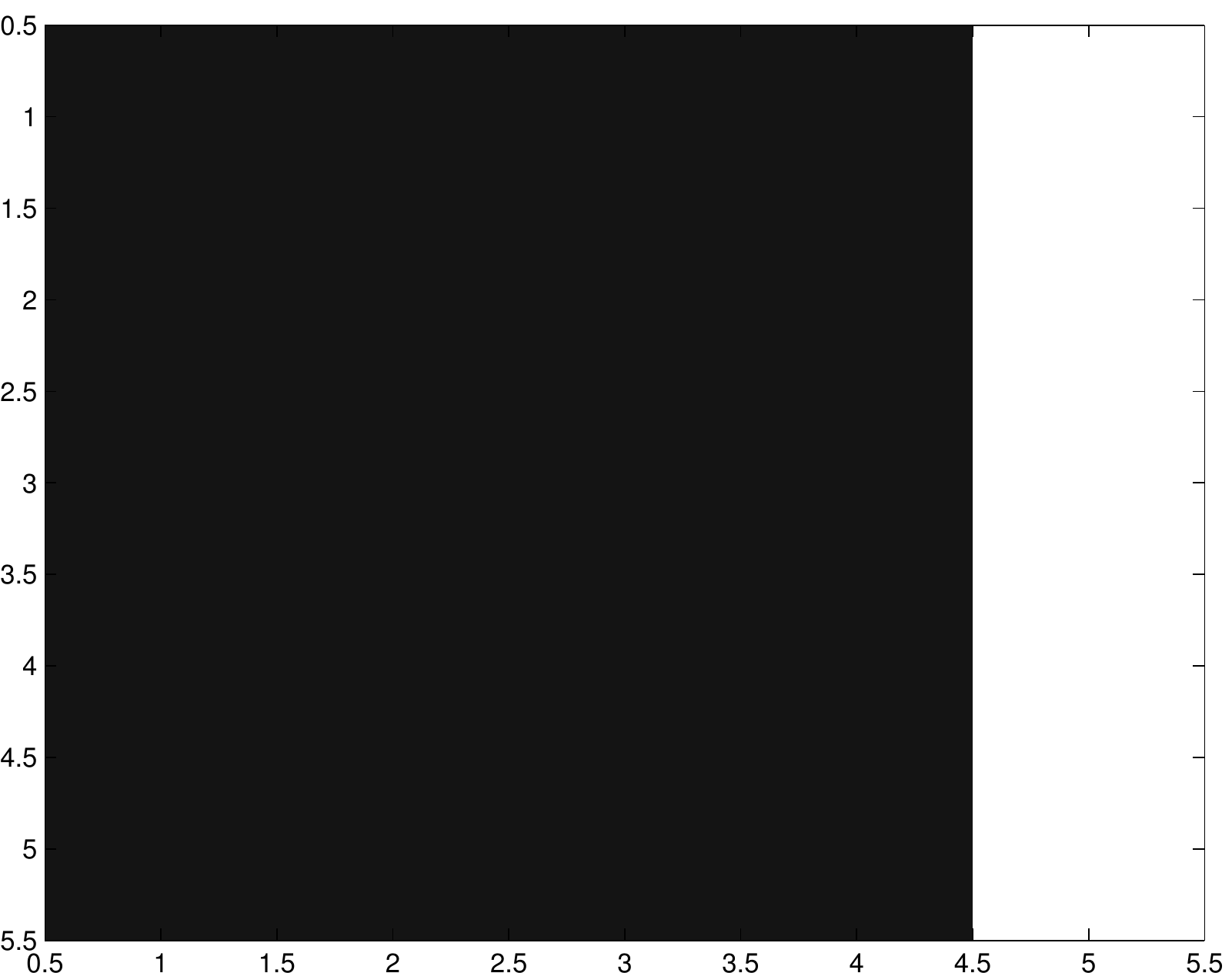}
} 
\\
\centering
\subfigure[]{
\includegraphics[width=0.08\textwidth]{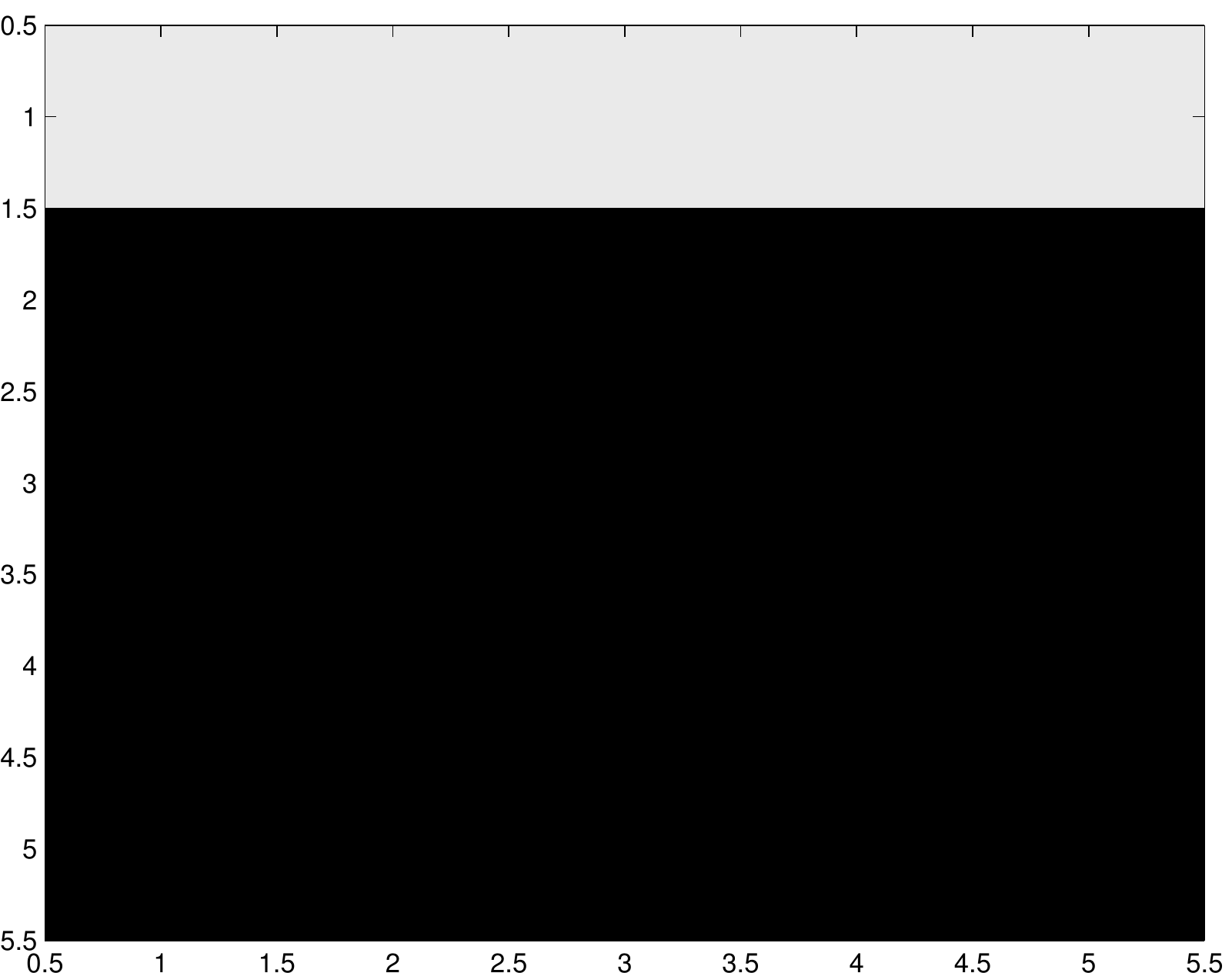}
}
\centering
\subfigure[]{
\includegraphics[width=0.08\textwidth]{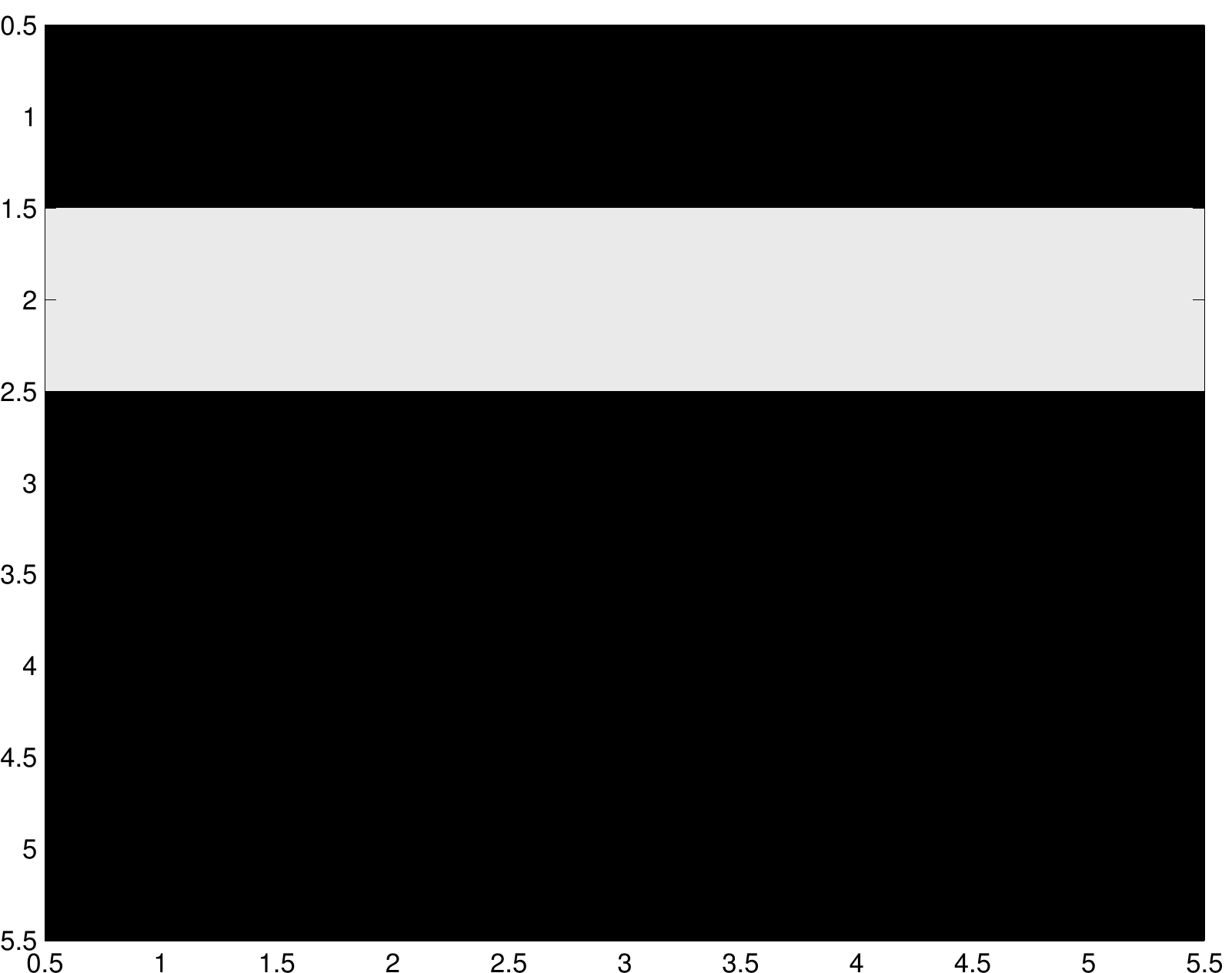}
} 
\centering
\subfigure[]{
\includegraphics[width=0.08\textwidth]{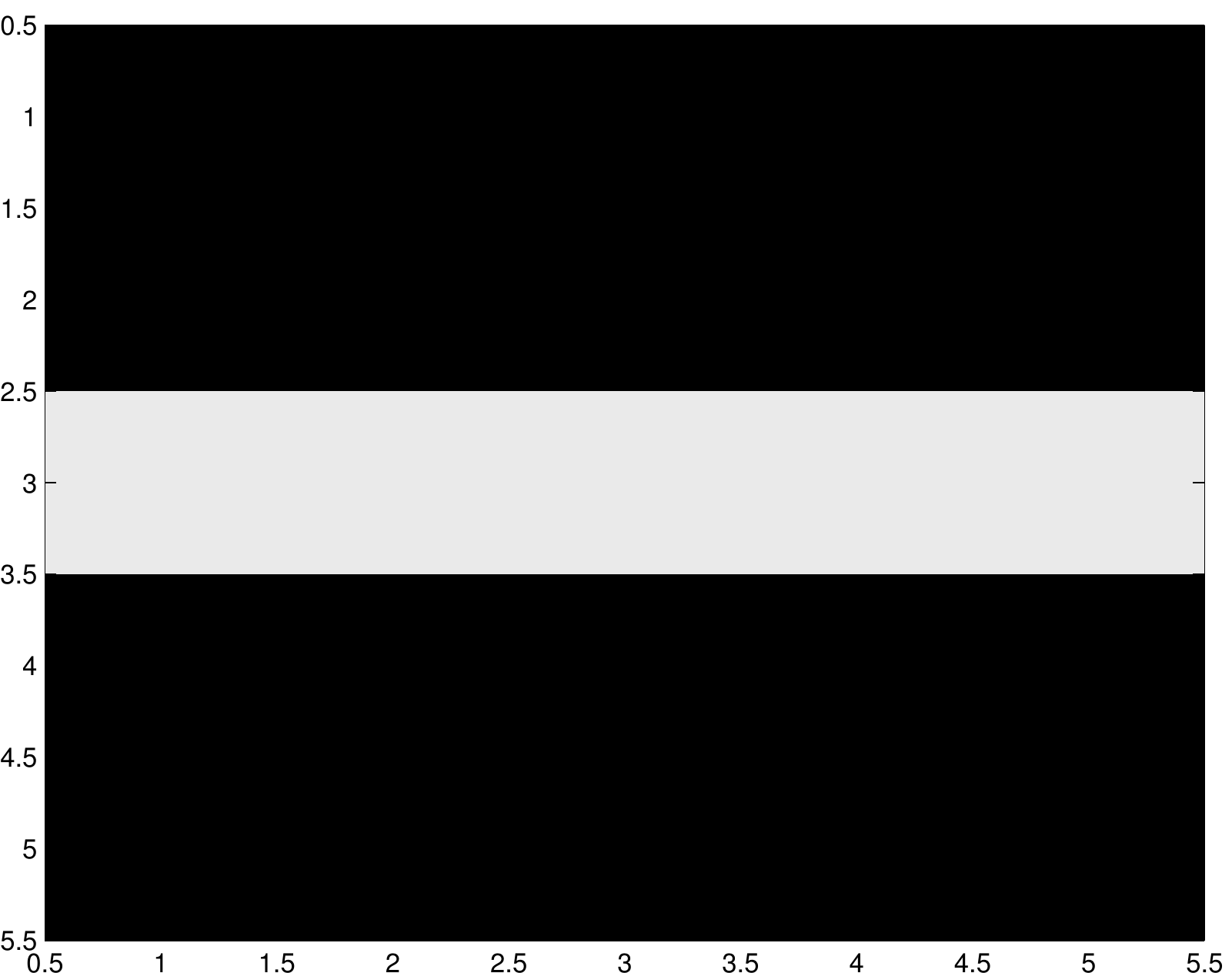}
}
\centering
\subfigure[]{
\includegraphics[width=0.08\textwidth]{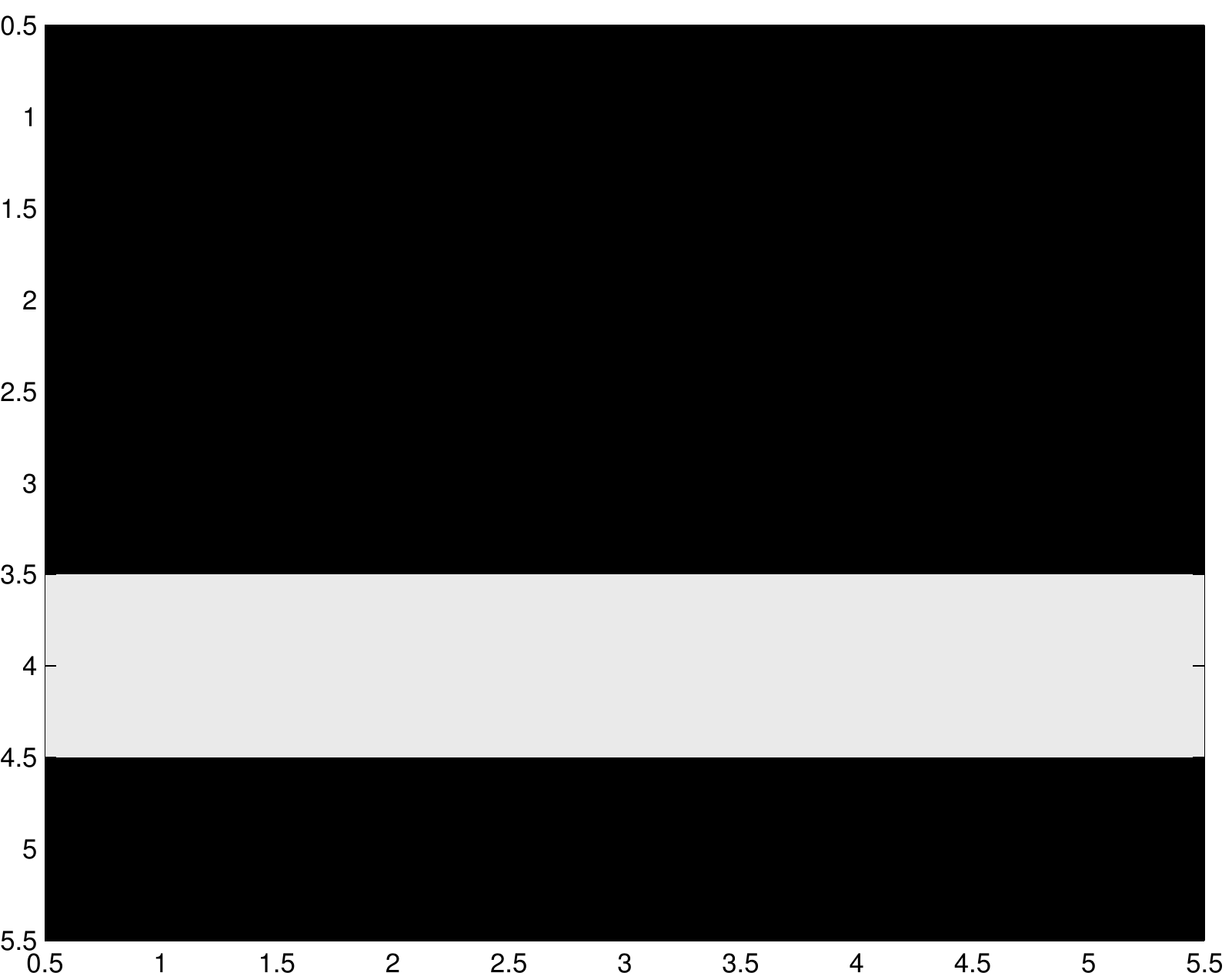}
} 
\centering
\subfigure[]{
\includegraphics[width=0.08\textwidth]{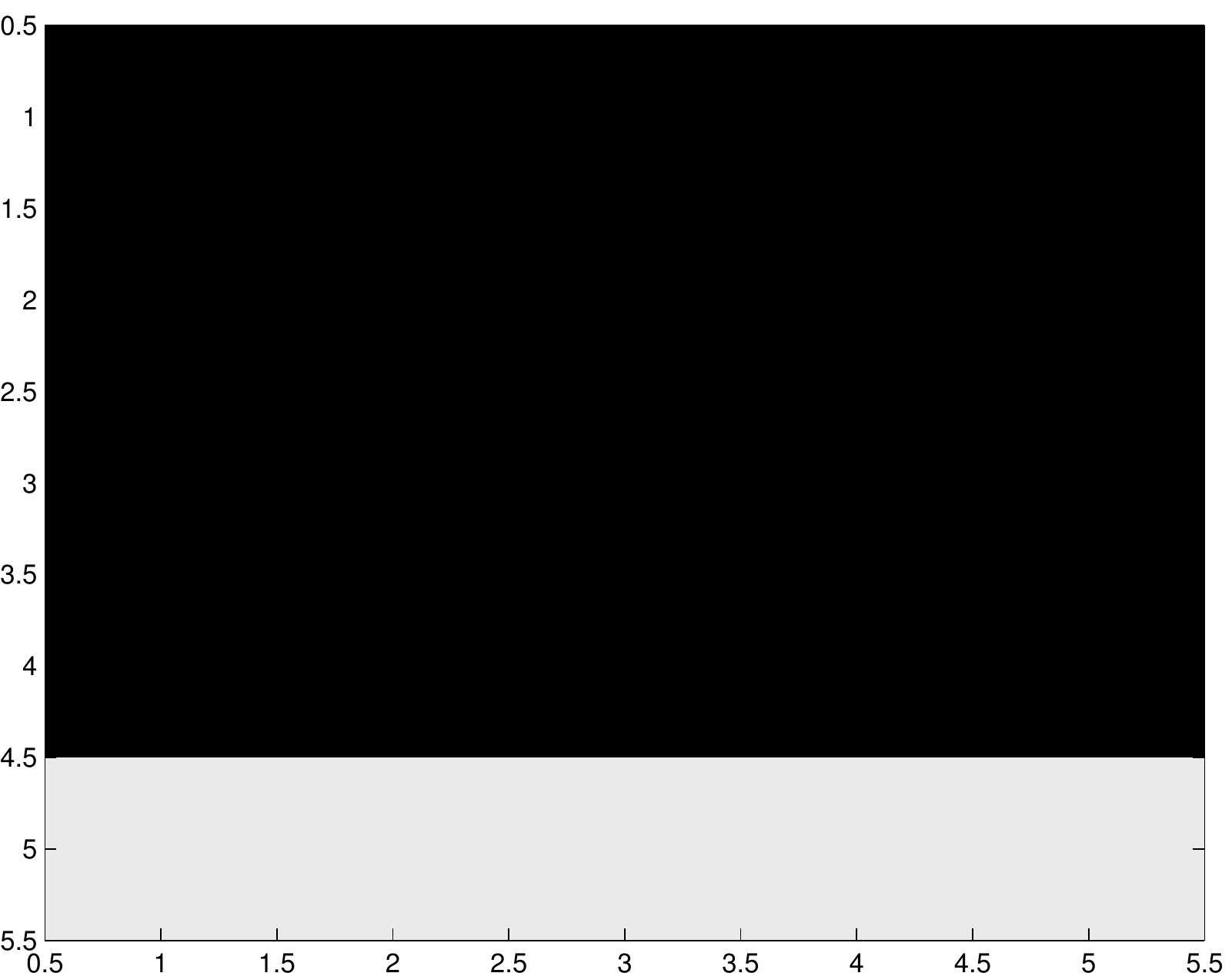}
} 
\caption{\label{fig:hor_vert_images}(a...e): Class 1 images. (f...j): Class -1 images.}
\end{figure}

We introduce in this section a running example used throughout the paper to illustrate the notion of adversarial robustness, and highlight its difference with the notion of risk.
We consider a binary classification task on square images of size $\sqrt{d} \times \sqrt{d}$. 
Images of class $1$ (resp. class $-1$) contain exactly one vertical line (resp. horizontal line), and a small constant positive number $a$ (resp. negative number $-a$) is added to all the pixels of the images. That is, for class $1$ (resp. $-1$) images, background pixels are set to $a$ (resp. $-a$), and pixels belonging to the line are equal to $1+a$ (resp. $1-a$). Fig. \ref{fig:hor_vert_images} illustrates the classification problem for $d = 25$. The number of datapoints to classify is $N = 2 \sqrt{d}$. 

Clearly, the most relevant concept (in terms of visual appearance) that permits to separate the two classes is the \textit{orientation} of the line (i.e., horizontal vs. vertical). 
The \textit{bias} of the image (i.e., the sum of all its pixels) is also a valid concept for this task, as it separates the two classes, despite being much more difficult to detect visually. The class of an image can therefore be correctly estimated from its orientation \textit{or} from the bias.
Let us first consider the linear classifier defined by
\begin{align}
\label{eq:f_lin}
f_{\text{lin}} (x) = \frac{1}{\sqrt{d}} \mathbf{1}^T x - 1,
\end{align}
where $\mathbf{1}$ is the vector of size $d$ whose entries are all equal to $1$, and $x$ is the vectorized image, exploits the difference of bias between the two classes and achieves a perfect classification accuracy for all $a > 0$.
Indeed, a simple computation gives $f_{\text{lin}} (x) = \sqrt{d} a$ (resp. $f_{\text{lin}} (x) = -\sqrt{d} a$) for class $1$ (resp. class $-1$) images. 
Therefore, the risk of $f_{\text{lin}}$ is $R(f_{\text{lin}}) = 0$. It is important to note that $f_{\text{lin}}$ only achieves zero risk because it captures the bias, but fails to distinguish between the images based on the orientation of the line. Indeed, when $a = 0$, the datapoints are not linearly separable. Despite its perfect accuracy for any $a > 0$, $f_{\text{lin}}$ is \textit{not} robust to small adversarial perturbations when $a$ is small, as a minor perturbation of the bias switches the estimated label. Indeed, a simple computation gives $\rho_{\text{adv}} (f_{\text{lin}}) = \sqrt{d} a$; therefore, the adversarial robustness of $f_{\text{lin}}$ can be made arbitrarily small by choosing $a$ to be small enough. 
More than that, among all linear classifiers that satisfy $R(f) = 0$, $f_{\text{lin}}$ is the one that maximizes $\rho_{\text{adv}} (f)$ (as we show later in Section \ref{sec:lin_classifiers}). Therefore, \textit{all} zero-risk linear classifiers are not robust to adversarial perturbations, for this classification task.

\begin{figure}[t]
\centering
\subfigure[Original]{
\includegraphics[width=0.08\textwidth]{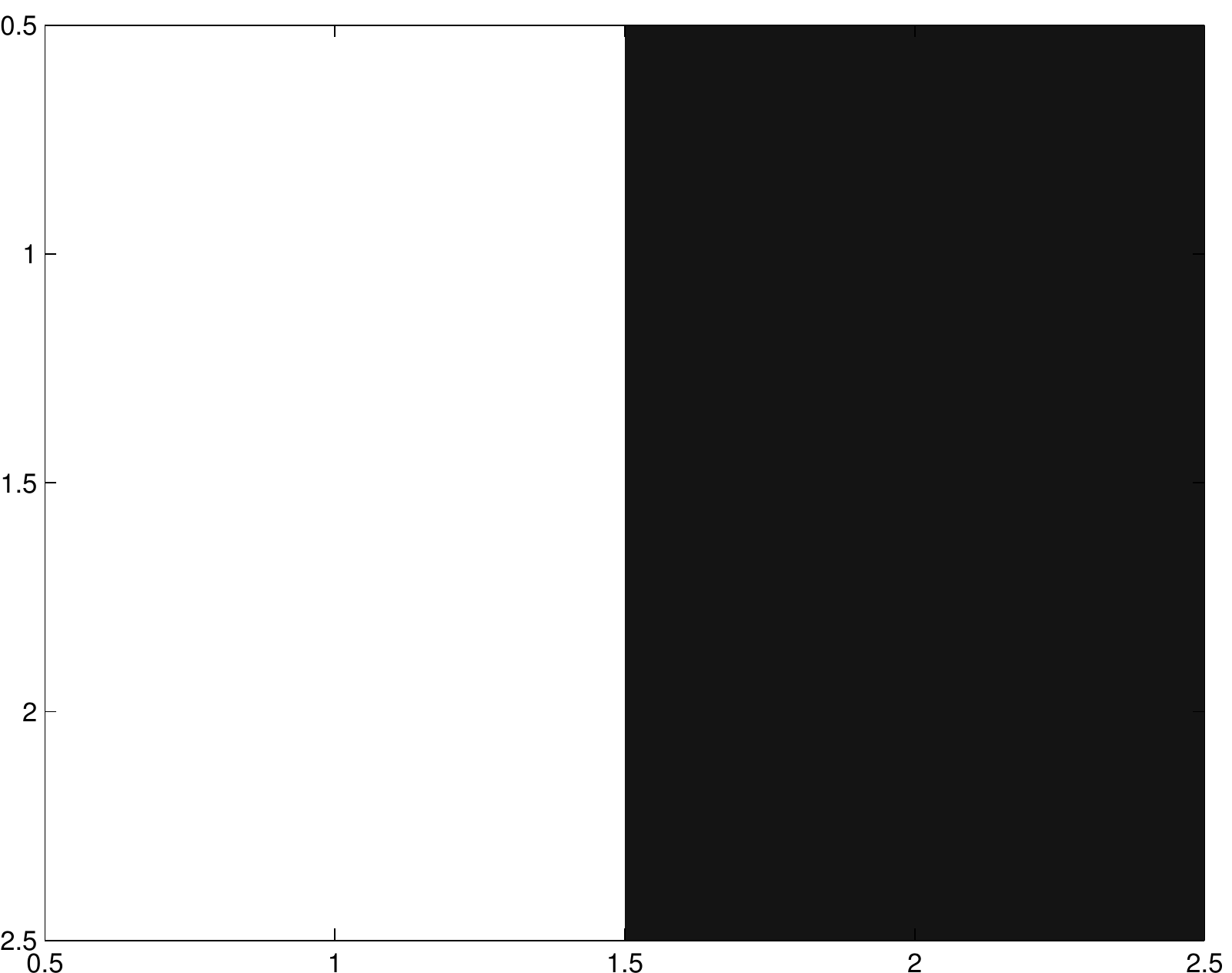}
}
\centering
\subfigure[$f_{\text{lin}}$]{
\includegraphics[width=0.08\textwidth]{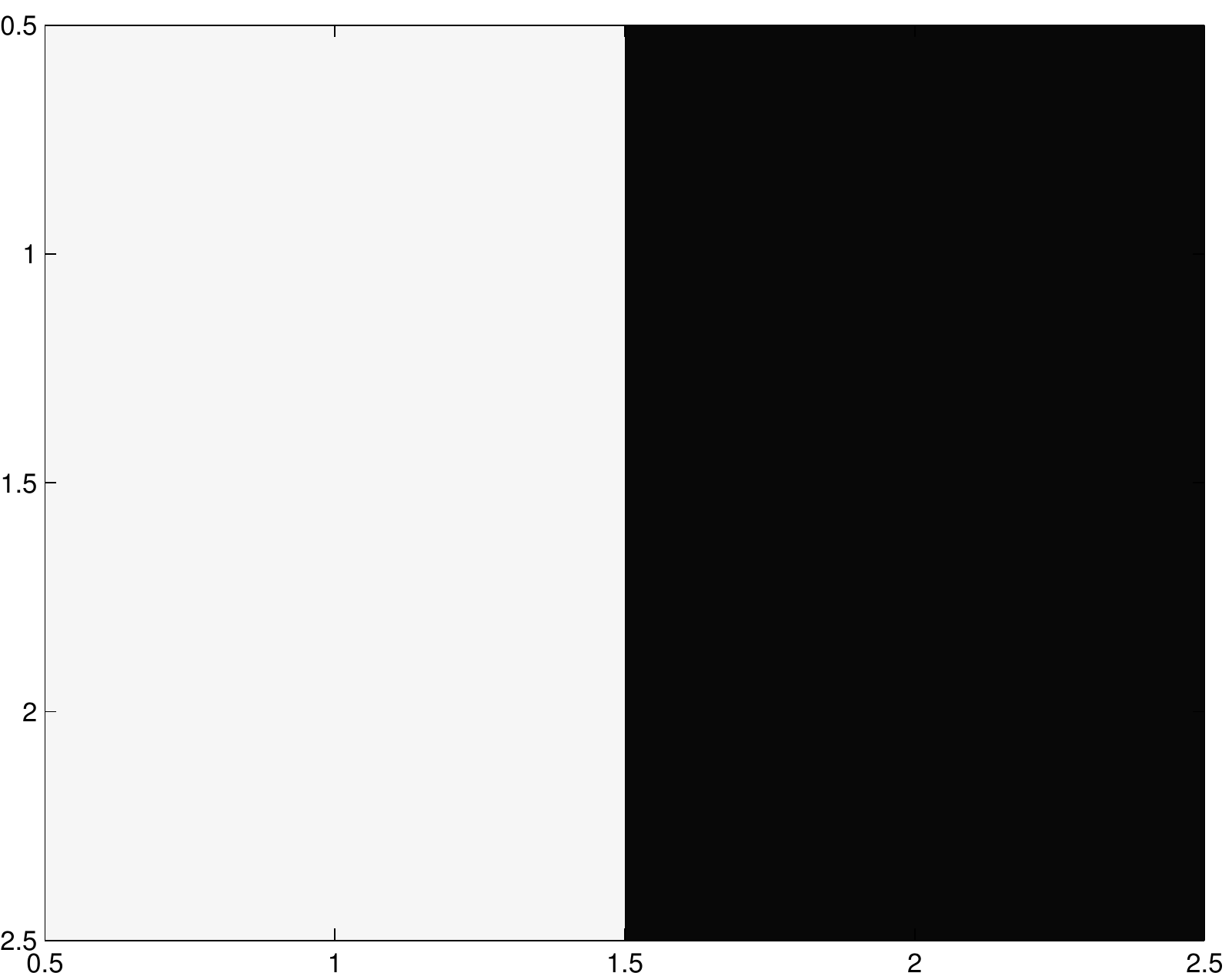}
}
\centering
\subfigure[$f_{\text{quad}}$]{
\includegraphics[width=0.08\textwidth]{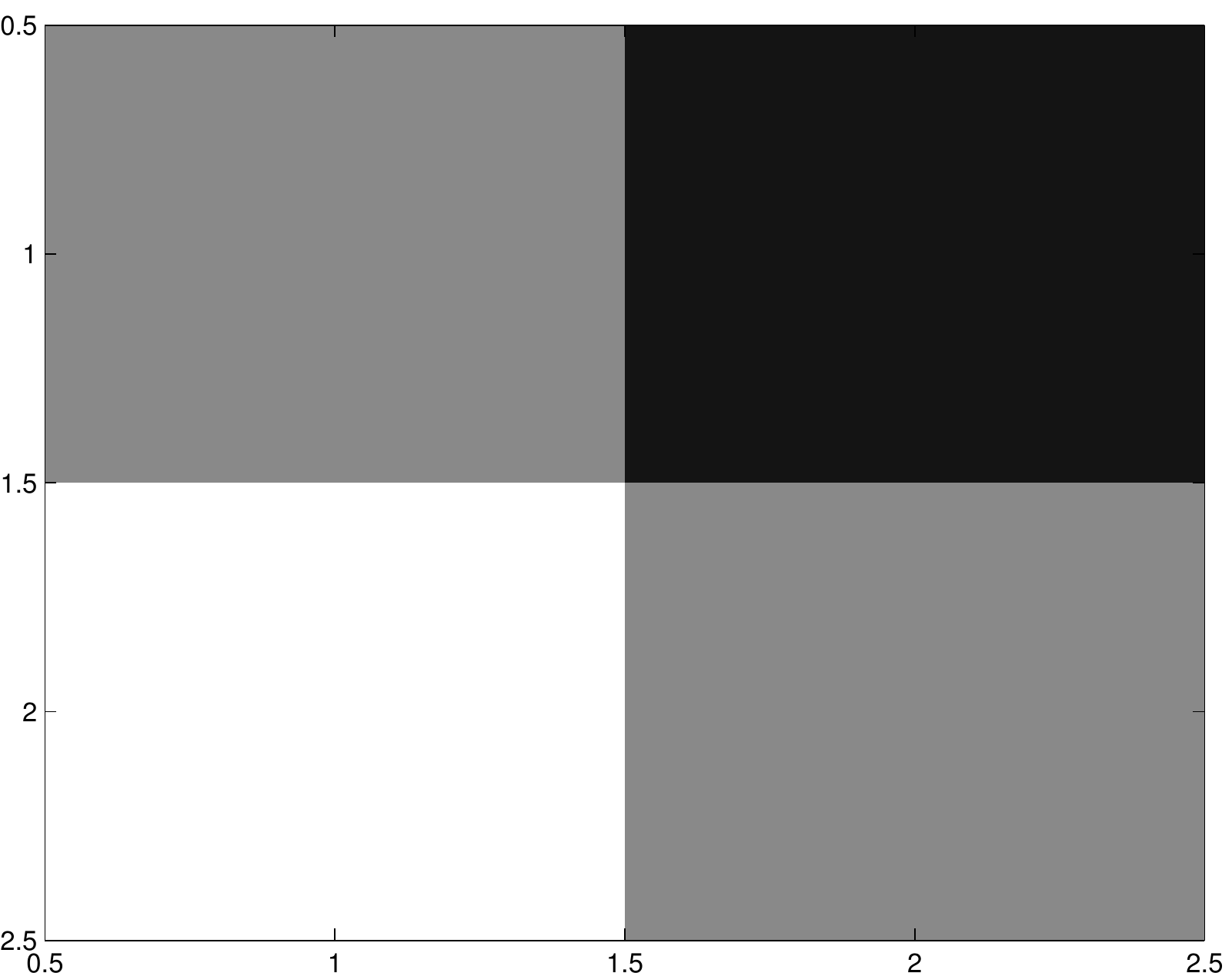}
}
\caption{\label{fig:vertical_example_2x2_2} Robustness to adversarial noise of linear and quadratic classifiers. (a): Original image ($d = 4$, and $a = 0.1/\sqrt{d}$), (b,c): Minimally perturbed image that switches the estimated label of (b) $f_{\text{lin}}$, (c) $f_{\text{quad}}$. Note that the difference between (b) and (a) is hardly perceptible, this demonstrates that $f_{\text{lin}}$ is not robust to adversarial noise. On the other hand images (c) and (a) are clearly different, which indicates that $f_{\text{quad}}$ is more robust to adversarial noise}
\end{figure}

Unlike linear classifiers, a more \textit{flexible} classifier that correctly captures the orientation of the lines in the images will be robust to adversarial perturbation, unless this perturbation significantly alters the image and modifies the direction of the line. 
To illustrate this point, we compare the adversarial robustness of $f_{\text{lin}}$ to that of a second order polynomial classifier $f_{\text{quad}}$ that achieves zero risk in Fig. \ref{fig:vertical_example_2x2_2}, for $d = 4$.\footnote{We postpone the detailed analysis of $f_{\text{quad}}$ to Section \ref{sec:quad_classifiers}.}
While a hardly perceptible change of the image is sufficient to switch the estimated label for the linear classifier, the minimal perturbation for $f_{\text{quad}}$ is one that modifies the direction of the line, to a great extent.

The above example highlights several important facts, which are summarized as follows:
\begin{itemize}
\item \textbf{Risk and adversarial robustness are two distinct properties of a classifier.} While $R(f_{\text{lin}}) = 0$, $f_{\text{lin}}$ is definitely not robust to small adversarial perturbations.\footnote{The opposite is also possible, since a constant classifier (e.g., $f(x) = 1$ for all $x$) is clearly robust to perturbations, but does not achieve good accuracy.} This is due to the fact that $f_{\text{lin}}$ only captures the bias in the images and ignores the orientation of the line. 
\item \textbf{To capture orientation (i.e., the most visual concept), one has to use a classifier that is flexible enough for the task.} Unlike the class of linear classifiers, the class of polynomial classifiers of degree $2$ correctly captures the line orientation, for $d = 4$. 
\item \textbf{The robustness to adversarial perturbations provides a quantitative measure of the strength of a concept.}  Since $\rho_{\text{adv}}(f_{\text{lin}}) \ll \rho_{\text{adv}}(f_{\text{quad}})$, one can confidently say that the concept captured by $f_{\text{quad}}$ is \textit{stronger} than that of $f_{\text{lin}}$, in the sense that the essence of the classification task is captured by $f_{\text{quad}}$, but not by $f_{\text{lin}}$ (while they are equal in terms of misclassification rate). In general classification problems, the quantity $\rho_{\text{adv}} (f)$ provides a natural way to evaluate and compare the learned concept; larger values of $\rho_{\text{adv}} (f)$ indicate that stronger concepts are learned, for comparable values of the risk.
\end{itemize}

As illustrated in the above toy example, the robustness to adversarial perturbations is key to assess the strength of a concept. In real-world classification tasks, weak concepts correspond to partial information about the classification task (which are possibly sufficient to achieve a good accuracy), while strong concepts capture the essence of the classification task.

\section{Upper limit on the adversarial robustness}
\label{sec:upper_limit}

We now introduce our theoretical framework for analyzing the robustness to adversarial perturbations. We first present a key assumption on the classifier $f$ for the analysis of adversarial robustness.

\noindent \textbf{Assumption (A).} There exist $\tau > 0$ and $0 < \gamma \leq 1$ such that, for all $x \in \mathcal{B}$,
\begin{equation}
\begin{aligned}
\text{dist} (x, S_{-}) & \leq \tau \max(0, f(x))^\gamma, \\
\text{dist} (x, S_{+}) & \leq \tau \max(0, -f(x))^\gamma,
\end{aligned}
\label{eq:assumption_A}
\end{equation} 
where $\dist(x, S) = \min_{y} \{ \| x - y \|_2: y \in S \}$ and $S_+$ (resp. $S_{-}$) is the set of points $x$ such that $f(x) \geq 0$ (resp. $f(x) \leq 0$): 
\begin{align*}
S_+ & = \{ x: f(x) \geq 0\}, \\
S_- & = \{ x: f(x) \leq 0 \}.
\end{align*}

In words, the assumption (A) states that for any datapoint $x$, the \textit{residual} $\max(0, f(x))$ (resp. $\max(0, -f(x))$) can be used to bound the distance from $x$ to a datapoint $y$ classified $-1$ (resp. $1$). 

Bounds of the form Eq. (\ref{eq:assumption_A}) have been established for various classes of functions since the early of work of Lojasiewicz \citep{lojasiewicz1961probleme} in algebraic geometry and have found applications in areas such as mathematical optimization \citep{pang1997error, lewis1998error}. For example, Lojasiewicz \citep{lojasiewicz1961probleme} and later \citep{luo1994error} have shown that, quite remarkably, \textit{assumption (A) holds for the general class of analytic functions}. In \citep{ng2003error}, (A) is shown to hold with $\gamma=1$ for piecewise linear functions. In \citep{luo1994extension}, error bounds on polynomial systems are studied. Proving inequality \eqref{eq:assumption_A} with explicit constants $\tau$ and $\gamma$ for different classes of functions is still an active area of research \citep{li2014new}. In Sections \ref{sec:lin_classifiers} and \ref{sec:quad_classifiers}, we provide examples of function classes for which (A) holds, and explicit formulas for the parameters $\tau$ and $\gamma$. 

The following result establishes a general upper bound on the robustness to adversarial perturbations:
\begin{lemma}
\label{lemma:main_result}
Let $f$ be an arbitrary classifier that satisfies (A) with parameters $(\tau, \gamma)$. Then,
\begin{align*}
\rho_{\text{adv}} (f) \leq 4^{1-\gamma} \tau \left( p_1 \mathbb{E}_{\mu_1}  (f(x))  - p_{-1} \mathbb{E}_{\mu_{-1}} (f(x)) + 2 \| f \|_{\infty} R(f) \right)^{\gamma}.
\end{align*}
\end{lemma}
The proof can be found in Appendix \ref{sec:proof_lemma_main}.
The above result provides an upper bound on the adversarial robustness that depends on the \textit{risk} of the classifier, as well as a measure of the separation between the \textit{expectations} of the classifier values computed on distribution $\mu_1$ and $\mu_{-1}$.
This result is general, as we only assume that $f$ satisfies assumption $(A)$. In the next two sections, we apply Lemma \ref{lemma:main_result} to two classes of classifiers, and derive interpretable upper bounds in terms of a distinguishibality measure (that depends only on the dataset) which quantifies the notion of difficulty of a classification task. Studying the general result in Lemma \ref{lemma:main_result} through two practical classes of classifiers shows the implications of such a fundamental limit on the adversarial robustness, and illustrates the methodology for deriving class-specific and practical upper bounds on adversarial robustness from the general upper bound.




\section{Robustness of linear classifiers to adversarial and random perturbations}
\label{sec:lin_classifiers}

The goal of this section is two-fold; first, we specialize Lemma \ref{lemma:main_result} to the class of linear functions, and derive interpretable upper bounds on the robustness of classifiers to adversarial perturbations (Section \ref{sec:lin_adv}). Then, we derive a formal relation between the robustness of linear classifiers to adversarial robustness, and the robustness to random uniform noise (Section \ref{sec:random_uniform_noise}).


\subsection{Adversarial perturbations}
\label{sec:lin_adv}
We define the classification function $f(x) = w^T x + b$. 
Note that any linear classifier for which $|b| > M \| w \|_2$ is a trivial classifier that assigns the same label to all points, and we therefore assume that $|b| \leq M \| w \|_2$. 

We first show that the family of linear classifiers satisfies assumption (A), with explicit parameters $\tau$ and $\gamma$. 
\begin{lemma}
\label{lemma:tau_gamma_linear}
Assumption (A) holds for linear classifiers $f(x) = w^T x + b$ with $\tau = 1/\| w \|_2$ and $\gamma = 1$.
\end{lemma}
\begin{proof}
Let $x$ be such that $f(x) \geq 0$, and the goal is to prove that $\text{dist}(x, S_{-}) \leq \tau f(x)^{\gamma}$ (the other inequality can be handled in a similar way). We have $f(x) = w^T x + b$. $\text{dist}(x, S_-) = f(x) / \| w \|_2 \implies \tau = 1/\|w\|_2, \gamma=1$.
\end{proof}
Using Lemma \ref{lemma:main_result}, we now derive an interpretable upper bound on the robustness to adversarial perturbations. In particular, the following theorem bounds $\rho_{\text{adv}} (f)$ from above in terms of the first moments of the distributions $\mu_1$ and $\mu_{-1}$, and the classifier's risk:
\begin{theorem}
Let $f(x) = w^T x + b$ such that $|b| \leq M \| w \|_2$. Then,
\begin{align}
\label{eq:first_result_linear}
\rho_{\text{adv}} (f) \leq \| p_1 \bb{E}_{\mu_1} (x) - p_{-1} \bb{E}_{\mu_{-1}} (x) \|_2 + M (|p_1 - p_{-1}| + 4 R(f)).
\end{align}
In the balanced setting where $p_{1} = p_{-1} = 1/2$, and if the intercept $b = 0$ the following inequality holds:
\begin{align}
\label{eq:second_result_linear}
\rho_{\text{adv}} (f) \leq \frac{1}{2} \| \bb{E}_{\mu_1} (x) - \bb{E}_{\mu_{-1}} (x) \|_2 + 2 M R(f).
\end{align}
\label{th:linear_expectation}
\end{theorem}
\begin{small}
\begin{proof}
Using Lemma \ref{lemma:main_result} with $\tau = 1/\|w\|_2$ and $\gamma = 1$, we have
\begin{align}
\label{eq:linear_upper_bound}
\rho_{\text{adv}} (f) \leq \frac{1}{\| w \|_2} \left( w^T \left( p_1 \mathbb{E}_{\mu_1} (x) - p_{-1} \mathbb{E}_{\mu_{-1}} (x) \right) + b (p_1 - p_{-1}) + 2 \| f \|_{\infty} R(f) \right)
\end{align}
Observe that
\begin{itemize}[topsep=0pt,itemsep=-1ex,partopsep=1ex,parsep=1ex]
\item[i.]  $w^T \left( p_1 \mathbb{E}_{\mu_1} (x) - p_{-1} \mathbb{E}_{\mu_{-1}} (x) \right) \leq \| w \|_2 \| p_1 \mathbb{E}_{\mu_1} (x) - p_{-1} \mathbb{E}_{\mu_{-1}} (x) \|_2$ using Cauchy-Schwarz inequality.
\item[ii.] $b (p_1 - p_{-1}) \leq M \| w \|_2 |p_{1} - p_{-1}|$ using the assumption $|b| \leq M \|w\|_2$,
\item[iii.] $\| f \|_{\infty} = \max_{x: \| x \|_2 \leq M} \{ |w^T x + b| \} \leq 2 M \| w \|_2$.
\end{itemize}
By plugging the three inequalities in Eq. (\ref{eq:linear_upper_bound}), we obtain the desired result in Eq. (\ref{eq:first_result_linear}).

When $p_1 = p_{-1} = 1/2$, and the intercept $b=0$, inequality (iii) can be tightened to $\| f \|_{\infty} \leq M \| w \|_2$, and directly leads to the stated result Eq. (\ref{eq:second_result_linear}). 
\end{proof}
\end{small}

\begin{figure}[t]
\centering
\subfigure[]{
\includegraphics[width=0.4\textwidth]{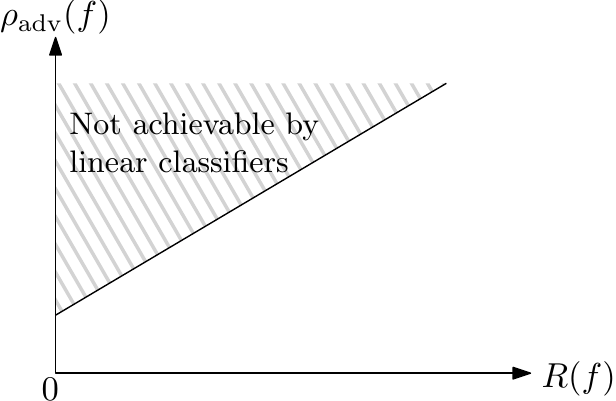}
}
\subfigure[]{
\includegraphics[width=0.3\textwidth]{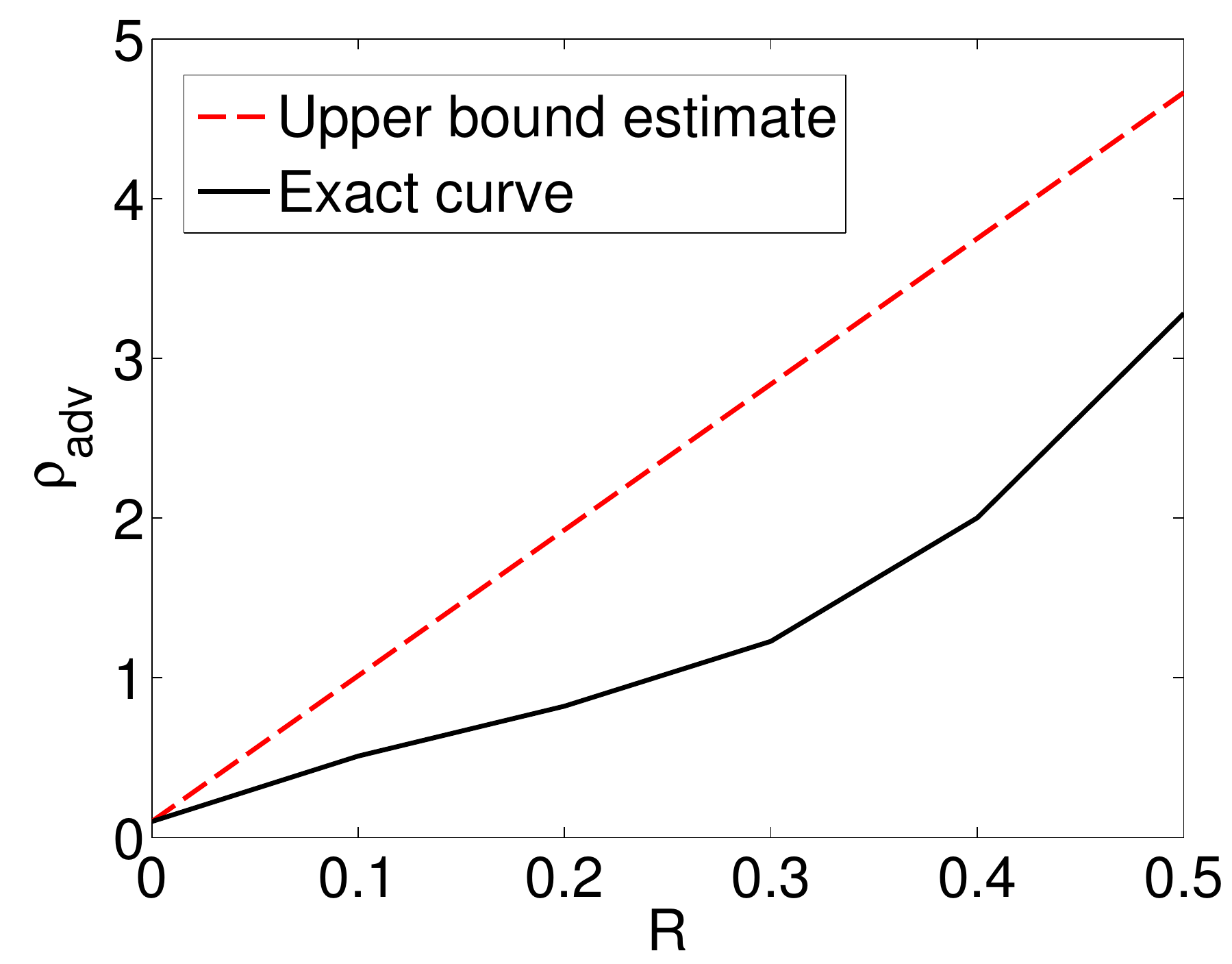}
}
\caption{\label{fig:curve_rhoR_mixed} Adversarial robustness $\rho_{\text{adv}}$ versus risk diagram for linear classifiers. Each point in the plane represents a linear classifier $f$. (a): Illustrative diagram, with the non-achievable zone (Theorem \ref{th:linear_expectation}). (b): The exact $\rho_{\text{adv}}$ versus risk achievable curve, and our upper bound estimate on the running example.}
\end{figure}

Our upper bound on $\rho_{\text{adv}} (f)$ depends on the difference of means $\| \bb{E}_{\mu_1} (x) - \bb{E}_{\mu_{-1}} (x) \|_2$, which measures the distinguishability between the classes. Note that this term is classifier-independent, and is only a property of the classification task. The only dependence on $f$ in the upper bound is through the risk $R(f)$.
%
Thus, in classification tasks where the means of the two distributions are close (i.e., $\| \bb{E}_{\mu_1} (x) - \bb{E}_{\mu_{-1}} (x) \|_2$ is small), \textit{any linear classifier} with small risk will necessarily have a small robustness to adversarial perturbations. Note that the upper bound logically increases with the risk, as there clearly exist robust linear classifiers that achieve high risk (e.g., constant classifier). Fig. \ref{fig:curve_rhoR_mixed} (a) pictorially represents the $\rho_{\text{adv}}$ vs $R$ diagram as predicted by Theorem \ref{th:linear_expectation}. Each linear classifier is represented by a point on the $\rho_{\text{adv}}$--$R$ diagram, and our result shows the existence of a region that linear classifiers cannot attain. 

Quite importantly, in many interesting classification problems, the quantity $\| \bb{E}_{\mu_1} (x) - \bb{E}_{\mu_{-1}} (x) \|_2$ is small due to large intra-class variability (e.g., due to complex intra-class geometric transformations in computer vision applications). Therefore, even if a linear classifier can achieve a good classification performance on such a task, it will not be robust to small adversarial perturbations. In simple tasks involving distributions with significantly different averages, it is likely that there exists a linear classifier that can separate correctly the classes, and have a large robustness to adversarial perturbations.

\subsection{Random uniform noise}
\label{sec:random_uniform_noise}
We now examine the robustness of linear classifiers to random uniform noise. The following theorem compares the robustness of linear classifiers to random uniform noise with their robustness to adversarial perturbations. 
\begin{theorem}
Let $f(x) = w^T x + b$. 
For any $\epsilon \in [0, 1/12)$, we have the following bounds on $\rho_{\text{unif}, \epsilon} (f)$:
\begin{align}
\rho_{\text{unif}, \epsilon} (f) & \geq \max\left(C_1(\epsilon) \sqrt{d}, 1\right) \rho_{\text{adv}} (f), \label{eq:unif_1} \\
\rho_{\text{unif}, \epsilon} (f) & \leq \widetilde{C_2}(\epsilon,d) \rho_{\text{adv}} (f) \leq C_2(\epsilon) \sqrt{d} \rho_{\text{adv}} (f) \label{eq:unif_2},
\end{align}
with $C_1(\epsilon) = (2 \ln(2/\epsilon))^{-1/2}$, 
$\widetilde{C_2}(\epsilon,d) = (1 - (12 \epsilon)^{1/d})^{-1/2}$ and
$C_2(\epsilon) = (1 - 12 \epsilon)^{-1/2}$.
\label{th:uniform_noise_linear}
\end{theorem}
The proof can be found in appendix \ref{sec:proof_theorm_proba}. In words, $\rho_{\text{unif}, \epsilon} (f)$ behaves as $\sqrt{d} \rho_{\text{adv}} (f)$ for linear classifiers (for constant $\epsilon$). Linear classifiers are therefore more robust to random noise than adversarial perturbations, by a factor of $\sqrt{d}$. In typical high dimensional classification problems, this shows that a linear classifier can be robust to random noise even if $\| \bb{E}_{\mu_1} (x) - \bb{E}_{\mu_{-1}} (x) \|_2$ is small. 
 Note moreover that our result is tight for $\epsilon = 0$, as we get $\rho_{\text{unif}, 0} (f)= \rho_{\text{adv}} (f)$.

Our results can be put in perspective with the empirical results of \cite{szegedy2013intriguing}, that showed a large gap between the two notions of robustness on neural networks. Our analysis provides a confirmation of this high dimensional phenomenon on linear classifiers.

%


\subsection{Illustration of the results on the running example}

We now illustrate our theoretical results on the example of Section \ref{sec:horizontal_vertical}. In this case, we have $\| \bb{E}_{\mu_1} (x) - \bb{E}_{\mu_{-1}} (x) \|_2 = 2 \sqrt{d} a$.
By using Theorem \ref{th:linear_expectation}, \textit{any} zero-risk linear classifier satisfies $\rho_{\text{adv}} (f) \leq \sqrt{d} a$. As we choose $a \ll 1/\sqrt{d}$, accurate linear classifiers are therefore not robust to adversarial perturbations for this task. 
We note that $f_{\text{lin}}$ (defined in Eq.(\ref{eq:f_lin})) achieves the upper bound and is therefore the most robust accurate linear classifier one can get, as it can easily be checked that $\rho_{\text{adv}} (f_{\text{lin}}) = \sqrt{d} a$. In Fig. \ref{fig:curve_rhoR_mixed} (b) the exact $\rho_{\text{adv}}$ vs $R$ curve is compared to our theoretical upper bound\footnote{The exact curve is computed using a bruteforce approach.}, for $d = 25$, $N = 10$ and a bias $a = 0.1/\sqrt{d}$. Besides the zero-risk case where our upper bound is tight, the upper bound is reasonably close to the exact curve for other values of the risk (despite not being tight).

\begin{figure}[t]
\centering
\includegraphics[scale=0.3]{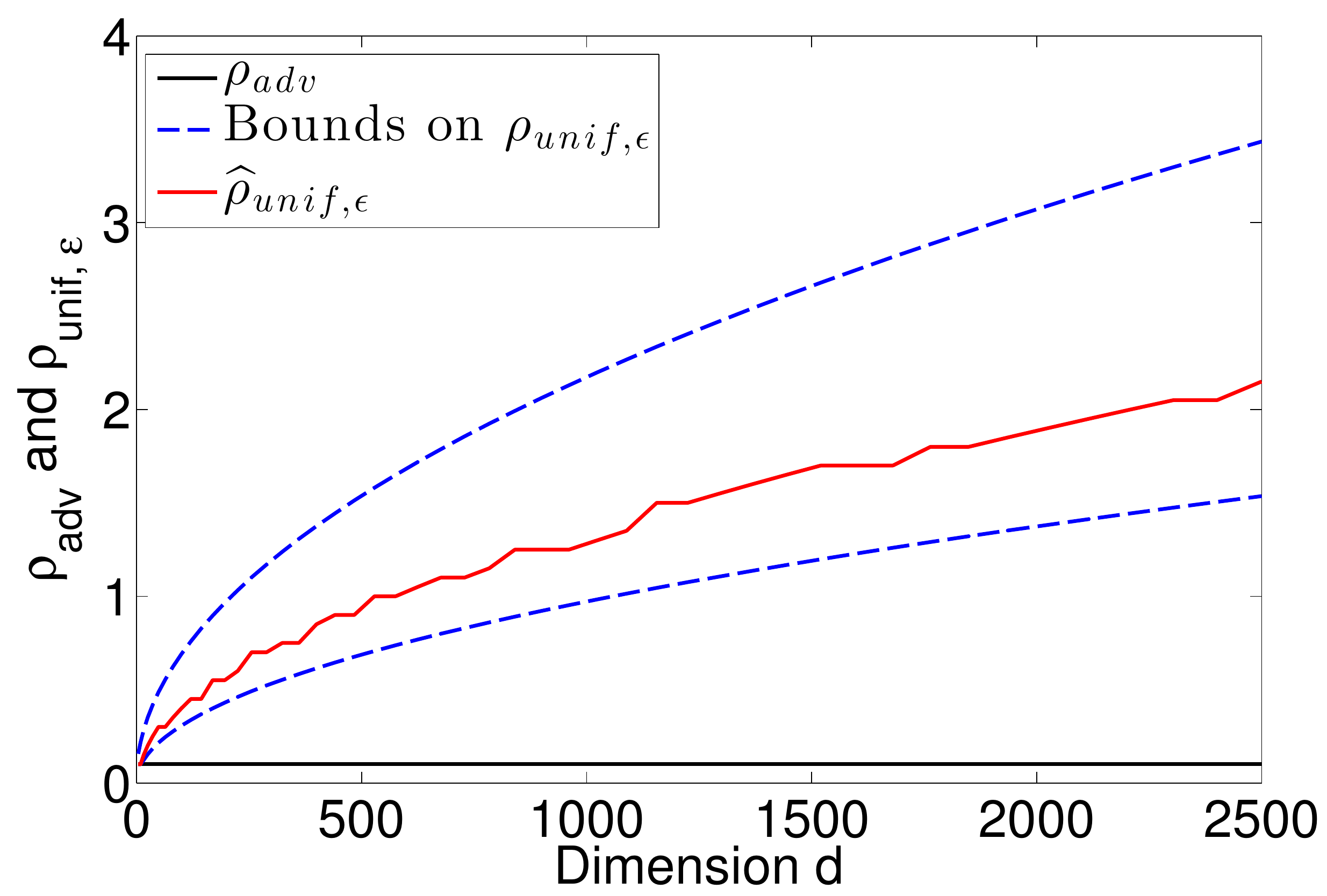}
\caption{\label{fig:uniform_random_noise_vert_horiz} Adversarial robustness and robustness to random uniform noise of $f_{\text{lin}}$ versus the dimension $d$. We used $\epsilon = 0.01$, and $a = 0.1/\sqrt{d}$. The lower bound is given in Eq. (\ref{eq:unif_1}), and the upper bound is the first inequality in Eq. (\ref{eq:unif_2}).}
\end{figure}

We now focus on the robustness to uniform random noise of $f_{\text{lin}}$. For various values of $d$, we compute the upper and lower bounds on the robustness to random uniform noise (Theorem \ref{th:uniform_noise_linear}) of $f_{\text{lin}}$, where we fix $\epsilon$ to $0.01$. In addition, we compute a simple empirical estimate $\widehat{\rho}_{\text{unif}, \epsilon}$ of the robustness to random uniform noise of $f_{\text{lin}}$ (see Sec. \ref{sec:exp_results} for details on the computation of this estimate). The results are illustrated in Fig. \ref{fig:uniform_random_noise_vert_horiz}. While the adversarial noise robustness is constant with the dimension (equal to $0.1$, as $\rho_{\text{adv}} (f_{\text{lin}}) = \sqrt{d} a$ and $a = 0.1/\sqrt{d}$), the robustness to random uniform noise \textit{increases} with $d$. For example, for $d = 2500$, the value of $\rho_{\text{unif}, \epsilon}$ is at least $15$ times larger than the adversarial robustness $\rho_{\text{adv}}$. In high dimensions, a linear classifier is therefore much more robust to random uniform noise than adversarial noise. 

\section{Adversarial robustness of quadratic classifiers}
\label{sec:quad_classifiers}

In this section, we derive specialized upper bounds on the robustness to adversarial perturbations of quadratic classifers using Lemma \ref{lemma:main_result}. 

\subsection{Analysis of adversarial perturbations}

We study the robustness to adversarial perturbations of quadratic classifiers of the form $f(x) = x^T A x$, where $A$ is a symmetric matrix. Besides the practical use of quadratic classifiers in some applications \citep{goldberg2008splitsvm, chang2010training}, they represent a natural extension of linear classifiers. The study of linear vs. quadratic classifiers provides insights into how adversarial robustness depends on the family of considered classifiers.
Similarly to the linear setting, we exclude the case where $f$ is a trivial classifier that assigns a constant label to all datapoints. That is, we assume that $A$ satisfies
\begin{align}
\label{eq:lambda_min_max}
\lambda_{\min} (A) < 0, \quad \lambda_{\max} (A) > 0,
\end{align}
where $\lambda_{\min} (A)$ and $\lambda_{\max} (A)$ are the smallest and largest eigenvalues of $A$. We moreover impose that the eigenvalues of $A$ satisfy 
\begin{align}
\label{eq:condition_K}
\max\left( \left| \frac{\lambda_{\min} (A)}{\lambda_{\max} (A)} \right|,  \left| \frac{\lambda_{\max} (A)}{\lambda_{\min} (A)} \right| \right) \leq K,
\end{align}
for some constant value $K \geq 1$. This assumption imposes an approximate symmetry around $0$ of the extremal eigenvalues of $A$, thereby disallowing a large bias towards any of the two classes. 

We first show that the assumption (A) is satisfied for quadratic classifiers, and derive explicit formulas for $\tau$ and $\gamma$.
\begin{lemma}
\label{lemma:tau_gamma_quadratic}
Assumption (A) holds for the class of quadratic classifiers $f(x) = x^T A x$ where $\lambda_{\min} (A) < 0$, $\lambda_{\max} (A) > 0$ with $\tau = \max(|\lambda_{\min} (A)|^{-1/2}, |\lambda_{\max} (A)|^{-1/2})$, and $\gamma = 1/2$,
\end{lemma}
\begin{proof}
Let $x$ be such that $f(x) \geq 0$, and the goal is to prove that $\text{dist}(x, S_{-}) \leq \tau f(x)^{\gamma}$ (the other inequality can be handled in a similar way). Assume without loss of generality that $A$ is diagonal (this can be done using an appropriate change of basis). 
Let $\nu = -\lambda_{\min} (A)$.
We have $f(x) = \sum_{i=1}^{d-1} \lambda_i x_i^2 - \nu x_d^2$. By setting $r_i = 0$ for all $i \in \{1, \dots, d-1\}$ and $r_d = \text{sign} (x_d) \sqrt{f(x)/\nu}$, (where $\text{sign}(x) = 1$ if $x \geq 0$ and $-1$ otherwise) we have
\begin{align*}
f(x+r) & = \sum_{i=1}^{d-1} \lambda_i x_i^2 - \nu (x_d + \text{sgn} (x_d) \sqrt{f(x) / \nu} )^2 \\
		   & = f(x) - 2 \nu x_d \text{sgn} (x_d) \sqrt{f(x) / \nu} - f(x) \\
		   & = - 2 \nu | x_d | \sqrt{f(x) / \nu} \leq 0.
\end{align*}
Hence, $\text{dist} (x, S_{-}) \leq \| r \|_2 = \nu^{-1/2} \sqrt{f(x)} \implies \tau = \nu^{-1/2}, \gamma=1/2$. 
\end{proof}
The following result builds on Lemma \ref{lemma:main_result} and bounds the adversarial robustness of quadratic classifiers as a function of the second order moments of the distribution and the risk.
\begin{theorem}
\label{th:quadratic_expectation}
Let $f(x) = x^T A x$, where $A$ satisfies Eqs. \eqref{eq:lambda_min_max} and \eqref{eq:condition_K}. Then, 
\begin{align*}
\rho_{\text{adv}} (f) \leq 2 \sqrt{K \| p_1 C_1 - p_{-1} C_{-1} \|_{*} + 2 M K R(f)},
\end{align*}
where $C_{\pm 1} (i,j) = (\bb{E}_{\mu_{\pm 1}} (x_i x_j))_{1 \leq i,j \leq d}$, and $\| \cdot \|_{*}$ denotes the nuclear norm defined as the sum of the singular values of the matrix.
\end{theorem}
\begin{small}
\begin{proof}
The class of classifiers under study satisfies assumption (A) with $\tau = \max(|\lambda_{\min} (A)|^{-1/2}, |\lambda_{\max} (A)|^{-1/2})$, and $\gamma = 1/2$ (see Lemma \ref{lemma:tau_gamma_quadratic}). By applying Lemma \ref{lemma:main_result}, we have
\begin{align*}
\rho_{\text{adv}} (f) \leq 2 \tau \left( \mathbb{E}_{\mu_1} (x^T A x) - \mathbb{E}_{\mu_{-1}} (x^T A x) + 2 \| f \|_{\infty} R(f) \right)^{1/2}.
\end{align*}
Observe that
\begin{itemize}[topsep=0pt,itemsep=-1ex,partopsep=1ex,parsep=1ex]
\item[i.] $p_1 \mathbb{E}_{\mu_1} (x^T A x) - p_{-1} \mathbb{E}_{\mu_{-1}} (x^T A x) = \langle A, p_1 C_1 - p_{-1} C_{-1} \rangle \leq \| A \| \| p_1 C_1 - p_{-1} C_{-1} \|_{*}$ using the generalized Cauchy-Schwarz inequality, where $\| \cdot \|$ and $\| \cdot \|_{*}$ denote respectively the spectral and nuclear matrix norms.
\item[ii.] $|f(x)| = |x^T A x| \leq \| A \| \| x \| \leq \| A \| M$,
\item[iii.] $\| A \|^{1/2} \tau = \max(|\lambda_{\min}(A)|, |\lambda_{\max}(A)|)^{1/2} \max(|\lambda_{\min} (A)|^{-1/2}, |\lambda_{\max} (A)|^{-1/2}) \leq \sqrt{K}$.
\end{itemize}
Applying these three inequalities, we obtain
\begin{align*}
\rho_{\text{adv}} (f) \leq 2 \| A \|^{1/2} \tau \left( \| p_1 C_1 - p_{-1} C_{-1} \|_{*} + 2 M R(f) \right)^{1/2} \leq 2 \sqrt{K} \left( \| p_1 C_1 - p_{-1} C_{-1} \|_{*} + 2 M R(f) \right)^{1/2}.
\end{align*}

\end{proof}
\end{small}

In words, the upper bound on the adversarial robustness depends on a distinguishability measure, defined by $\| C_1 - C_{-1} \|_{*}$, and the classifier's risk. In difficult classification tasks, where $\| C_1 - C_{-1} \|_{*}$ is small, any quadratic classifier with low risk that satisfies our assumptions in Eq. (\ref{eq:lambda_min_max}, \ref{eq:condition_K}) is not robust to adversarial perturbations. 

It should be noted that, while the distinguishability is measured with the distance between the means of the two distributions in the linear case, it is defined here as the difference between the second order moments matrices $\| C_1 - C_{-1} \|_{*}$. Therefore, in classification tasks involving two distributions with close means, and different second order moments, any zero-risk linear classifier will not be robust to adversarial noise, while zero-risk and robust quadratic classifiers are a priori possible according to our upper bound in Theorem \ref{th:quadratic_expectation}. This suggests that robustness to adversarial perturbations can be larger for more flexible classifiers, for comparable values of the risk. 



\subsection{Illustration of the results on the running example}
We now illustrate our results on the running example of Section \ref{sec:horizontal_vertical}, with $d = 4$. In this case, a simple computation gives $\| C_1 - C_{-1} \|_{*} = 2 + 8a \geq 2$. This term is significantly larger than the difference of means (equal to $4a$), and there is therefore hope to have a quadratic classifier that is accurate \textit{and} robust to small adversarial perturbations, according to Theorem \ref{th:quadratic_expectation}. In fact, the following quadratic classifier
\begin{align*}
f_{\text{quad}} (x) = x_1 x_2 + x_3 x_4 - x_1 x_3 - x_2 x_4,
\end{align*}
outputs $1$ for vertical images, and $-1$ for horizontal images (independently of the bias $a$). Therefore, $f_{\text{quad}}$ achieves zero risk on this classification task, similarly to $f_{\text{lin}}$. The two classifiers however have different robustness properties to adversarial perturbations. Using straightforward calculations, it can be shown that $\rho_{\text{adv}} (f_{\text{quad}}) = 1/\sqrt{2}$, for any value of $a$ (see Appendix \ref{sec:vertical_horizontal_details_quadratic} for more details). For small values of $a$, we therefore get $\rho_{\text{adv}} (f_{\text{lin}}) \ll \rho_{\text{adv}} (f_{\text{quad}})$. This result is intuitive, as $f_{\text{quad}}$ differentiates the images from their \textit{orientation}, unlike $f_{\text{lin}}$ that uses the \textit{bias} to distinguish them. The minimal perturbation required to switch the estimated label of $f_{\text{quad}}$ is therefore one that modifies the direction of the line, while a hardly perceptible perturbation that modifies the bias is enough to flip the label for $f_{\text{quad}}$. This explains the result originally illustrated in Fig. \ref{fig:vertical_example_2x2_2}. 

\section{Experimental results}
\label{sec:exp_results}

\subsection{Setting}
In this section, we illustrate our results on practical classification examples. Specifically, through
experiments on real data, we seek to confirm the identified limit on the robustness of classifiers, and
we show the large gap between adversarial and random robustness on real data. 
We also study more general classifiers to suggest that the trends obtained with our theoretical results are not limited to linear and quadratic classifiers.

Given a binary classifier $f$, and a datapoint $x$, we use an approach close to that of \citet{szegedy2013intriguing} to approximate $\Delta_{\text{adv}} (x;f)$. Specifically, we perform a line search to find the maximum $c > 0$ for which the minimizer of the following problem satisfies $f(x) f(x+r) \leq 0$:
\begin{align*}
\min_{r} c \| r \|_2 + L(f(x+r) \text{sign} (f(x))),
\end{align*}
where we set $L(x) = \max(0, x)$. The above problem (for $c$ fixed) is solved with a subgradient procedure, and we denote by $\widehat{\Delta}_{\text{adv}} (x;f)$ the obtained solution.\footnote{This procedure is not guaranteed to provide the optimal solution (for arbitrary classifiers $f$), as the problem is clearly non convex. Strictly speaking, the optimization procedure is only guaranteed to provide an upper bound on $\Delta_{\text{adv}} (x;f)$.} 
The empirical robustness to adversarial perturbations is then defined by $\widehat{\rho}_{\text{adv}} (f) = \frac{1}{m} \sum_{i=1}^m \widehat{\Delta}_{\text{adv}} (x_i;f)$, 
%
where $x_1, \dots, x_m$ denote the training points. To evaluate the robustness of $f$, we compare $\widehat{\rho}_{\text{adv}} (f)$ to the following quantity:
\begin{align}
\label{eq:kappa}
\kappa = \frac{1}{m} \sum_{i=1}^m \min_{j: y(x_j) \neq y(x_i)} \| x_i - x_j \|_2.
\end{align}
It represents the average norm of the minimal perturbation required to ``transform'' a training point to a training point of the opposite class, and can be seen as a distance measure between the two classes.
The quantity $\kappa$ therefore provides a baseline for comparing the robustness to adversarial perturbations, and we say that $f$ is not robust to adversarial perturbations when $\widehat{\rho}_{\text{adv}} (f) \ll \kappa$.
We also compare the adversarial robustness of the classifiers with their robustness to random uniform noise. We estimate $\Delta_{\text{unif}, \epsilon} (x;f)$ using a line search procedure that finds the largest $\eta$ for which the condition
\begin{align*}
\frac{1}{J} \# \{1 \leq j \leq J: f(x+n_j) f(x) \leq 0  \} \leq \epsilon,
\end{align*}
is satisfied, where $n_1, \dots, n_J$ are iid samples from the sphere $\eta \bb{S}$. By calling this estimate $\widehat{\Delta}_{\text{unif}, \epsilon} (x;f)$, the robustness of $f$ to uniform random noise is the empirical average over all training points, i.e.,  $\widehat{\rho}_{\text{unif}, \epsilon} (f) = \frac{1}{m} \sum_{i=1}^m \widehat{\Delta}_{\text{unif}, \epsilon} (x_i;f)$.
In the experiments, we set $J = 500$, and $\epsilon = 0.01$.\footnote{We compute the robustness to uniform random noise of all classifiers, except RBF-SVM, as this classifier is often asymmetric, assigning to one of the classes ``small pockets'' in the ambient space, and the rest of the space is assigned to the other class. In these cases, the robustness to uniform random noise can be equal to infinity for one of the classes, for a given $\epsilon$.}

\subsection{Binary classification using SVM}

We perform experiments on several classifiers: linear SVM (denoted \textit{L-SVM}), SVM with polynomial kernels of degree $q$ (denoted \textit{poly-SVM ($q$)}), and SVM with RBF kernel with a width parameter $\sigma^2$ (\textit{RBF-SVM($\sigma^2$)}). To train the classifiers, we use the efficient Liblinear \citep{REF08a} and LibSVM \citep{CC01a} implementations, and we fix the regularization parameters using a cross-validation procedure.

\begin{figure*}[ht]
\centering
\subfigure[]{
\includegraphics[width=0.12\textwidth]{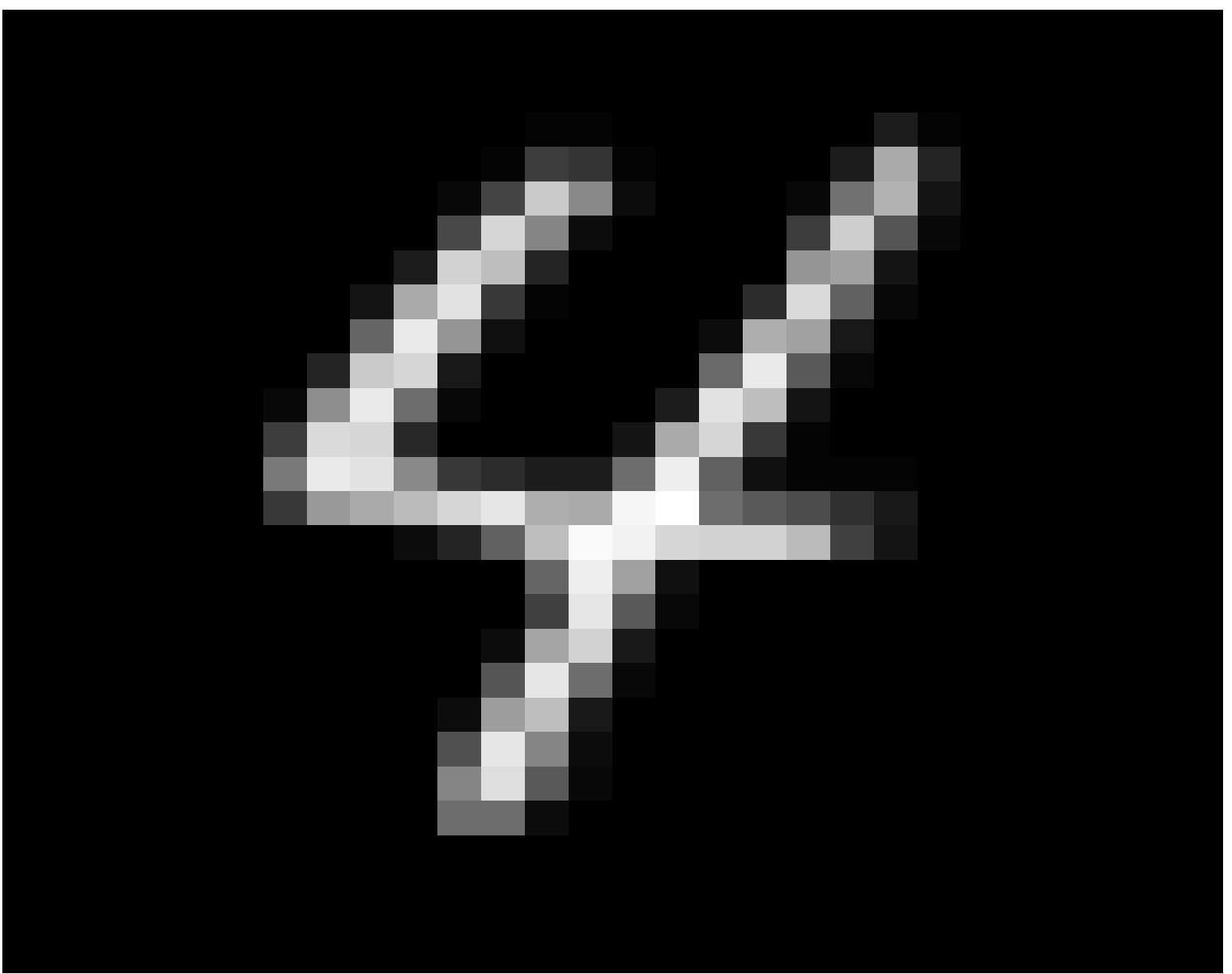}
}
\centering
\subfigure[$\Delta_{\text{adv}} = 0.08$]{
\includegraphics[width=0.12\textwidth]{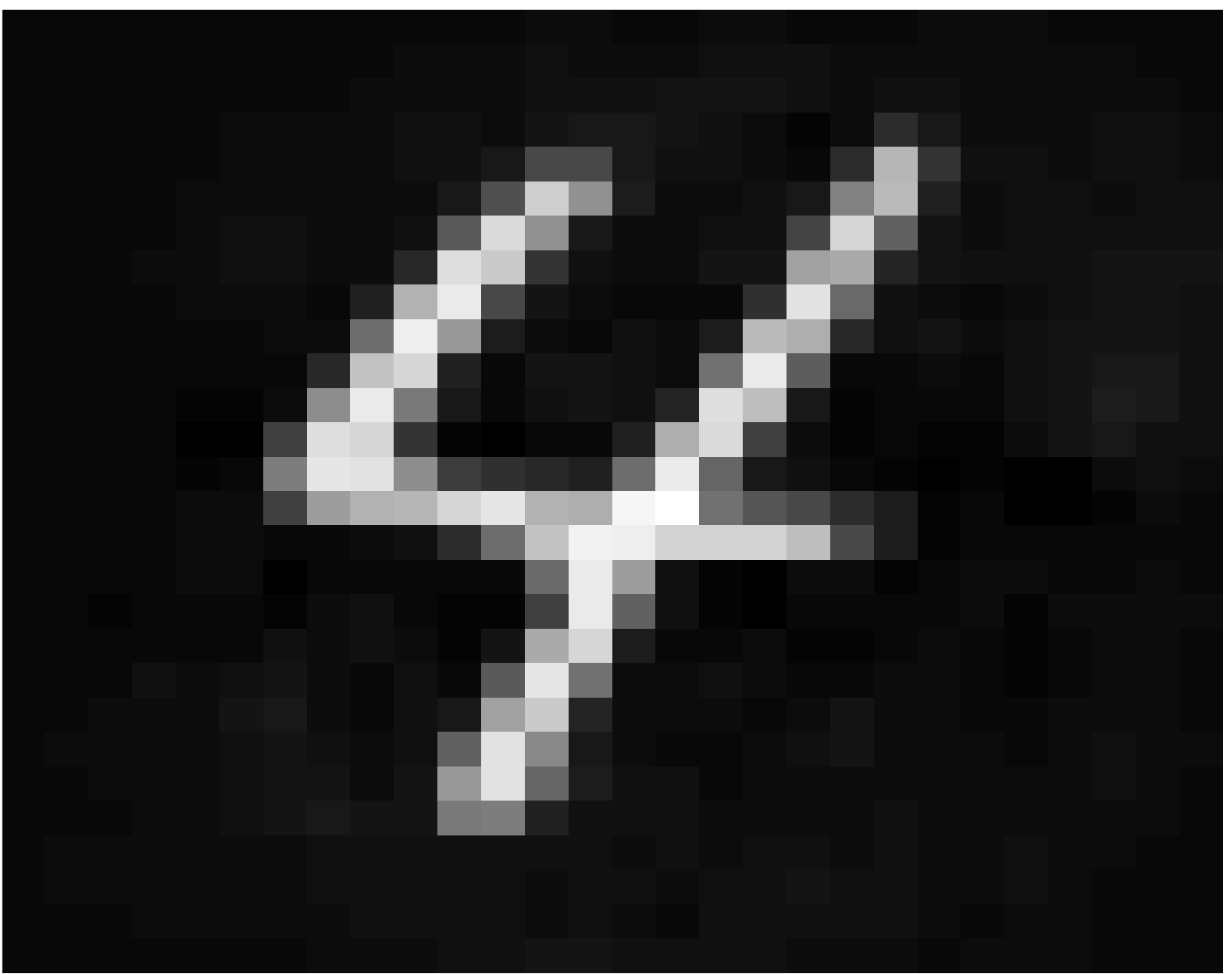}
}
\centering
\subfigure[$\Delta_{\text{adv}} = 0.19$]{
\includegraphics[width=0.12\textwidth]{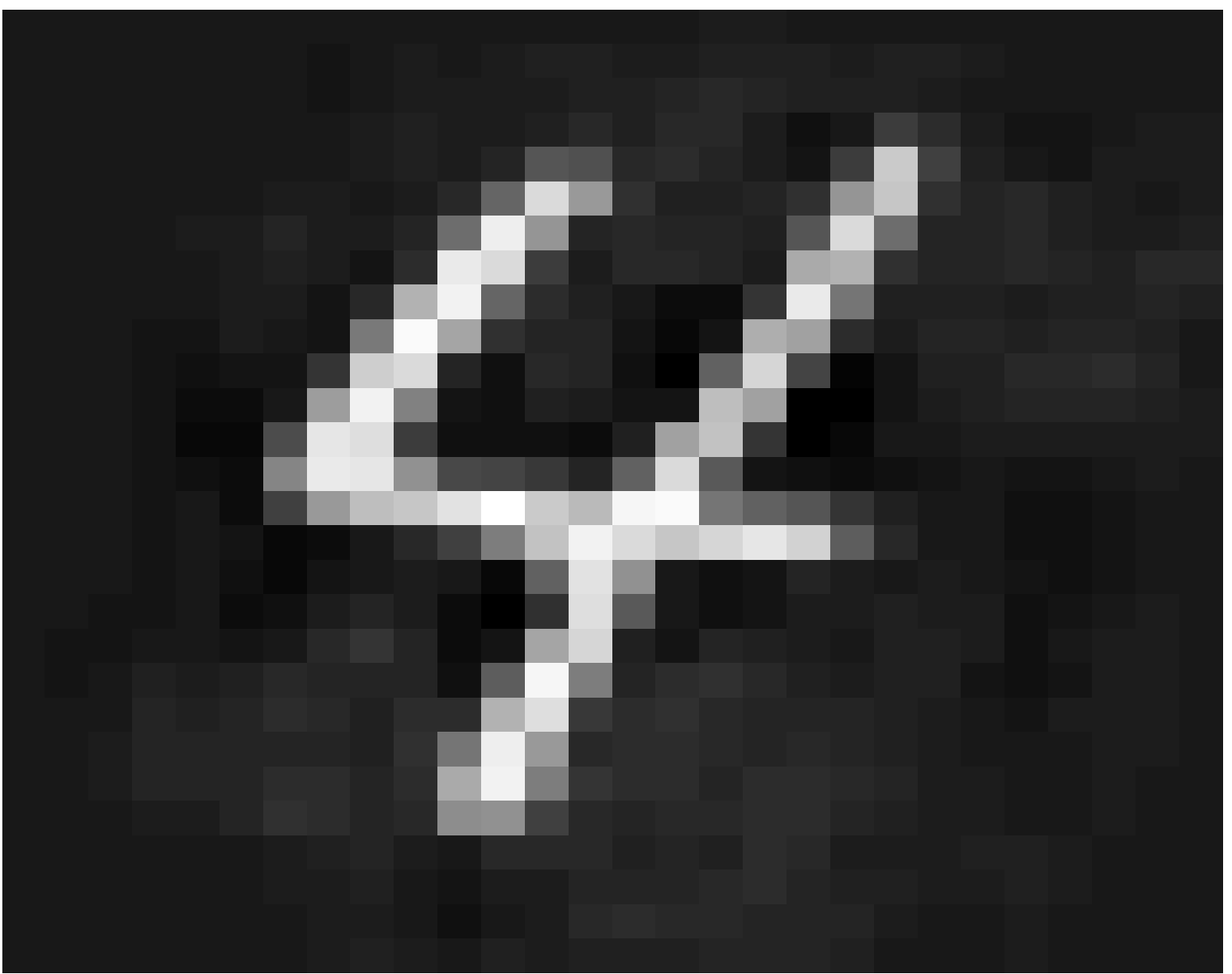}
}
\centering
\subfigure[$\Delta_{\text{adv}} = 0.21$]{
\includegraphics[width=0.12\textwidth]{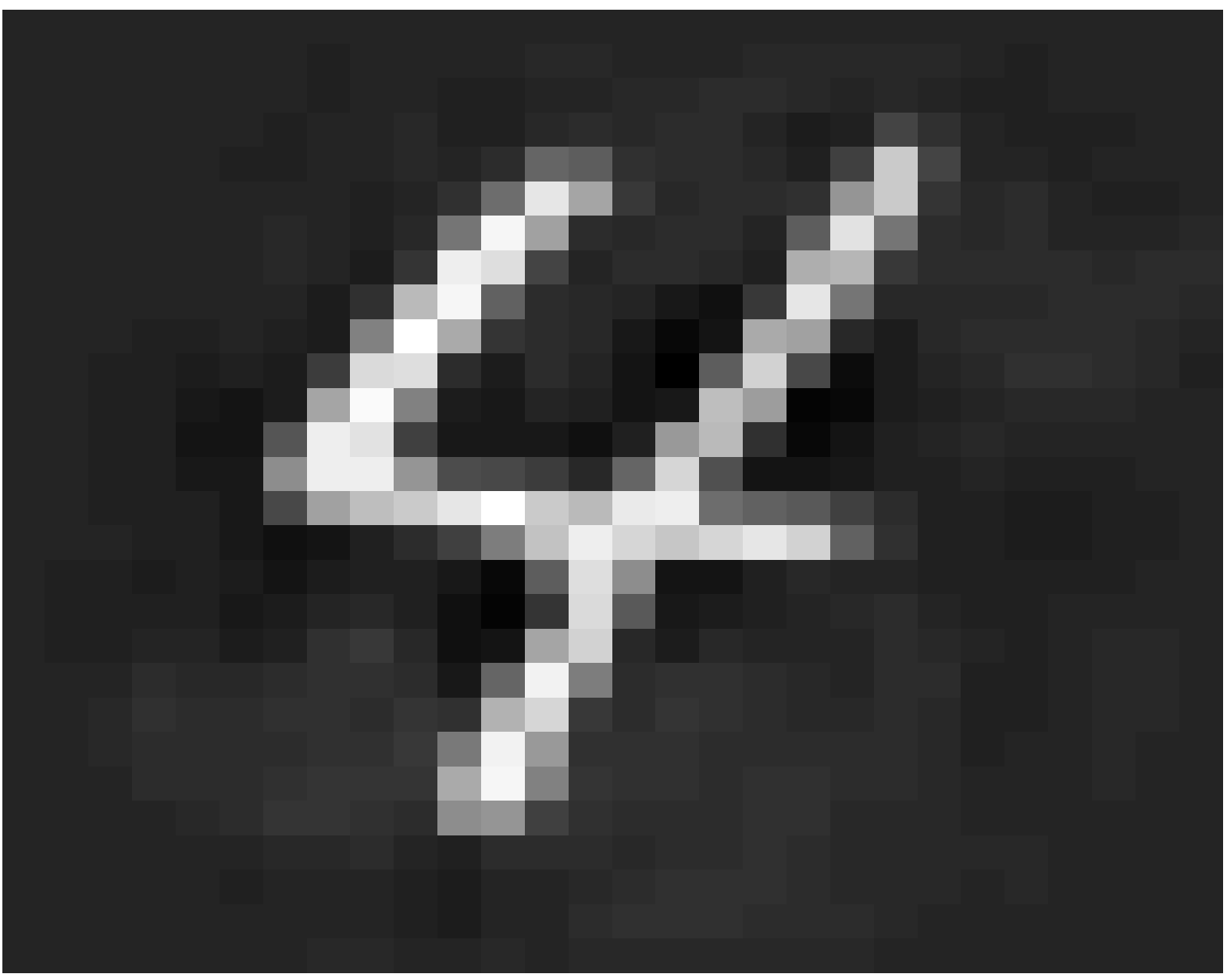}
}
\centering
\subfigure[$\Delta_{\text{adv}} = 0.15$]{
\includegraphics[width=0.12\textwidth]{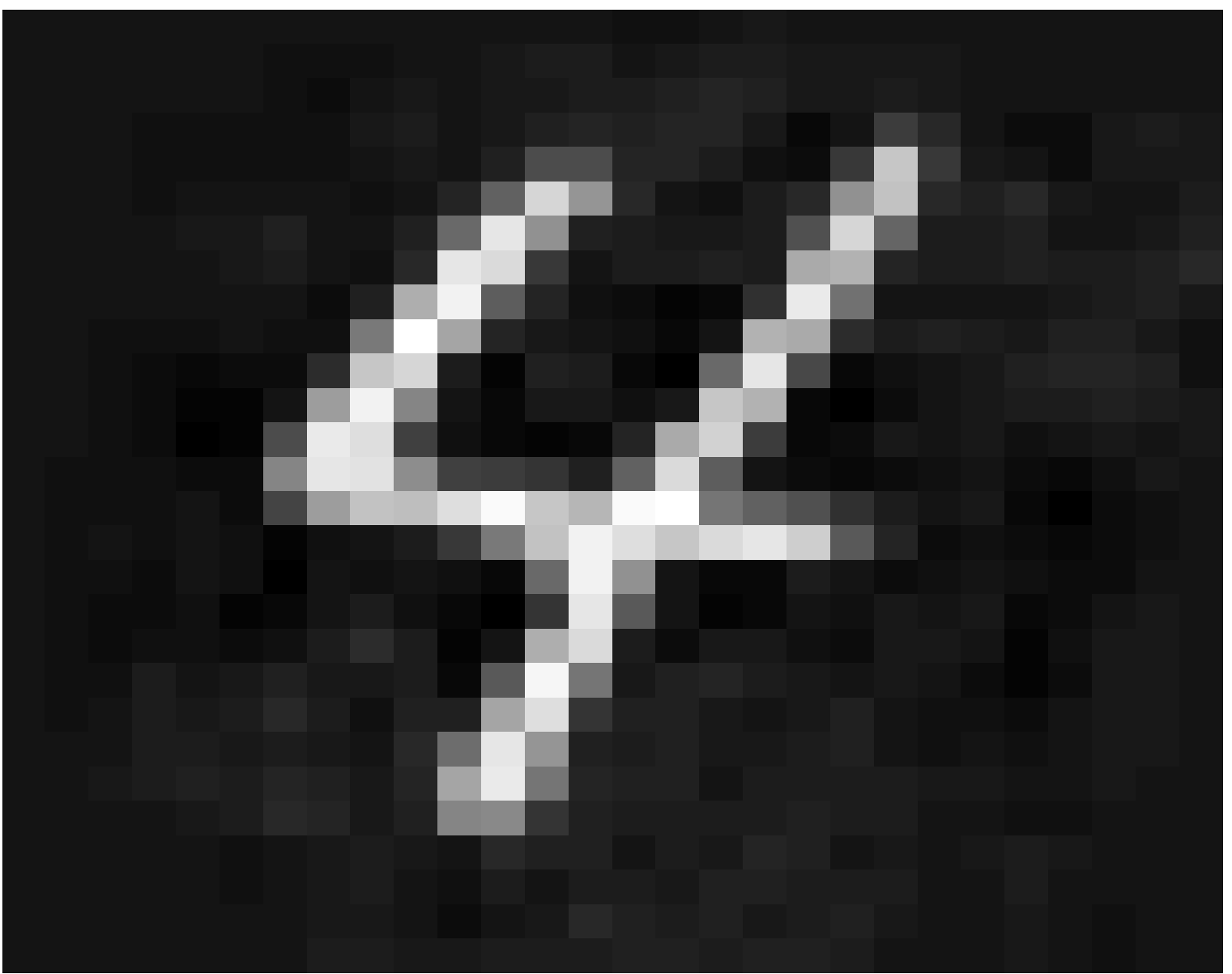}
}
\centering
\subfigure[$\Delta_{\text{adv}} = 0.41$]{
\includegraphics[width=0.12\textwidth]{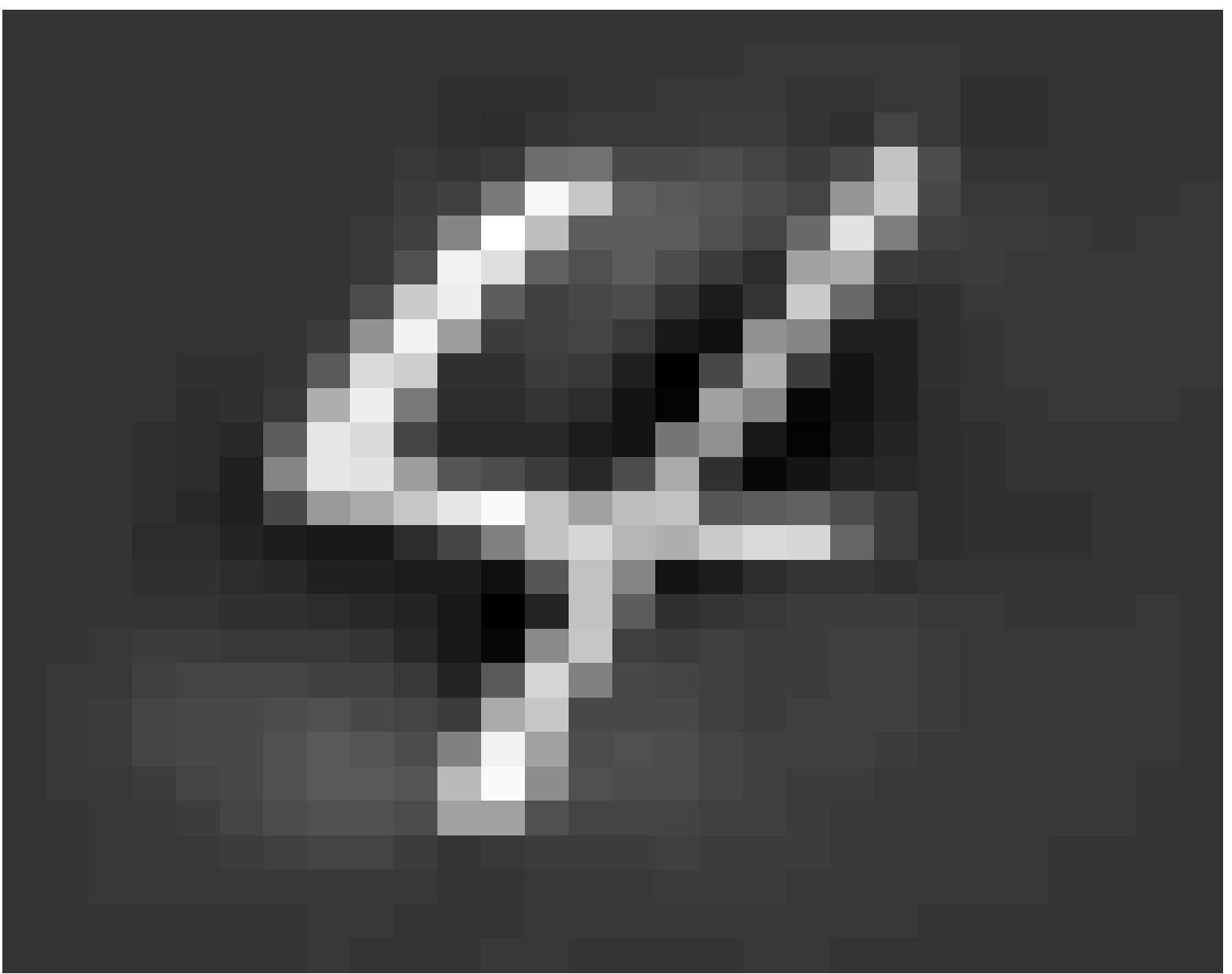}
}
\centering
\subfigure[$\Delta_{\text{unif}, \epsilon} = 0.8$]{
\includegraphics[width=0.12\textwidth]{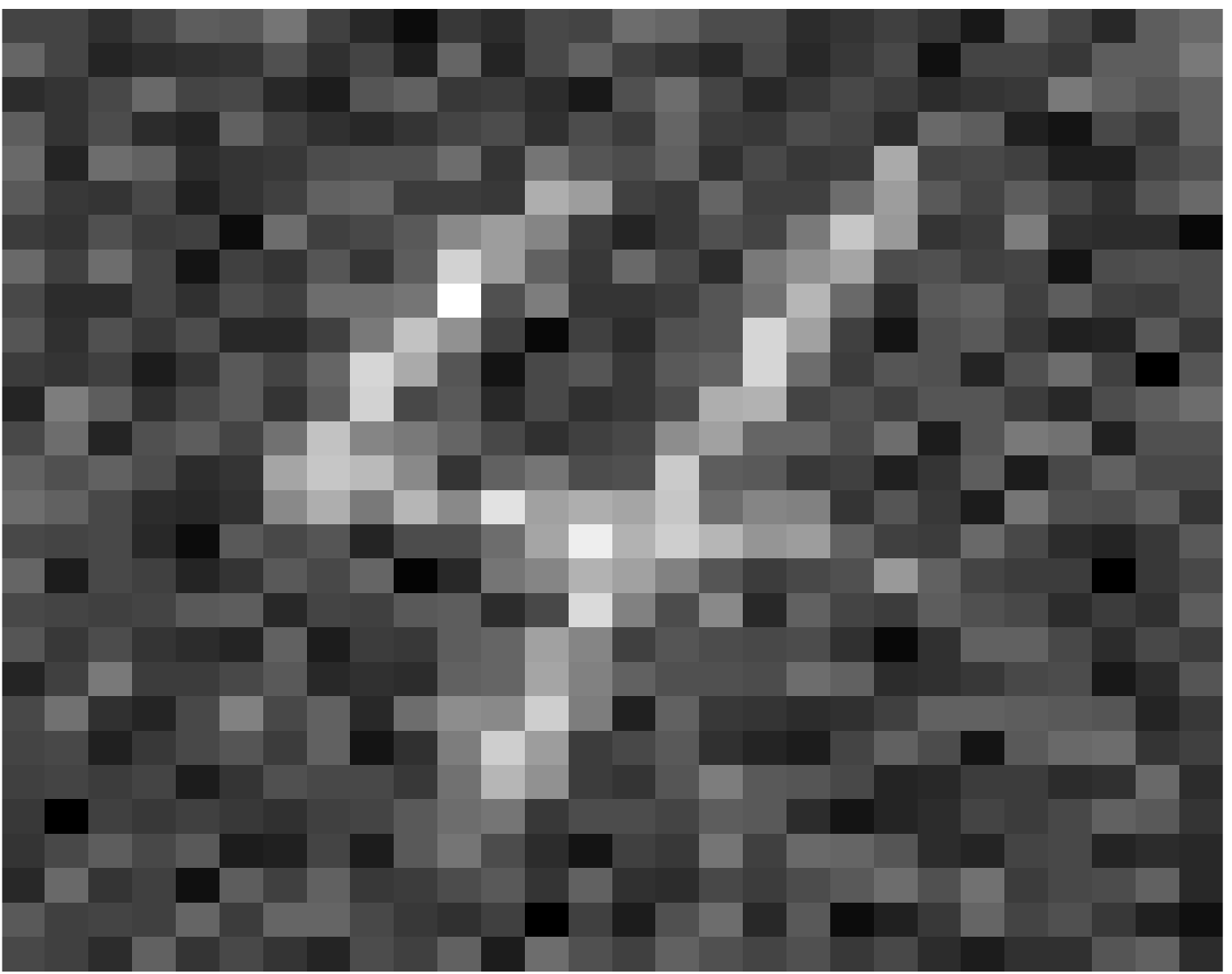}
}
\caption{\label{fig:fours_complexity}Original image (a) and minimally perturbed images (b-f) that switch the estimated label of linear (b), quadratic (c), cubic (d), RBF(1) (e), RBF(0.1) (f) classifiers. The image in (g) corresponds to the original image perturbed with a random uniform noise of norm $\Delta_{\text{unif}, \epsilon} (x; f)$, where $f$ is the learned linear classifier. That is, the linear classifier gives the same label to (a) and (g), with high probability. The norms of the perturbations are reported in each case.}
\end{figure*}

We first consider a classification task on the MNIST handwritten digits dataset \citep{lecun1998gradient}. We consider a digit ``4'' vs. digit ``5'' binary classification task, with $2,000$ and $1,000$ randomly chosen images for training and testing, respectively. In addition, a small random translation is applied to all images, and the images are normalized to be of unit Euclidean norm.
Table \ref{tab:table_perf_digits} reports the accuracy of the different classifiers, and their robustness to adversarial and random perturbations. Despite the fact that L-SVM performs fairly well on this classification task (both on training and testing), it is highly non robust to small adversarial perturbations. Indeed, $\widehat{\rho}_{\text{adv}} (f)$ is one order of magnitude smaller than $\kappa = 0.72$. 
Visually, this translates to an adversarial perturbation that is hardly perceptible. The instability of the linear classifier to adversarial perturbations is not surprising in the light of Theorem \ref{th:linear_expectation}, as the distinguishability term $\frac{1}{2} \| \bb{E}_{\mu_1} (x) - \bb{E}_{\mu_{-1}} (x) \|_2$ is small (see Table \ref{tab:digits_natural_images_datasets}).
In addition to improving the accuracy, the more flexible classifiers are also more robust to adversarial perturbations, as predicted by our theoretical analysis.
That is, the third order classifier is slightly more robust than the second order one, and RBF-SVM with small width $\sigma^2 = 0.1$ is more robust than with $\sigma^2 = 1$. Note that $\sigma$ controls the flexibility of the classifier in a similar way to the degree in the polynomial kernel. Interestingly, in this relatively easy classification task, RBF-SVM(0.1) achieves both a good performance, and a high robustness to adversarial perturbations. Concerning the robustness to random uniform noise, the results in Table \ref{tab:table_perf_digits} confirm the large gap between adversarial and random robustness for the linear classifier, as predicted by Theorem \ref{th:uniform_noise_linear}. Moreover, the results suggest that this gap is maintained for polynomial SVM. Fig. \ref{fig:fours_complexity} illustrates the robustness of the different classifiers on an example image.

\begin{table}[ht]
\centering
\begin{small}
\begin{tabular}{|>{\hspace{-3pt}}l<{\hspace{-4pt}}|>{\hspace{-4pt}}c<{\hspace{-4pt}}|>{\hspace{-4pt}}c<{\hspace{-4pt}}|c|c|}
\hline
Model & Train error (\%) & Test error (\%) & $\widehat{\rho}_{\text{adv}}$ & $\widehat{\rho}_{\text{unif}, \epsilon}$ \\ \hline
L-SVM & 4.8 &  7.0 & 0.08 & 0.97 \\ \hline \hline
poly-SVM($2$) & 0 & 1 & 0.19 & 2.15 \\ \hline
poly-SVM($3$) & 0 & 0.6 & 0.24 & 2.51 \\ \hline\hline
RBF-SVM($1$) & 0 & 1.1 & 0.16 & - \\ \hline
RBF-SVM($0.1$) & 0 & 0.5 & 0.32 & - \\ \hline
\end{tabular}
\end{small}
\caption{\label{tab:table_perf_digits} Training and testing accuracy of different models, and robustness to adversarial noise for the MNIST task. Note that for this example, we have $\kappa = 0.72$.}
\end{table}

\begin{figure*}[ht]
\centering
\subfigure[]{
\includegraphics[width=0.12\textwidth]{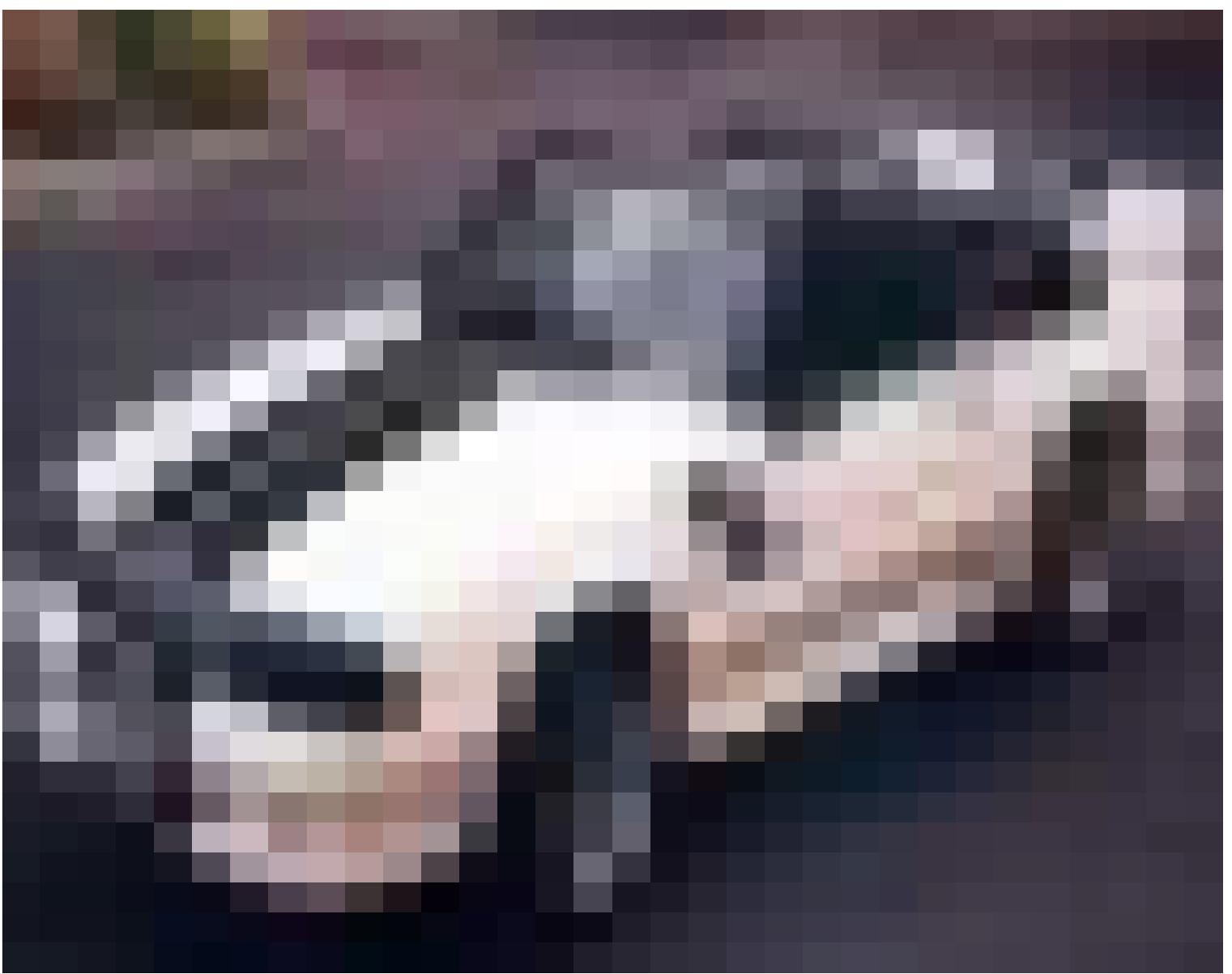}
}
\centering
\subfigure[$\Delta_{\text{adv}} = 0.04$]{
\includegraphics[width=0.12\textwidth]{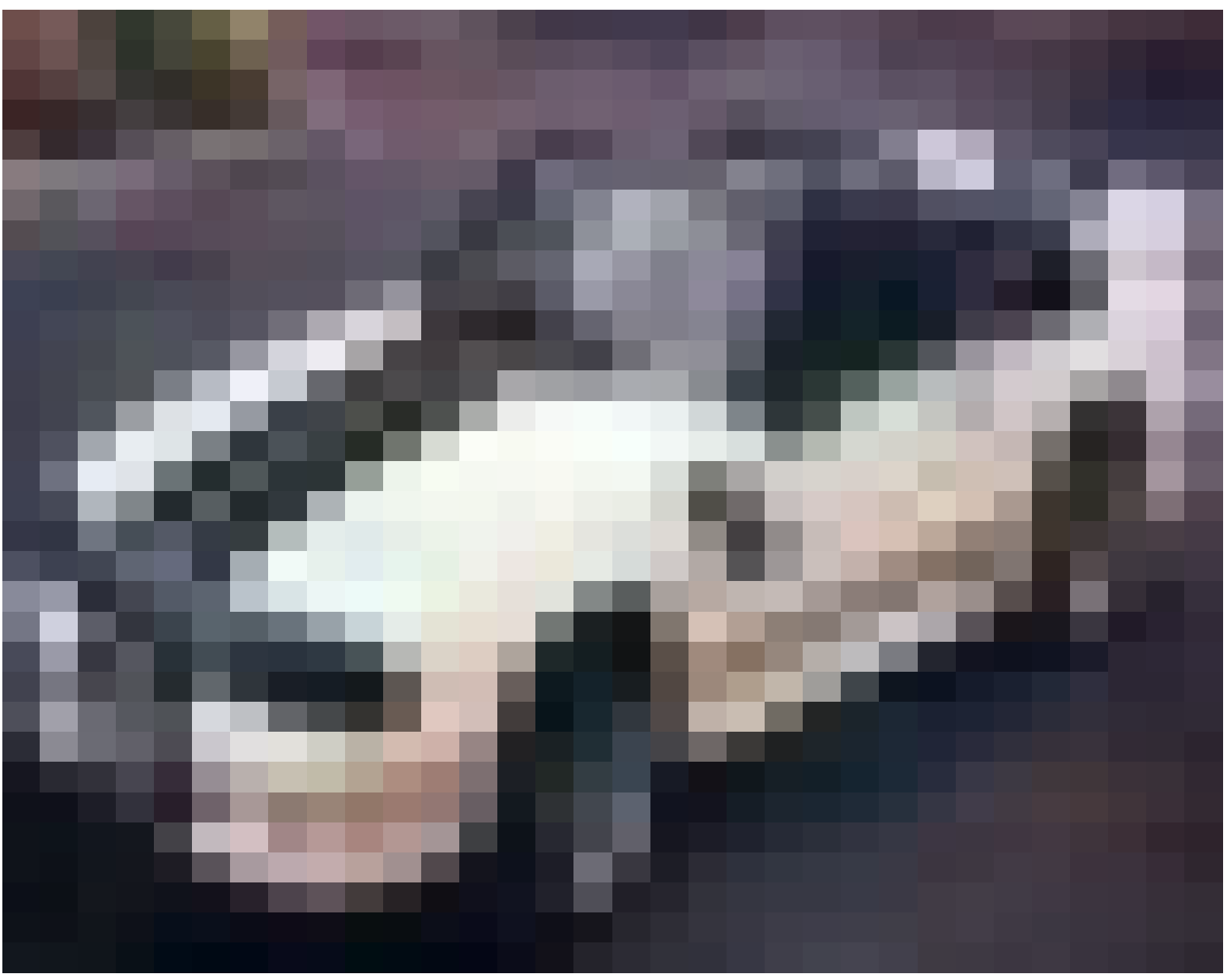}
}
\centering
\subfigure[$\Delta_{\text{adv}} = 0.02$]{
\includegraphics[width=0.12\textwidth]{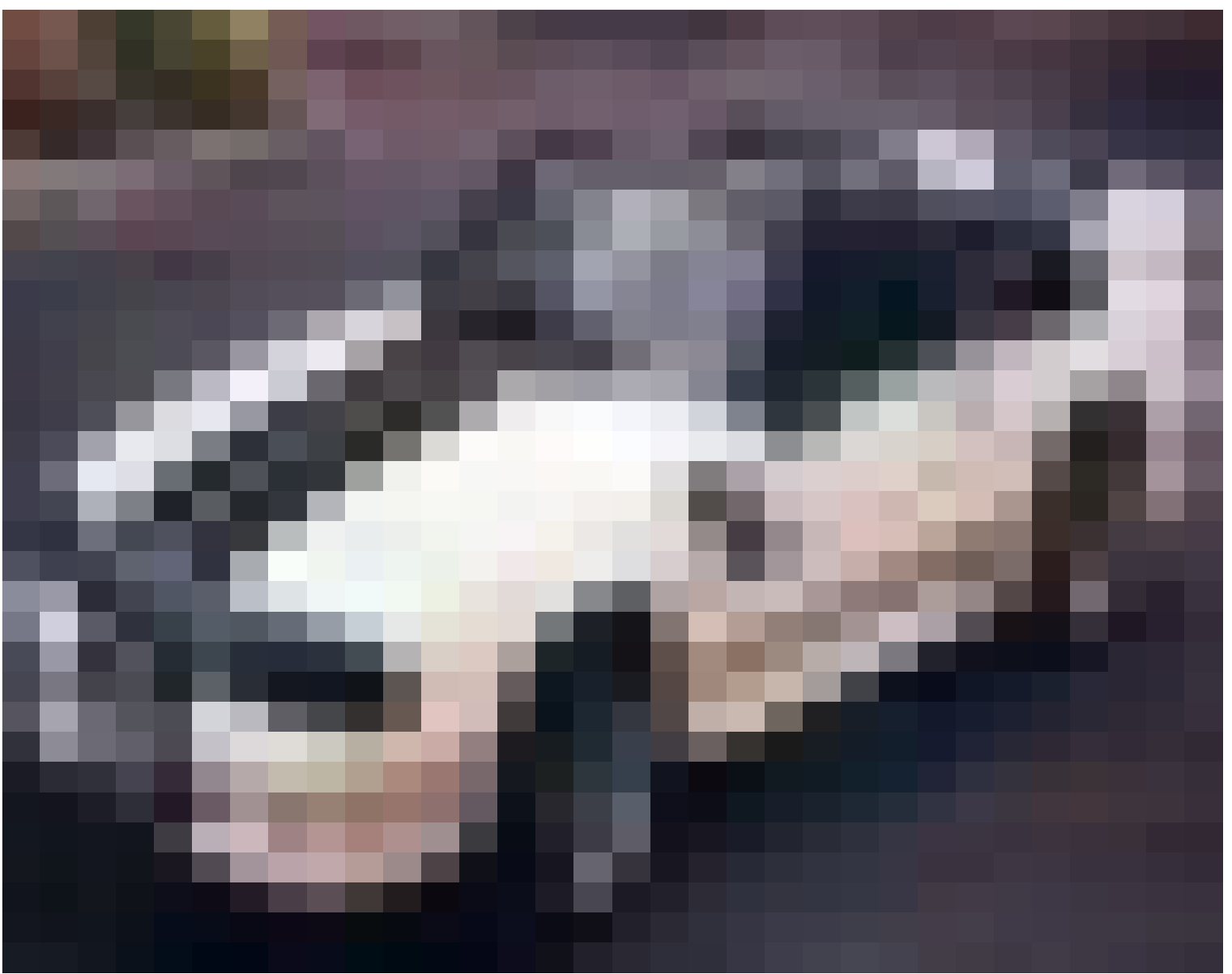}
}
\centering
\subfigure[$\Delta_{\text{adv}} = 0.03$]{
\includegraphics[width=0.12\textwidth]{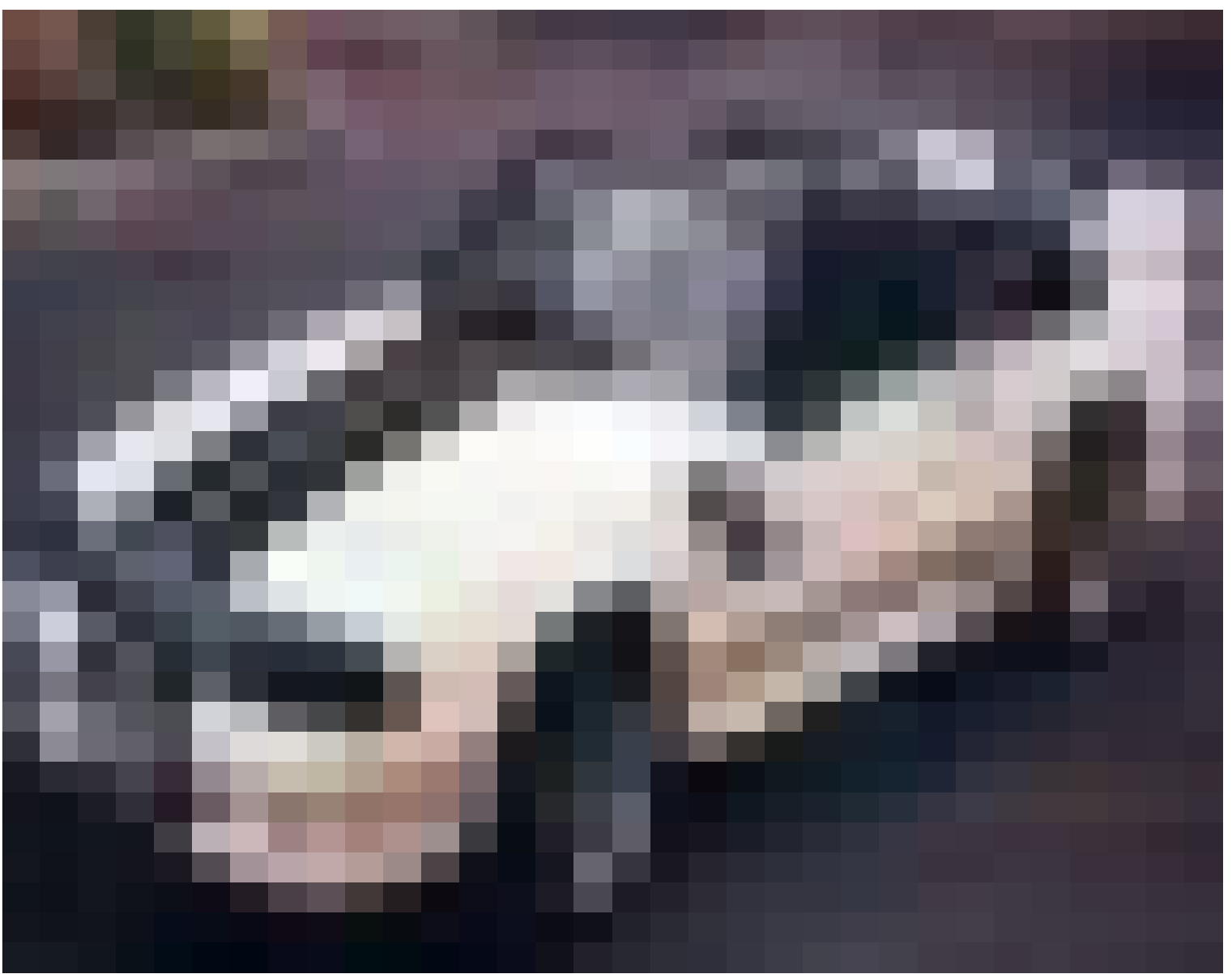}
}
\centering
\subfigure[$\Delta_{\text{adv}} = 0.03$]{
\includegraphics[width=0.12\textwidth]{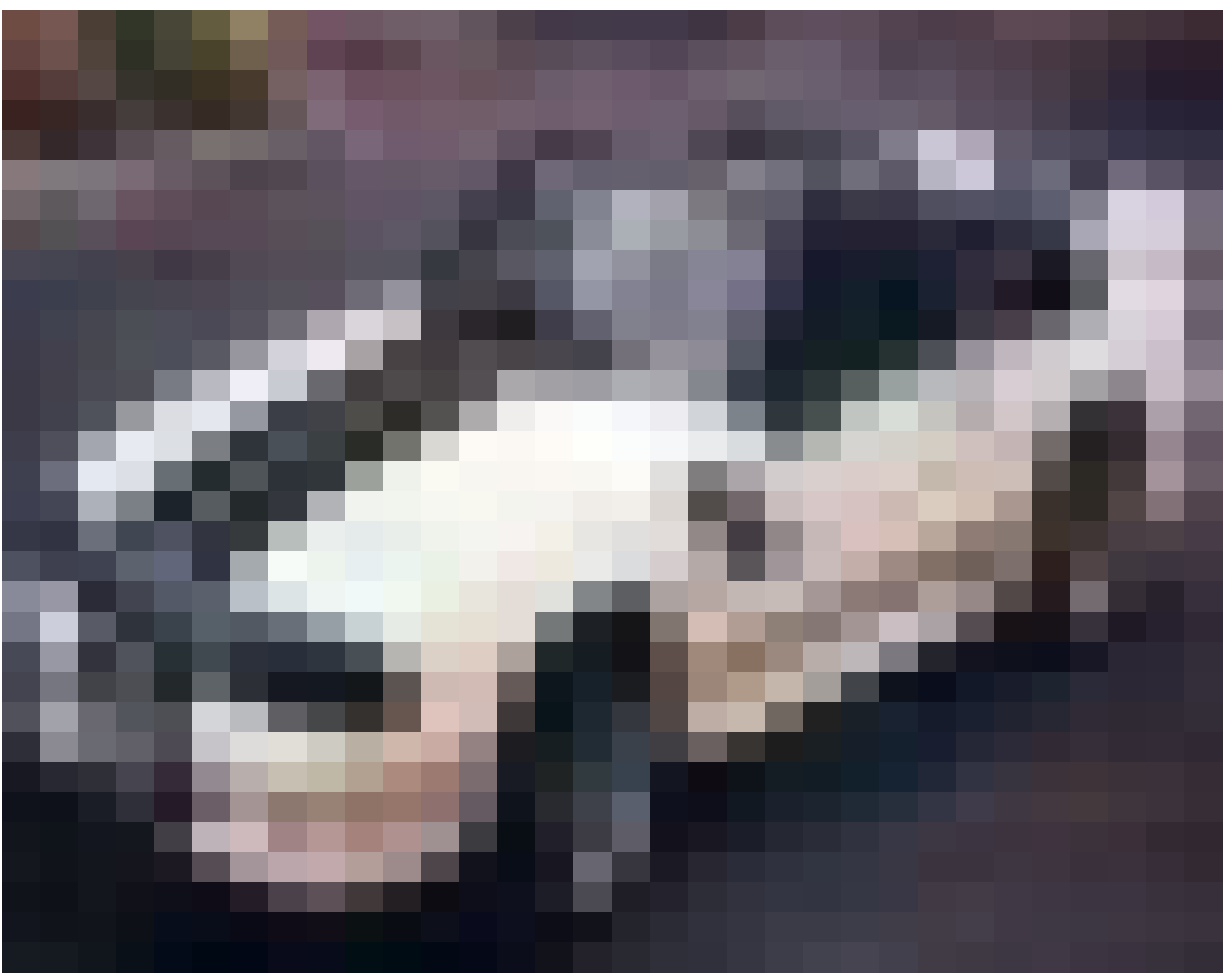}
}
\centering
\subfigure[$\Delta_{\text{adv}} = 0.05$]{
\includegraphics[width=0.12\textwidth]{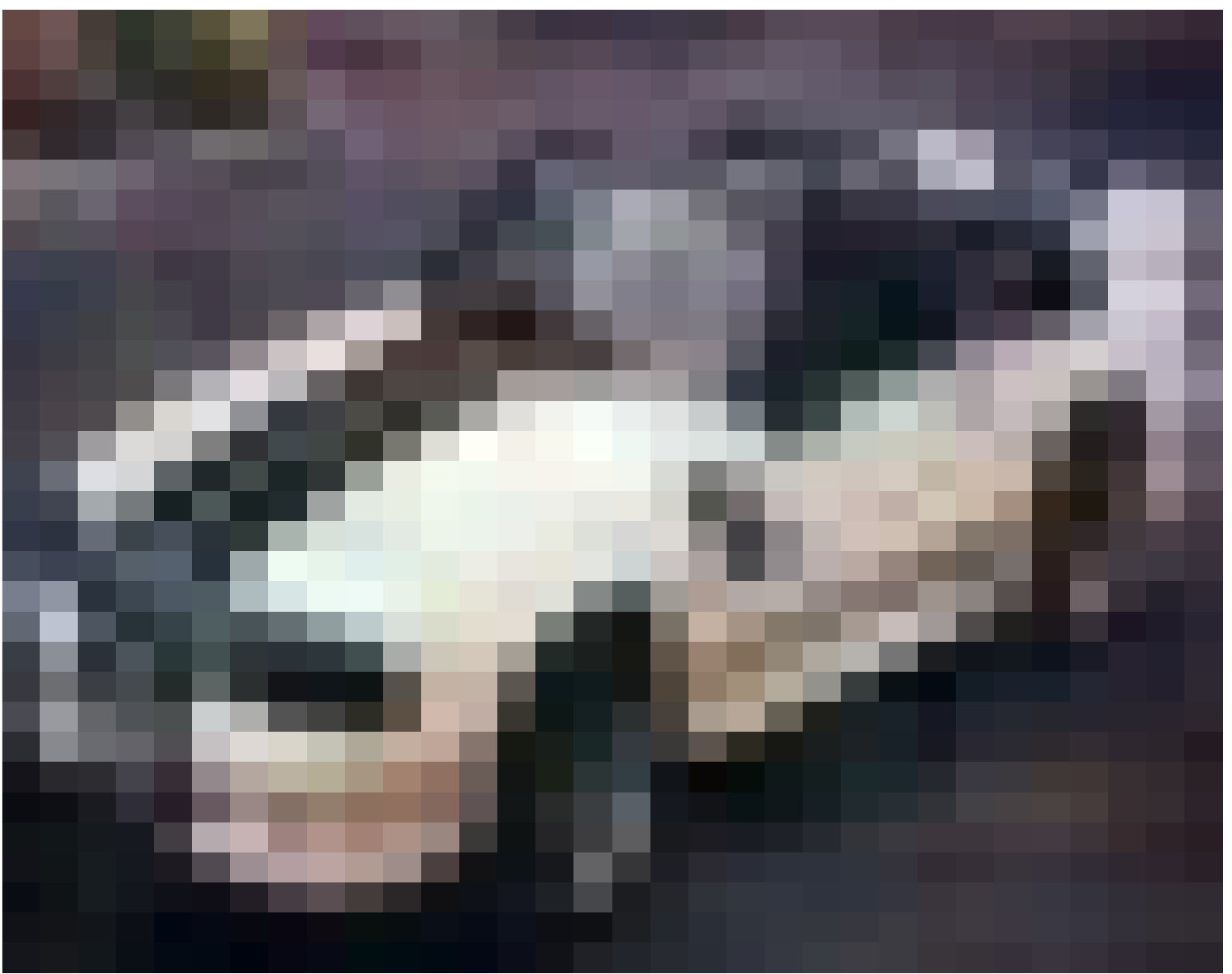}
}
\centering
\subfigure[$\Delta_{\text{unif}, \epsilon} = 0.8$]{
\includegraphics[width=0.12\textwidth]{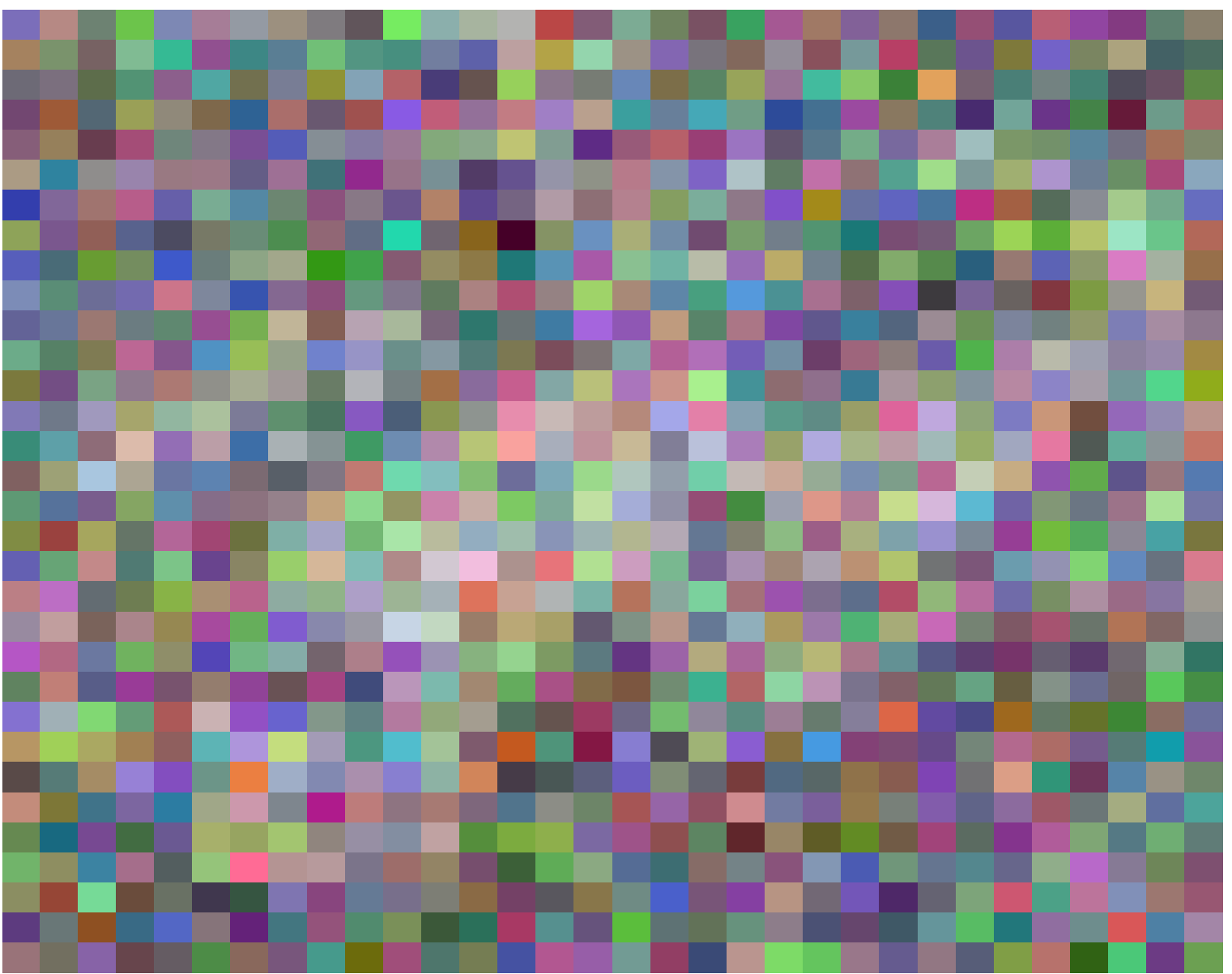}
}
\caption{\label{fig:cifar_complexity} Same as Fig. \ref{fig:fours_complexity}, but for the ``airplane'' vs. ``automobile'' classification task.}
\end{figure*}

We now turn to a natural image classification task, with images taken from the CIFAR-10 database \citep{krizhevsky2009learning}. The database contains $10$ classes of $32 \times 32$ RGB images. We restrict the dataset to the first two classes (``airplane'' and ``automobile''), and consider a subset of the original data, with $1,000$ images for training, and $1,000$ for testing. Moreover, all images are normalized to be of unit Euclidean norm. Compared to the first dataset, this task is more difficult, as the variability of the images is much larger than for digits. 
We report the results in Table \ref{tab:table_perf_digits_CIFAR}.
It can be seen that \textit{all} classifiers are not robust to adversarial perturbations for this experiment, as $\rho_{\text{adv}} (f) \ll \kappa = 0.39$. Despite that, all classifiers (except L-SVM) achieve an accuracy around $85 \%$, and a training accuracy above $92\%$, and are robust to uniform random noise. Fig. \ref{fig:cifar_complexity} illustrates the robustness to adversarial and random noise of the learned classifiers, on an example image of the dataset. Compared to the digits dataset, the distinguishability measures for this task are smaller (see Table \ref{tab:digits_natural_images_datasets}). Our theoretical analysis therefore predicts a lower limit on the adversarial robustness of linear and quadratic classifiers for this task (even though the bound for quadratic classifiers is far from the achieved robustness of poly-SVM(2) in this example).

\begin{table}[t]
\centering
\begin{small}
\begin{tabular}{|>{\hspace{-3pt}}l<{\hspace{-4pt}}|>{\hspace{-4pt}}c<{\hspace{-4pt}}|>{\hspace{-4pt}}c<{\hspace{-4pt}}|c|c|}\hline
Model & Train error (\%) & Test error (\%) & $\widehat{\rho}_{\text{adv}}$ & $\widehat{\rho}_{\text{unif}, \epsilon}$ \\ \hline
L-SVM & 14.5 & 21.3 & 0.04 & 0.94 \\ \hline \hline
poly-SVM($2$) & 4.2 & 15.3 &  0.03 & 0.73 \\ \hline
poly-SVM($3$) & 4 & 15 &  0.04 & 0.89 \\ \hline\hline
RBF-SVM($1$) & 7.6 & 16 & 0.04 & - \\ \hline
RBF-SVM($0.1$) & 0 & 13.1 & 0.06 & - \\ \hline
\end{tabular}
\end{small}
\caption{\label{tab:table_perf_digits_CIFAR} Training and testing accuracy of different models, and robustness to adversarial noise for the CIFAR task. Note that for this example, we have $\kappa = 0.39$.}
\end{table}

\begin{table}[t]
\centering
\begin{small}
\begin{tabular}{|l|l|c|c|}
\hline
 Quantity & Definition & Digits & Natural images \\ \hline
Distance between classes & $\kappa$ (see Eq. (\ref{eq:kappa})) & 0.72 & 0.39 \\ \hline
Distinguishability (linear class.) & $\| p_1 \bb{E}_{\mu_1} (x) - p_{-1} \bb{E}_{\mu_{-1}} (x) \|_2$ & 0.14 & 0.06 \\\hline
Distinguishability (quadratic class.) & $2 \sqrt{K \| p_{1} C_1 - p_{-1} C_{-1} \|_{*}}$ & 1.4 & 0.87 \\ \hline
\end{tabular}
\end{small}
\caption{\label{tab:digits_natural_images_datasets} The parameter $\kappa$, and distinguishability measures for the two classification tasks. For the numerical computation, we used $K = 1$.}
\end{table}

The instability of all classifiers to adversarial perturbations on this task suggests that the essence of the classification task was not correctly captured by these classifiers, even if a fairly good test accuracy is reached. To reach better robustness, two possibilities exist: use a more flexible family of classifiers (as our theoretical results suggest that more flexible families of classifiers achieve better robustness), or use a better training algorithm for the tested nonlinear classifiers. The latter solution seems possible, as the theoretical limit for quadratic classifiers suggests that there is still room to improve the robustness of these classifiers.

\subsection{Multiclass classification using CNN}
\label{sec:exp_deep}
Since our theoretical results suggest that more flexible classifiers achieve better robustness to adversarial perturbations in the binary case, we now explore empirically whether the same intuitions hold in scenarios that depart from the theory in two different ways: (i) we consider \textit{multiclass} classification problems, and (ii) we consider convolutional neural network architectures. While classifiers' flexibility is relatively well quantified for polynomial classifiers by the degree of the polynomials, this is not straightforward to do for neural network architectures. In this section, we examine the effect of \textit{breadth} and \textit{depth} on the robustness to adversarial perturbations of classifiers.

\begin{figure}[ht]
\centering
\subfigure[Evolution with depth] {
\includegraphics[scale=0.25]{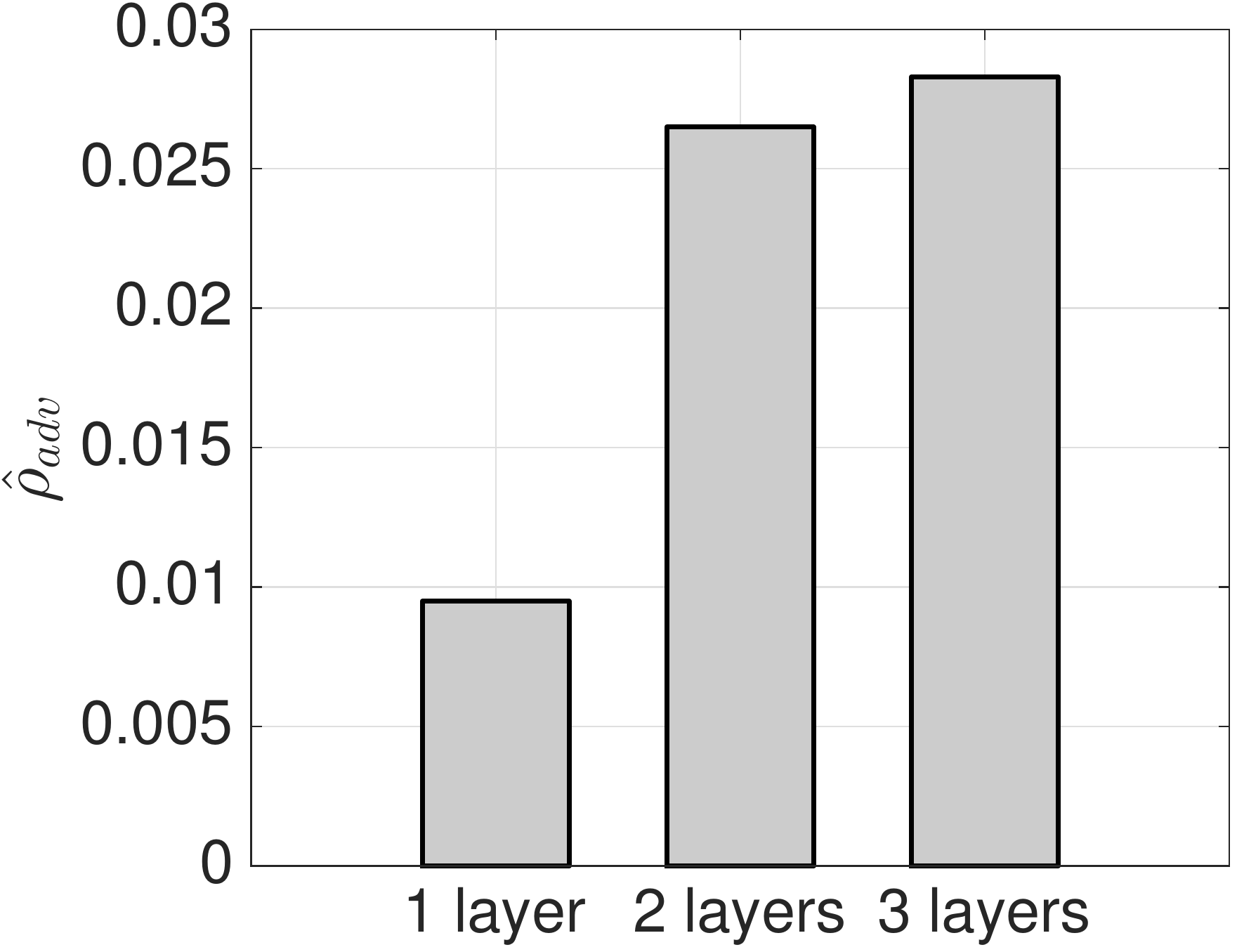}
}
\subfigure[Evolution with breadth] {
\includegraphics[scale=0.25]{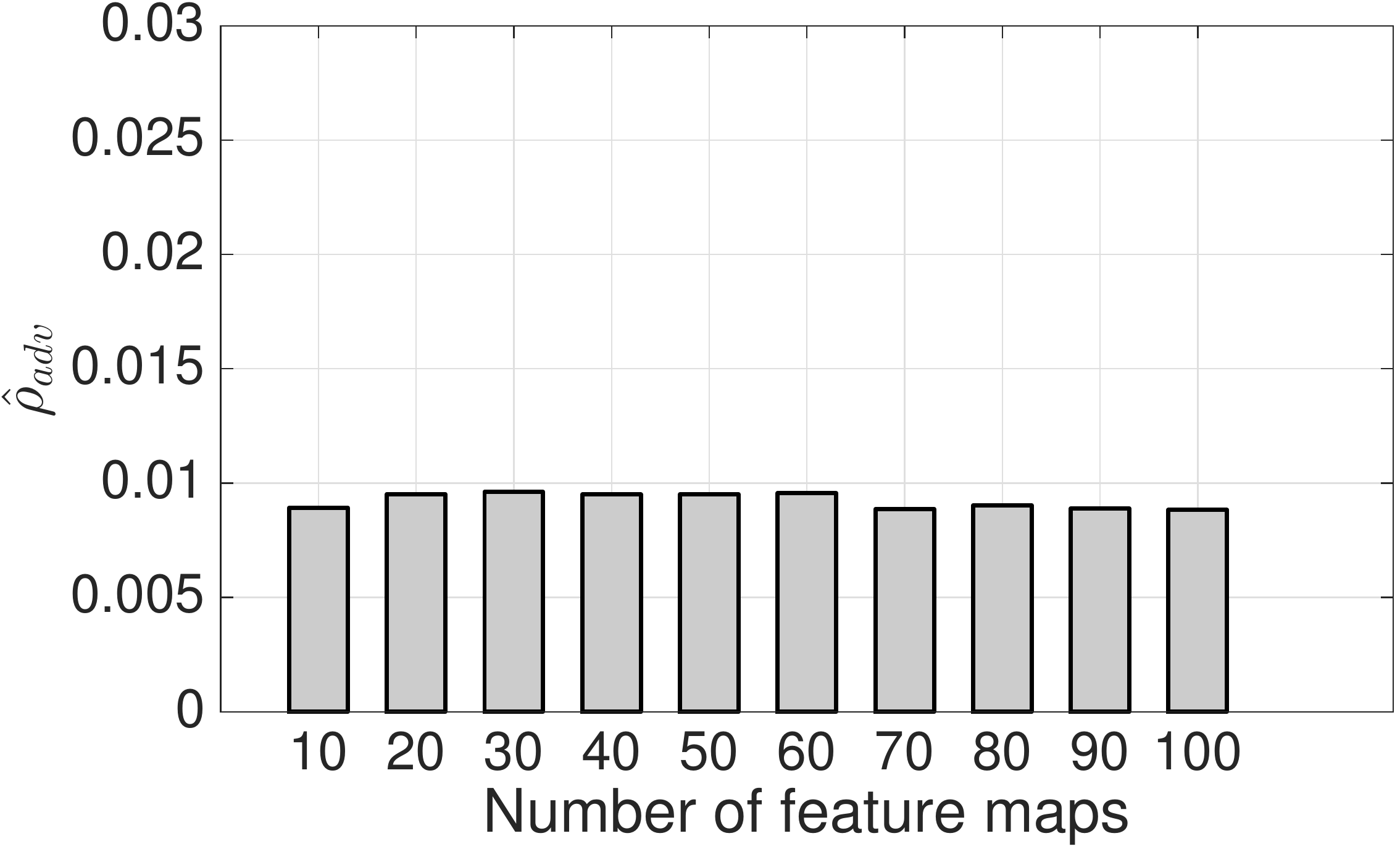}
}
\caption{\label{fig:evolution_numLayers}Evolution of the normalized robustness of classifiers with respect to (a) the depth of a CNN for CIFAR-10 task, and (b) the number of feature maps.}
\end{figure}

We perform experiments on the multiclass CIFAR-10 classification task, and use the recent method in \citep{moosavi2015deepfool} to compute adversarial examples in the multiclass case. We focus on baseline CNN classifiers, and learn architectures with 1, 2 and 3 hidden layers. Specifically, each layer consists of a successive combination of convolutional, rectified linear units and pooling operations. The convolutional layers consist of 5 $\times$ 5 filters with 50 feature maps for each layer, and the pooling operations are done on a
window of size 3$\times$3 with a stride parameter of 2. We build the three architectures gradually, by successively stacking a new hidden layer on top of the previous architecture (kept fixed). The last hidden layer is then connected to a fully connected layer, and the softmax loss is used. All architectures are trained with stochastic gradient descent. To provide a fair comparison of the different classifiers, all three classifiers have approximately similar classification error ($35\%$). To ensure similar accuracies, we perform an early stop of the training procedure when necessary. The empirical normalized robustness to adversarial perturbations of the three networks are compared in Figure \ref{fig:evolution_numLayers} (a).\footnote{More precisely, we report $\frac{1}{m} \sum_{i=1}^m \frac{\hat{\Delta}_{\text{adv}} (x_i; f)}{\| x_i \|_2}$.} 

We observe first that increasing the depth of the network leads to a significant increase in the robustness to adversarial perturbations, especially from $1$ to $2$ layers. The depth of a neural network has an important impact on the robustness of the classifier, just like the degree of a polynomial classifier is an important factor for the robustness. Going from $2$ to $3$ layers however seems to have a marginal effect on the robustness. It should be noted that, despite the increase of the robustness with the depth, the normalized robustness computed for all classifiers is relatively small, which suggests that none of these classifiers is really robust to adversarial perturbations. Note also that the results in Figure   \ref{fig:evolution_numLayers} (a) showing an increase of the robustness with the depth are inline with recent results showing that depth provides robustness to adversarial \textit{geometric} transformations \citep{BMVC2015_106}. 
In Fig. \ref{fig:evolution_numLayers} (b), we show the effect of the number of feature maps in the CNN (for a one layer CNN) on the estimated normalized robustness to adversarial perturbations. Unlike the effect of depth, we observe that the number of feature maps has barely any effect on the robustness to adversarial perturbations. 
Finally, a comparison of the normalized robustness measures of very deep networks VGG-16 and VGG-19 \citep{simonyan2014very} on ImageNet shows that these two networks behave very similarly in terms of robustness (both achieve a normalized robustness of $3 \cdot 10^{-3}$). This experiment, along with the experiment in Figure \ref{fig:evolution_numLayers} (a), empirically suggest that adding layers on top of shallow network helps in terms of adversarial robustness, but if the depth of the network is already sufficiently large, then adding layers only moderately changes that robustness. 

\section{Discussion and perspectives}
\label{sec:conclusion}

The existence of a general limit on the adversarial robustness of classifiers (established in Lemma \ref{lemma:main_result}) is an important phenomenon with many practical implications. To better understand the implications of this limit, we derived specialized upper bounds for two families of classifiers. 
For the family of linear classifiers, the established limit is very small for most problems of interest. Hence, linear classifiers are usually not robust to adversarial noise (even though robustness to random noise might be achieved). This is however different for nonlinear classifiers: for the family of quadratic classifiers, the limit on adversarial robustness is usually larger than for linear classifiers, which gives hope to have classifiers that are robust to adversarial perturbations. In fact, by using an appropriate training procedure, it might be possible to get closer to the theoretical bound. For general nonlinear classifiers (e.g., neural networks), designing training procedures that specifically take into account the robustness in the learning is an important future work. We also believe that the application of our general upper bound in Lemma \ref{lemma:main_result} to derive explicit upper bounds that are specific to e.g., deep neural networks is an important future work. To do that, we believe that it is important to derive explicitly the parameters $(\tau, \gamma)$ of assumption (A) for the class of functions under consideration. Results from algebraic geometry seem to suggest that establishing such results might be possible for general classes of functions (e.g., piecewise linear functions). In addition, experimental results suggest that, unlike the breadth of the neural network, the depth plays a crucial role in the adversarial robustness. 
Identifying an upper bound on the adversarial robustness of deep neural networks in terms of the depth of the network would be a great step towards having a better understanding of such systems. 

\begin{small}
\section*{Acknowledgments}

We thank Hamza Fawzi, Ian Goodfellow for discussions and comments on an early draft of the paper, and Guillaume Aubrun for pointing out a reference for Theorem \ref{th:matousek}. We also thank Seyed Mohsen Moosavi for his help in preparing experiments.
\end{small}

\bibliography{reference.bib}
\bibliographystyle{spbasic}

\newpage
\appendix

\section{Proofs}
\label{sec:proofs}

\subsection{Proof of Lemma \ref{lemma:main_result}}
\label{sec:proof_lemma_main}
We begin by proving the following inequality:
\begin{lemma}
\label{lemma:lemma_inequality}
Let $z_1, \dots, z_n$ be non-negative real numbers, and let $0 \leq \gamma \leq 1$. Then, 
\begin{align*}
\sum_{i=1}^n z_i^{\gamma} \leq n^{1-\gamma} \left( \sum_{i=1}^n z_i\right)^{\gamma}.
\end{align*}
\end{lemma}
\begin{proof}
We prove that the quantity
\begin{align*}
\frac{\sum_{i=1}^n z_i^{\gamma}}{\left(\sum_{i=1}^n z_i\right)^{\gamma}} = \sum_{i=1}^n \left( \frac{z_i}{\sum_{i=1}^n z_i} \right)^{\gamma}
\end{align*}
is bounded from above by $n^{1-\gamma}$. To do so, let $u_i = \frac{z_i}{\sum_{i=1}^n z_i}$, and let us determine the maximum of the concave function $g(u_1, \dots, u_{n-1}) = u_1^{\gamma} + \dots + (1-u_1-\dots-u_{n-1})^{\gamma}$. Setting the derivative of $g$ with respect to $u_i$ to zero, we get
\begin{align*}
u_i^{\gamma-1} - (1-u_1-\dots-u_{n-1})^{\gamma-1} = 0,
\end{align*}
hence $u_i = 1 - u_1 - \dots - u_{n-1}$. We therefore get $u_1 = \dots = u_{n-1}$, and conclude that the maximum of $\sum_{i=1}^n \left( \frac{z_i}{\sum_{i=1}^n z_i} \right)^{\gamma}$ is reached when $z_1 = \dots = z_n$ and the value of the maximum is $n^{1-\gamma}$. 
\end{proof}

We now prove Lemma \ref{lemma:main_result}.

\begin{proof}
The goal is to find an upper bound on $\rho_{\text{adv}} (f) = \mathbb{E}_{\mu} (\Delta_{\text{adv}} (x; f))$.
\begin{align*}
\rho_{\text{adv}} (f) & = p_1 \bb{E}_{\mu_1} (\Delta_{\text{adv}} (x;f)) + p_{-1} \bb{E}_{\mu_{-1}} (\Delta_{\text{adv}} (x;f)) \\
							  & = p_1 \Bigl( \bb{E}_{\mu_1} (\Delta_{\text{adv}} (x;f) | f(x) \geq 0) \bb{P}_{\mu_1} (f(x) \geq 0) + \bb{E}_{\mu_{1}} (\Delta_{\text{adv}} (x;f) | f(x) < 0) \bb{P}_{\mu_{1}} (f(x) < 0) \Bigr) \\
							  & + p_{-1} \Bigl( \bb{E}_{\mu_{-1}} (\Delta_{\text{adv}} (x;f) | f(x) < 0) \bb{P}_{\mu_{-1}} (f(x) < 0) + \bb{E}_{\mu_{-1}} (\Delta_{\text{adv}} (x;f) | f(x) \geq 0) \bb{P}_{\mu_{-1}} (f(x) \geq 0) \Bigr).
\end{align*}
Using assumption (A), the following upper bounds hold:
\begin{align*}
\bb{E}_{\mu_{\pm1}} (\Delta_{\text{adv}} (x;f) | f(x) \geq 0) & \leq \tau \bb{E}_{\mu_{\pm1}} (f(x)^\gamma | f(x) \geq 0) \\
\bb{E}_{\mu_{\pm1}} (\Delta_{\text{adv}} (x;f) | f(x) < 0) & \leq \tau \bb{E}_{\mu_{\pm1}} ((-f(x))^\gamma | f(x) < 0)
\end{align*}
Hence, we obtain the following inequality on $\rho_{\text{adv}} (f)$:
\begin{align*}
\rho_{\text{adv}} (f) & \leq \tau p_1 \Bigl( \bb{E}_{\mu_1} (f(x)^\gamma | f(x) \geq 0) \bb{P}_{\mu_1} (f(x) \geq 0) + \bb{E}_{\mu_{1}} ((-f(x))^\gamma | f(x) < 0) \bb{P}_{\mu_{1}} (f(x) < 0) \Bigr) \\
							  & + \tau p_{-1} \Bigl( \bb{E}_{\mu_{-1}} ((-f(x))^\gamma | f(x) < 0) \bb{P}_{\mu_{-1}} (f(x) < 0) + \bb{E}_{\mu_{-1}} (f(x)^\gamma | f(x) \geq 0) \bb{P}_{\mu_{-1}} (f(x) \geq 0) \Bigr).
\end{align*}
Using Jensen's inequality, we have $\mathbb{E}(X^\gamma) \leq \mathbb{E}(X)^\gamma$, for any random variable $X$, and $\gamma \leq 1$. Using this inequality together with $\mathbb{P} (A) \leq \mathbb{P} (A)^\gamma$, we obtain
\begin{align*}
\rho_{\text{adv}} (f) & \leq \tau \Bigl( \left( p_1 \bb{E}_{\mu_1} (f(x) | f(x) \geq 0)\bb{P}_{\mu_1} (f(x) \geq 0) \right)^{\gamma} + \left( p_1 \bb{E}_{\mu_{1}} (-f(x) | f(x) < 0) \bb{P}_{\mu_{1}} (f(x) < 0) \right)^{\gamma} \\
							  & + \left( p_{-1} \bb{E}_{\mu_{-1}} (-f(x) | f(x) < 0) \bb{P}_{\mu_{-1}} (f(x) < 0) \right)^{\gamma} + \left( p_{-1} \bb{E}_{\mu_{-1}} (f(x) | f(x) \geq 0) \bb{P}_{\mu_{-1}} (f(x) \geq 0) \right)^{\gamma} \Bigr).
\end{align*}
We use the result in Lemma \ref{lemma:lemma_inequality} with $n=4$, and obtain
\begin{align*}
\rho_{\text{adv}} (f) & \leq \tau 4^{1-\gamma} \Bigl( p_1 \bb{E}_{\mu_1} (f(x) | f(x) \geq 0)\bb{P}_{\mu_1} (f(x) \geq 0) + p_1 \bb{E}_{\mu_{1}} (-f(x) | f(x) < 0) \bb{P}_{\mu_{1}} (f(x) < 0) \\
							  & + p_{-1} \bb{E}_{\mu_{-1}} (-f(x) | f(x) < 0) \bb{P}_{\mu_{-1}} (f(x) < 0) + p_{-1} \bb{E}_{\mu_{-1}} (f(x) | f(x) \geq 0) \bb{P}_{\mu_{-1}} (f(x) \geq 0) \Bigr)^{\gamma}.
\end{align*}
Note moreover that the following equality holds
\begin{align*}
& -p_1 \bb{P}_{\mu_1}(f(x) < 0) \bb{E}_{\mu_1} (f(x) | f(x) < 0) \\ 
& = 2 p_1 \bb{P}_{\mu_1}(f(x) < 0) | \bb{E}_{\mu_1} (f(x) | f(x) < 0) | + p_{1} \bb{P}_{\mu_1}(f(x) < 0) \bb{E}_{\mu_1} (f(x) | f(x) < 0),
\end{align*}
Using the above equality along with a similar one for $p_{-1} \bb{P}_{\mu_{-1}} (f(x) \geq 0) \bb{E}_{\mu_{-1}} (f(x) | f(x) \geq 0)$, the following upper bound is obtained
\begin{align*}
\rho_{\text{adv}} (f) & \leq \tau 4^{1-\gamma} \Bigl( p_1 \bb{E}_{\mu_1} (f(x) | f(x) \geq 0)\bb{P}_{\mu_1} (f(x) \geq 0) + p_1 \bb{E}_{\mu_{1}} (f(x) | f(x) < 0) \bb{P}_{\mu_{1}} (f(x) < 0) \\
							  & - p_{-1} \bb{E}_{\mu_{-1}} (f(x) | f(x) < 0) \bb{P}_{\mu_{-1}} (f(x) < 0) - p_{-1} \bb{E}_{\mu_{-1}} (f(x) | f(x) \geq 0) \bb{P}_{\mu_{-1}} (f(x) \geq 0) \\
							  & + 2 p_1 \mathbb{P}_{\mu_1} (f(x) < 0) |\mathbb{E}_{\mu_1}(f(x)|f(x)<0| + 2 p_{-1} \mathbb{P}_{\mu_{-1}} (f(x) \geq 0) |\mathbb{E}_{\mu_{-1}}(f(x)|f(x) \geq 0|) \Bigr)^{\gamma},
\end{align*}
which simplifies to
\begin{align*}
\rho_{\text{adv}} (f) & \leq \tau 4^{1-\gamma} \Bigl( p_1 \mathbb{E}_{\mu_1} (f(x)) - p_{-1} \bb{E}_{\mu_{-1}} (f(x)) + 2 p_1 \mathbb{P}_{\mu_1} (f(x) < 0) |\mathbb{E}_{\mu_1}(f(x)|f(x)<0)| \\ 
& + 2 p_{-1} \mathbb{P}_{\mu_{-1}} (f(x) \geq 0) |\mathbb{E}_{\mu_{-1}} (f(x)|f(x) \geq 0)| \Bigr)^{\gamma},
\end{align*}
Observe moreover that $R(f) = p_1 \bb{P}_{\mu_1} (f(x) < 0) + p_{-1} \bb{P}_{\mu_{-1}} (f(x) \geq 0)$, and that $|\mathbb{E}_{\mu_{-1}} (f(x)|f(x) \geq 0)|$ is bounded from above by $\| f \|_{\infty}$. We therefore conclude that 
\begin{align*}
\rho_{\text{adv}} (f) & \leq \tau 4^{1-\gamma} \Bigl( p_1 \mathbb{E}_{\mu_1} (f(x)) - p_{-1} \bb{E}_{\mu_{-1}} (f(x)) + 2 R(f) \| f \|_{\infty} \Bigr)^{\gamma}.
\end{align*}

\end{proof}

\subsection{Proof of Theorem \ref{th:uniform_noise_linear}}
\label{sec:proof_theorm_proba}
The proof of this theorem relies on the concentration of measure on the sphere. The following result from \citep{matouvsek2002lectures} precisely bounds the measure of a spherical cap.
\begin{theorem}
Let $\mathscr{C} (\tau) = \{ x \in \bb{S}^{d-1}: x_1 \geq \tau \}$ denote the spherical cap of height $1-\tau$. Then for $0 \leq \tau \leq \sqrt{2/d}$, we have $\frac{1}{12} \leq \bb{P} (\mathscr{C} (\tau)) \leq \frac{1}{2}$, and for $\sqrt{2/d} \leq \tau < 1$, we have:
\begin{align*}
\frac{1}{6 \tau \sqrt{d}} (1 - \tau^2)^{\frac{d-1}{2}} \leq \bb{P} (\mathscr{C}(\tau)) \leq \frac{1}{2 \tau \sqrt{d}} (1 - \tau^2)^{\frac{d-1}{2}}.
\end{align*}
\label{th:matousek}
\end{theorem}
\noindent Based on Theorem \ref{th:matousek}, we show the following result:
\begin{lemma}
\label{lemma:lemma_concentration_measure}
Let $w$ be a vector of unit $\ell_2$ norm in $\bb{R}^d$. Let $\tau \in [0, 1)$, and $x$ be a vector sampled uniformly at random from the unit sphere in $\bb{R}^d$. Then,
\begin{align*}
\frac{1}{12} (1-\tau^2)^{d} \leq \bb{P} (\{w^T x \geq \tau\}) \leq 2 \exp\left( - \frac{\tau^2 d}{2} \right).
\end{align*}
\end{lemma}
\begin{proof}
Using an appropriate change of basis, we can assume that $w = (1, 0, \dots, 0)^T$.
For $\tau \in [\sqrt{2/d}, 1)$, we have
\begin{align*}
\bb{P} (\{x_1 \geq \tau\}) & \overset{(a)}{\leq} \frac{1}{2 \tau \sqrt{d}} (1 - \tau^2)^{\frac{d-1}{2}} \overset{(b)}{\leq} 2 \exp(-\tau^2 d/2),
\end{align*}
where (a) uses the upper bound of Theorem \ref{th:matousek}, and (b) uses the inequality $(1 - \tau^2) \leq \exp(-\tau^2)$. 
Note moreover that for $\tau \in [0, \sqrt{2/d})$, the inequality $2 \exp(-\tau^2 d/2) \geq 2 \exp(-1) \geq 1/2$ holds, which proves the upper bound. 

We now prove the lower bound. Observe that the following lower bound holds for $(\tau \sqrt{d})^{-1}$, for any $\tau \in [\sqrt{2/d}, 1)$:
\begin{align*}
\frac{1}{\tau \sqrt{d}} \geq \exp(-\tau^2 d/2).
\end{align*}
To see this, note that the maximum of the function $a \mapsto \ln(a)/a^2$ is equal to $1/(2e) \leq 1/2$. Therefore, $\ln(\tau \sqrt{d}) / (\tau^2 d) \leq 1/2$, or equivalently, $(\tau \sqrt{d})^{-1} \geq \exp(-\tau^2d/2)$.
Therefore, we get $\frac{1}{\tau \sqrt{d}} \geq (1 - \tau^2)^{d/2}$, and using Theorem \ref{th:matousek}, we obtain for any $\tau \in [\sqrt{2/d}, 1)$:
\begin{align*}
\bb{P} (\{x_1 \geq \tau\}) \geq \frac{1}{6 \tau \sqrt{d}} (1-\tau^2)^{\frac{d-1}{2}} \geq \frac{1}{12} (1-\tau^2)^{d}.
\end{align*}
Note also that this inequality holds for $\tau \in [0, \sqrt{2/d}]$, as $\frac{1}{12} (1-\tau^2)^{d} \leq \frac{1}{12}$. 
\end{proof}

\noindent Armed with the concentration of measure result on the sphere, we now focus on the proof of Theorem \ref{th:uniform_noise_linear}.
Let $f(x) = w^T x + b$. Let $x$ be fixed such that $f(x) > 0$, and let $\eta > 0$ and $\epsilon \in (0, 1/12)$. Then, 
\begin{align*}
\mathbb{P}_{n \sim \eta \bb{S}} \left( f(x+n) \leq 0 \right) & = \mathbb{P}_{n \sim \eta \bb{S}} \left( w^T n \leq -w^T x -b \right) \\ 
																 & = \mathbb{P}_{n \sim \eta \bb{S}} \left( w^T n / \| w \|_2 \leq - \Delta_{\text{adv}} (x;f) \right) \\
																 & = \mathbb{P}_{n \sim \bb{S}} \left( w^T n / \| w \|_2 \leq - \Delta_{\text{adv}} (x;f) / \eta \right) \end{align*}
Using the upper bound in Lemma \ref{lemma:lemma_concentration_measure}, we obtain:
\begin{align*}
\mathbb{P}_{n \sim \eta \bb{S}} \left( f(x+n) \leq 0 \right) \leq
2 \exp\left( -\frac{\Delta_{\text{adv}} (x;f)^2 d}{2 \eta^2} \right).
\end{align*}
Therefore, for $\eta = (2 \ln(2/\epsilon))^{-1/2} \sqrt{d} \Delta_{\text{adv}} (x;f) = C_1(\epsilon) \sqrt{d} \Delta_{\text{adv}} (x;f)$, we obtain $\bb{P}_{n \sim \eta \bb{S}} ( f(x+n) \leq 0) \leq \epsilon$, and we deduce that 
\begin{align*}
\Delta_{\text{unif}, \epsilon} (x;f) \geq C_1(\epsilon) \sqrt{d} \Delta_{\text{adv}} (x;f).
\end{align*}
Using the lower bound result of Lemma \ref{lemma:lemma_concentration_measure}, we have:
\begin{align*}
\frac{1}{12} \left( 1 - \frac{\Delta_{\text{adv}}^2(x;f)}{\eta^2} \right)^{d} \leq \mathbb{P}_{n \sim \eta \bb{S}} \left( f(x+n) \leq 0 \right) 
\end{align*}
This implies that for any $\eta \geq \frac{\Delta_{\text{adv}} (x;f)}{\sqrt{1 - (12 \epsilon)^{1/d}}} = \widetilde{C_2}(\epsilon, d) \Delta_{\text{adv}} (x;f)$, we have $\mathbb{P}_{n \sim \eta \bb{S}} \left( f(x+n) \leq 0 \right)  \geq \epsilon$. Hence, we obtain the following upper bound on $\Delta_{\text{unif}, \epsilon} (x;f)$:
\begin{align*}
\Delta_{\text{unif}, \epsilon} (x;f) \leq \widetilde{C_2}(\epsilon, d) \Delta_{\text{adv}} (x;f).
\end{align*}
We also derive a lower bound on $\Delta_{\text{unif}, \epsilon} (x;f)$ of the form $C_2(\epsilon) \sqrt{d} \Delta_{\text{adv}} (x;f)$ by noting that
\begin{align*}
\widetilde{C_2}(\epsilon, d) d^{-1/2} = \frac{1}{\sqrt{d (1 - (12\epsilon)^{1/d})}} \leq \frac{1}{\sqrt{1-12\epsilon}} = C_2(\epsilon),
\end{align*}
where we have used the fact that $\frac{1}{\sqrt{d (1 - (12\epsilon)^{1/d})}} $ is a decreasing function of $d$. To see that this function is indeed decreasing, note that its derivative (with respect to $d$) can be written as $P(d) \left( d (\overline{\epsilon}^{1/d} - 1) - \overline{\epsilon}^{1/d} \ln(\overline{\epsilon}) \right)$, with $P(d)$ non-negative, and $\overline{\epsilon} = 12 \epsilon$. Then, by using the inequality $\ln((1/\overline{\epsilon})^{1/d}) \leq \left( 1/\overline{\epsilon} \right)^{1/d} - 1$, the negativity of the derivative follows.

\noindent By combining the lower and upper bounds, and taking the expectations on both sides of the inequality, we obtain:
\begin{align*}
C_1(\epsilon) \sqrt{d} \bb{E}_{\mu} \left( \Delta_{\text{adv}} (x;f) 1_{f(x) > 0} \right) \leq \bb{E}_{\mu} \left(\Delta_{\text{unif}, \epsilon} (x;f) 1_{f(x) > 0} \right) & \leq \widetilde{C_2} (\epsilon, d) \bb{E}_{\mu} \left( \Delta_{\text{adv}} (x;f) 1_{f(x) > 0} \right)  \\
& \leq C_2(\epsilon) \sqrt{d} \bb{E}_{\mu} \left( \Delta_{\text{adv}} (x;f) 1_{f(x) > 0} \right).
\end{align*}
A similar result can be proven for $x$ such that $f(x) \leq 0$. We therefore conclude that
\begin{align*}
\max(C_1(\epsilon) \sqrt{d}, 1) \rho_{\text{adv}} (f) \leq \rho_{\text{unif}, \epsilon} (f) & \leq \widetilde{C_2} (\epsilon, d) \rho_{\text{adv}} (f) \leq C_2(\epsilon) \sqrt{d} \rho_{\text{adv}} (f),
\end{align*}
where we have used the inequality $\rho_{\text{unif}, \epsilon} (f) \geq \rho_{\text{adv}} (f)$. 

\section{Vertical-horizontal example: quadratic classifier}
\label{sec:vertical_horizontal_details_quadratic}

We consider the quadratic classifier $f_{\text{quad}} (x) = x^T A x$, with 
\begin{align*}
A = \frac{1}{2} \begin{bmatrix} 0 & 1 & -1 & 0 \\ 1 & 0 & 0 & -1 \\ -1 & 0 & 0 & 1 \\ 0 & -1 & 1 & 0 \end{bmatrix}.
\end{align*}
We perform a change of basis, and work in the diagonalizing basis of $A$, denoted by $P$. We have
\begin{align*}
P & = \frac{1}{2} 
\begin{bmatrix} 1 & 1 & -1 & -1 \\ 0  & -\sqrt{2} & -\sqrt{2} & 0 \\ \sqrt{2} & 0 & 0 & \sqrt{2}\\ 1 & -1 & 1 & -1
                \end{bmatrix}, \\
A & = P^T \begin{bmatrix} 1 & 0 & 0 & 0 \\ 0 & 0 & 0 & 0 \\ 0 & 0 & 0 & 0 \\ 0 & 0 & 0 & -1 \end{bmatrix} P.
\end{align*}
By letting $\tilde{x} = P x$, we have: 
\begin{align*}
f_{\text{quad}} (\tilde{x}) = \tilde{x}_1^2 - \tilde{x}_4^2.
\end{align*}
Given a point $x$ and label $y$, the following problem is solved to find the minimal perturbation that switches the estimated label:
\begin{align*}
\min_{\tilde{r}} \tilde{r}_1^2 + \tilde{r}_4^2 \text{ s.t. } y ( (\tilde{x}_1 + \tilde{r}_1)^2 - ( \tilde{x}_4 + \tilde{r}_4)^2) \leq 0.
\end{align*}
Let us consider the first datapoint $x = [1+a, 1+a, a, a]^T$ (the other points can be handled in an exactly similar fashion). Then, it is easy to see that $\tilde{x}_1 = 1$ and $\tilde{x}_4 = 0$, and the optimal point is achieved for $\tilde{r}_1 = -1/2$ and $\tilde{r}_4 = 1/2$. In the original space, this point corresponds to $r = P^T \tilde{r} = [0, -1/2, 1/2, 0]^T$. Therefore, $\| r \|_2 = 1/\sqrt{2}$, and we obtain $\rho_{\text{adv}} (f_{\text{quad}}) = 1/\sqrt{2}$. 

\section{Discussion on the norms used to measure the magnitude of adversarial perturbations}
\label{sec:discussion_norms}
The goal of this section is to discuss different ways of measuring the robustness to adversarial perturbations.

Given a datapoint $x$, let $\eta > 0$ be such that we know a priori that all points in the region
\begin{align*}
\mathcal{R}(x) = \{ z: N(z - x) \leq \eta \},
\end{align*}
have the same true class as $x$ (i.e., a human observer would classify all images in this region similarly). Here $N: \mathbb{R}^d \rightarrow \mathbb{R}^+$ defines a norm in the image space. Note that $\mathcal{R}(x)$ only depends on the dataset, but does not depend on any classifier $f$. We defined the robustness of $f$ to adversarial perturbations, at $x$, to be
\begin{align*}
\Delta_{\text{adv}} (x; f) = \min_{r} N(r) \text{ subject to } f(x+r) f(x) \leq 0.
\end{align*}
The classifier $f$ is said to be \textit{not robust} at $x$ if 
\begin{align}
\label{eq:Delta_eta}
\Delta_{\text{adv}} (x; f) \leq \eta.
\end{align}
In words, this means that there exists a point $z$ in the region $\mathcal{R}(x)$ (i.e., $z$ and $x$ are classified in the same way by a human observer), but $z$ is classified differently than $x$ by $f$. Our main theoretical result  provides upper bounds to $\rho_{\text{adv}} (f)$ (the expectation of $\Delta_{\text{adv}} (x;f)$) in terms of interpretable quantities (i.e., distinguishability and risk): $\rho_{\text{adv}} (f) \leq U(\mu, R(f))$. Using this upper bound and Eq. (\ref{eq:Delta_eta}), we certify that $f$ is \textit{not} robust to adversarial perturbations when the following sufficient condition holds: 
\begin{align}
U(\mu, R(f)) \leq \eta.
\end{align}

The main difficulty in the above definitions lies in the choice of $N$ and $\eta$: how can $(N, \eta)$ be chosen to guarantee that $\mathcal{R}(x)$ contains all images of the same underlying class as $x$? In the original paper \citep{szegedy2013intriguing}, $N$ is set to be the $\ell_2$ norm, but no $\eta$ is formally derived; classifiers are said to be not robust to adversarial perturbations when $\rho_{\text{adv}} (f) / \sqrt{d}$ is judged to be ``sufficiently small''. For example, it appears from Table 1 in \citep{szegedy2013intriguing} that if $\rho_{\text{adv}} (f) / \sqrt{d} \lesssim 0.1$, the minimum required perturbation is thought to be small enough to guarantee that the images do not change their true underlying label. Motivated by the fact that pixels (or features) have limited precision, \citep{goodfellow2014explaining}
consider instead the $\ell_{\infty}$ norm, and ideally assume that a perturbation that have $\ell_{\infty}$ norm smaller than the precision of the pixels (e.g., 1/255 of the dynamic range for 8-bit images) is guaranteed to conserve the true underlying class. While this corresponds to setting $\eta$ to be the precision of the pixels, in practice it is set to be much larger for the MNIST case, as the images are essentially binary. In our case, the $\ell_2$ norm is considered, and we define the quantity $\kappa$ to be the average norm of the minimal perturbation required to transform a training point to a training point of the opposite class:
\begin{align*}
\kappa = \frac{1}{m} \sum_{i=1}^m \min_{j: y(x_j) \neq y(x_i)} \| x_i - x_j \|_2.
\end{align*}
We assume that the image $x+r$ is of the same underlying label as $x$ if $\| r \|_2$ is one order of magnitude smaller than $\kappa$. This corresponds to setting $\eta = \kappa / 10$. A summary of the different choices is shown in Table \ref{tab:comparison_different_N_eta}.

\begin{table}[ht]
\centering
\begin{tabular}{lcc}
\hline
& $N$ & $\eta$ \\ \hline
\citep{szegedy2013intriguing} & $\| \cdot \|_2$ & - \\ \hline
\citep{goodfellow2014explaining} & $\| \cdot \|_{\infty}$ & \tiny{Determined by the image precision (in theory). Larger in practice.} \\ \hline
Ours & $\| \cdot \|_2$ & $\kappa/10$ \\ \hline
\end{tabular}
\caption{\label{tab:comparison_different_N_eta} Different choices of $N$ and $\eta$ in different papers.}
\end{table}

All the above choices represent proxies of what we really would like to capture (i.e., the notion of perceptibility and class change). They all have some benefits and drawbacks, which we mention briefly now. We first acknowledge that the $\ell_{\infty}$ norm with $\eta \approx 0.1$ blocks class changes (and therefore provides a sufficient condition for certifying the non-robustness of classifiers) for images that are \textit{essentially binary} (e.g, MNIST digit images). In those cases, the $\ell_{\infty}$ norm seems more appropriate to use than the $\ell_2$ norm. 
However, in order to compare both norms, we need to carefully (and fairly) choose the $\eta$ parameter for both norms. 
In fact, if it is acknowledged that $N = \| \cdot \|_{\infty}$ and $\eta = 0.1$ provides a valid region $\mathcal{R}$ where underlying image classes do not change, then $N = \| \cdot \|_{2}$ and $\eta = 0.1$ \textit{ also provides a valid region}, as $\| r \|_{\infty} \leq \| r \|_{2}$ for any vector $r$. It is therefore all a matter of choosing a right threshold $\eta$ that is fair for all norms, if we wish to compare the norms for the task that we have at hand. 
A comparison between the $\ell_{\infty}$ and $\ell_2$ norm is provided in \citep{goodfellow_lecture}, and it is concluded that, while the $\ell_2$ norm allows for class changes within its region, the $\ell_{\infty}$ essentially blocks the class changes and therefore constitutes a better choice. In more details, the comparison goes as follows: it is first argued that by choosing $N = \| \cdot \|_2$ and $\eta = 3.96$, the region $\mathcal{R}$ contains both a ``natural'' $3$ and $7$, and therefore does not provide a valid region. To show the benefits of the $\ell_{\infty}$ norm, the author proceeds by considering $N = \| \cdot \|_{\infty}$ and $\eta = 3.96 / \sqrt{d} \approx 0.1414$. It is then argued that this region blocks previous attempts for class changes, and therefore the $\ell_{\infty}$ norm provides a better choice for the task at hand. 
While this type of comparison is important in order to reach a better understanding of the norms used to measure the adversarial examples, it is not conclusive as it is unfair to the $\ell_2$ norm.
Let us recall the following inequalities 
\begin{align}
\label{eq:equivalence_norms}
\forall r \in \mathbb{R}^d, \| r \|_{\infty} \leq \| r \|_{2} \leq \sqrt{d} \| r \|_{\infty}.
\end{align}
For a fixed $\eta_0 > 0$, define the regions:
\begin{align*}
\mathcal{R}_{\infty} & = \{ z: \| z - x \|_\infty \leq \eta_0 \}, \\
\mathcal{R}_{2} & = \{ z: \| z - x \|_2 \leq \eta_0 \sqrt{d} \}.
\end{align*}
It should be noted that for any $\eta_0$, we have $\mathcal{R}_{\infty} \subset \mathcal{R}_2$ using Eq. (\ref{eq:equivalence_norms}). Not only that, but $\mathcal{R}_{\infty}$ constitutes \textit{a tiny portion} of $\mathcal{R}_2$ in high dimensional spaces (i.e., the volume of $\mathcal{R}_{\infty}$ over that of $\mathcal{R}_2$ decays exponentially with the dimension). Therefore, a comparison of $\mathcal{R}_2$ to $\mathcal{R}_{\infty}$ will typically lead to problematic images in $\mathcal{R}_2$ but not in $\mathcal{R}_{\infty}$, as $\mathcal{R}_2$ is much bigger than $\mathcal{R}_{\infty}$. 
Therefore, the fact that $\mathcal{R}_{\infty}$ is a much smaller set than $\mathcal{R}_2$ (i.e., it contains much less images) is already known from Eq. (\ref{eq:equivalence_norms}) and is not conclusive in terms of the comparison of the two norms for measuring the robustness to adversarial perturbations.
Just like the comparison of $\mathcal{R}_2$ to $\mathcal{R}_{\infty}$ is unfair to the $\ell_2$ norm, saying that the $\ell_2$ norm is better than the $\ell_{\infty}$ norm because $\mathcal{R}_{\infty}$ contains much more images (potentially problematic ones with class changes, for sufficiently large $\eta_0$) that are not in $\mathcal{R}_{2}' = \{ z: \| z - x \|_2 \leq \eta_0 \}$ is unfair to the $\ell_{\infty}$ norm.

One possible way for providing a fair comparison between both norms is to find the coefficient $c$ such that $\mathcal{R}_{\infty}$ has the \textit{same volume} as the following $\ell_2$ ball
\begin{align*}
\mathcal{R}_{2}'' = \{ z: \| z - x \|_2 \leq \eta_0'' \sqrt{d} \}, \text{ with } \eta_0'' = \eta_0 c.
\end{align*}
Using mathematical derivations that we omit for the flow of this short discussion, we obtain $c = \sqrt{\frac{2}{e \pi}} \approx 0.48$ asymptotically as $d \to \infty$. 
We argue that the comparison of $\mathcal{R}_{\infty}$ to $\mathcal{R}_2''$ provides a more conclusive experiment than comparing $\mathcal{R}_{\infty}$ to $\mathcal{R}_2$, as it highlights the advantage of one norm with respect to the other without biases on the volume of the region.
In practice, this new comparison implies the following change for the ``3'' vs. ``7'' example in \cite{goodfellow_lecture}: instead of allowing perturbations of max-norm $0.1414$, perturbations with $\ell_\infty$ norm up to $\approx 0.3$ are allowed. This will result in images that are roughly twice as much perturbed, for the $\ell_{\infty}$ case. Even with this comparison, it is possible that the max-norm in this case will also block attempts to change the class, as the images are essentially binary. We believe that the $\ell_{\infty}$ is probably a better choice in this case.

\begin{figure}[t]
\centering
\subfigure[]{
\includegraphics[width=0.1\textwidth]{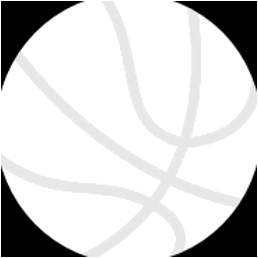}
}
\centering
\subfigure[]{
\includegraphics[width=0.1\textwidth]{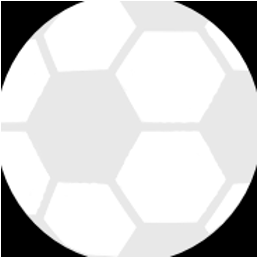}
}
\centering
\subfigure[]{
\includegraphics[width=0.1\textwidth]{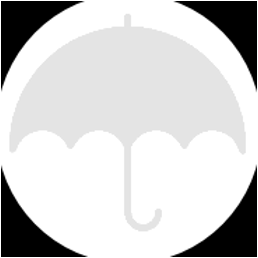}
}
\caption{\label{fig:toy_example_balls} Example images in a toy classification problem where the goal is to distinguish the different balls (a: basketball, b: soccer). (c) represents an umbrella that does not belong to any class. Black pixels are equal to $0$, white pixels are equal to $1$, grey pixels are set around $0.9$.}
\end{figure}

However, this is \textit{not} a general statement, as in some cases of non-binary images, the $\ell_2$ norm might be a better choice. We illustrate the above statement on a toy example where the goal is to classify sport balls. Some example images are shown in Fig. \ref{fig:toy_example_balls}. In this example, the $\ell_{\infty}$ norm between any two images is less than $0.11$. Setting $\eta_0 = 0.11$ with $N = \| \cdot \|_{\infty}$ does not define a valid region (i.e., it does \textit{not} guarantee that no class changes will occur within the region). On the other hand, the region $\mathcal{R}_2''$ computed with $\eta_0 = 0.11$ (i.e., $\eta_0'' = 0.0532$) rightfully excludes the images b) and c) from the space of valid perturbations of a). This toy example provides a proof of concept that shows that, in some cases, the $\ell_2$ norm might actually be a better choice than the $\ell_{\infty}$ norm.



In conclusion, we stress that this example has no intention of proving that the $\ell_2$ norm is universally better than the $\ell_{\infty}$ norm to measure the norm of adversarial perturbations. Through this discussion and example, we show that there is no universal answer to which norm one has to use to measure the robustness to adversarial perturbations, as it is strongly application-dependent. We believe more theoretical research in that area is needed in order to fully grasp the advantage of each norm, and probably design new norms that are suitable for measuring adversarial perturbations.

\end{document}